%% file: vonLaszewski-mlcommons-benchmark-carpentry.tex
\documentclass[conference]{IEEEtran}
\IEEEoverridecommandlockouts
\input{paper-format}

\definecolor{headerblue}{RGB}{100, 149, 237}
\definecolor{lightgray}{gray}{0.9}
\newcommand{\headerfont}{\fontsize{10pt}{12pt}\selectfont\bfseries}

\begin{document}

\newcommand{\TITLE}{AI Benchmark Democratization and Carpentry}

\onecolumn
\begin{center}
   {\bf\large \TITLE}
\end{center}

\tableofcontents
\twocolumn

\title{\TITLE}

\input{0000-authors}

\maketitle

\input{0010-abstract}
\input{0020-intro}
\input{0030-definitions}
\input{0040-formal}

\input{0050-benchmark-review}
\input{0100-profile}
\input{0060-review-gpu-benchmarks}
\input{0070-energy-benchmarks}
\input{0110-sim}
\input{0090-share}
\input{0400-carpentry}

\input{0410-democratizing}

\input{0530-conclusion}

\input{0910-acknowledgments}

\bibliographystyle{IEEEtran}
\bibliography{%
       ref-0-ontology,
       vonLaszewski-frontiers-citations,
       ref-carpentry,
       ref-sinclair,
       ref-energy,
       ref-hpcbench,
       ref-profile
     }
     
\input{0930-glossary}

\input{0940-contribution}

\end{document}

%% file: paper-format.tex
\usepackage[utf8]{inputenc}
\usepackage[T1]{fontenc}

\usepackage{lmodern}
\usepackage{textcomp}
\usepackage{microtype}
\usepackage{amsmath}

\usepackage{caption}
\usepackage{amsmath, amssymb, amsfonts}
\usepackage{bm} 
\usepackage{siunitx}
\DeclareSIUnit{\COtwoe}{\ensuremath{\mathrm{CO_2e}}}
\DeclareSIUnit{\gCOtwoe}{\ensuremath{\text{g\,CO_{2e}}}}
\DeclareSIUnit{\cent}{\text{\textcent}}

\usepackage{pdflscape}   
\usepackage{booktabs}    
\usepackage{tabularx}    
\usepackage{array}       
\usepackage[table]{xcolor}
\usepackage{longtable}
\usepackage{makecell}

\usepackage{listings}
\usepackage{xcolor}
\lstdefinestyle{pystyle}{
  language      = Python,
  basicstyle    = \ttfamily\small,
  keywordstyle  = \color{blue!70!black}\bfseries,
  commentstyle  = \color{green!50!black},
  stringstyle   = \color{orange!70!black},
  numbers       = left,
  numberstyle   = \tiny,
  numbersep     = 6pt,
  frame         = single,
  rulecolor     = \color{black!40},
  tabsize       = 4,
  showstringspaces = false,
  captionpos    = b
}

\usepackage{enumitem}
\newlist{goal}{enumerate}{1}
\setlist[goal,1]{label=\textbf{G\arabic*},leftmargin=1.8em,align=left}

\usepackage{graphicx}
\graphicspath{{ontology/images/}{images}}
\usepackage{tikz}
\usetikzlibrary{positioning,arrows.meta,shadows,trees,mindmap}
\usepackage{forest}
\usepackage{smartdiagram}
\usepackage[framemethod=tikz]{mdframed}

\forestset{
  skan tree/.style={
    for tree={
      drop shadow,
      text width=3cm,
      grow'=0,
      rounded corners,
      draw,
      top color=white,
      bottom color=blue!20,
      edge={Latex-},
      child anchor=parent,
      anchor=parent,
      tier/.wrap pgfmath arg={tier ##1}{level()},
      s sep+=2.5pt,
      l sep+=2.5pt,
      edge path'={
        (.child anchor) -- ++(-10pt,0) -- (!u.parent anchor)
      },
      node options={ align=center },
    },
    before typesetting nodes={
      for tree={
        content/.wrap value={\strut ##1},
      },
    },
  },
}

\usepackage[hyphens]{url}
\usepackage{hyperref}

\usepackage{comment}
\usepackage{stfloats}     
\usepackage{adjustbox}    
\usepackage{todonotes}    

\newcommand{\TODO}[1]{\todo[inline]{{\footnotesize #1}}}

\definecolor{darkgreen}{rgb}{0.0, 0.5, 0.0}

\definecolor{lightblue}{rgb}{0.7,0.85,1.0}

\DeclareUnicodeCharacter{2082}{\textsubscript{2}}

\makeatletter

\newcommand{\getcurrentlevel}{%
  \ifnum\c@subparagraph>0 5%
  \else\ifnum\c@paragraph>0 4%
  \else\ifnum\c@subsubsection>0 3%
  \else\ifnum\c@subsection>0 2%
  \else\ifnum\c@section>0 1%
  \else 0%
  \fi\fi\fi\fi\fi
}

\newcommand{\uselevel}[2]{%
  \ifcase#1%
    \section{#2}
  \or \section{#2}
  \or \subsection{#2}
  \or \subsubsection{#2}
  \or \paragraph{#2}
  \or \subparagraph{#2}
  \else \subparagraph{#2}
  \fi
}

\newcommand{\usedynlevel}[2]{%
  \edef\level{\number\numexpr\getcurrentlevel+#1\relax}%
  \ifnum\level>5 \def\level{5}\fi 
  \uselevel{\level}{#2}%
}

\makeatother

%% file: 0000-authors.tex
\author{
\IEEEauthorblockN{
Gregor von Laszewski\textsuperscript{1},
Wesley Brewer\textsuperscript{2},
Jeyan Thiyagalingam\textsuperscript{3},
Juri Papay\textsuperscript{3},
Armstrong Foundjem\textsuperscript{4}
}
\and
\IEEEauthorblockN{
Piotr Luszczek\textsuperscript{5},
Murali Emani\textsuperscript{6},
Shirley V. Moore\textsuperscript{7},
Vijay Janapa Reddi\textsuperscript{8},
Matthew D. Sinclair\textsuperscript{9}
}
\and
\IEEEauthorblockN{
Sebastian Lobentanzer\textsuperscript{10},
Sujata Goswami\textsuperscript{11},
Benjamin Hawks\textsuperscript{12},
Marco Colombo\textsuperscript{13},
Nhan Tran\textsuperscript{12}
}
\and
\IEEEauthorblockN{
Christine R. Kirkpatrick\textsuperscript{14},
Abdulkareem Alsudais\textsuperscript{15},
Gregg Barrett\textsuperscript{16},
Tianhao Li\textsuperscript{17},
Kirsten Morehouse\textsuperscript{18}
}
\and
\IEEEauthorblockN{
Shivaram Venkataraman\textsuperscript{9},
Rutwik Jain\textsuperscript{9},
Kartik Mathur\textsuperscript{20},
Victor Lu\textsuperscript{21},
Tejinder Singh\textsuperscript{22},
Khojasteh Z. Mirza\textsuperscript{23},
}
\and
\IEEEauthorblockN{
~~~~~~~~
Kongtao Chen\textsuperscript{24},
Sasidhar Kunapuli\textsuperscript{25},
Gavin Farrell\textsuperscript{26},
Renato Umeton\textsuperscript{27},
Geoffrey C. Fox\textsuperscript{1}
}
\\
\IEEEauthorblockA{
\begin{tabular}[t]{@{}l@{}}
\textsuperscript{1}\textit{Biocomplexity Institute, University of Virginia}, Charlottesville, VA, USA \\
~~(laszewski@gmail.com, gcfexchange@gmail.com)\\
\textsuperscript{2}\textit{Oak Ridge National Laboratory}, Oak Ridge, TN, USA (brewerwh@ornl.gov)\\
\textsuperscript{3}\textit{Rutherford Appleton Laboratory, STFC}, Harwell Campus, UK \\
~~(t.jeyan@stfc.ac.uk, juri.papay@stfc.ac.uk)\\
\textsuperscript{4}\textit{DEEL, Polytechnique Montreal}, Montreal, Canada (a.foundjem@polymtl.ca)\\
\textsuperscript{5}\textit{LLSC, MIT Lincoln Laboratory}, Lexington, MA, USA (luszczek@icl.utk.edu)\\
\textsuperscript{6}\textit{Argonne National Laboratory}, Lemont, IL, USA (memani@anl.gov)\\
\textsuperscript{7}\textit{Computer Science Department, UTEP}, El Paso, TX, USA (svmoore@utep.edu)\\
\textsuperscript{8}\textit{Harvard University}, Boston, MA, USA (vj@eecs.harvard.edu)\\
\textsuperscript{9}\textit{Computer Sciences Department, Univ. of Wisconsin–Madison}, Madison, WI, USA \\
~~(sinclair@cs.wisc.edu, shivaram@cs.wisc.edu, rnjain@wisc.edu)\\
\textsuperscript{10}\textit{Helmholtz Center Munich}, Munich, Germany \\
~~(sebastian.lobentanzer@helmholtz-munich.de)\\
\textsuperscript{11}\textit{ALS, LBNL}, Berkeley, CA, USA (sujatagoswami@lbl.gov)\\
\textsuperscript{12}\textit{Fermilab}, Batavia, IL, USA (ntran@fnal.gov, bhawks@fnal.gov)\\
\textsuperscript{13}\textit{Discovery Partners Institute, UIUC}, Chicago, IL, USA (mcolom4@illinois.edu)\\
\textsuperscript{14}\textit{SDSC, UC San Diego}, San Diego, CA, USA (christine@sdsc.edu)\\
\textsuperscript{15}\textit{Prince Sattam bin Abdulaziz University}, Saudi Arabia (a.alsudais@psau.edu.sa)\\
\textsuperscript{16}\textit{Cirrus AI}, Johannesburg, South Africa (gregg.barrett@cirrusai.net)\\
\textsuperscript{17}\textit{Duke University}, Durham, NC, USA (tianhao.li@duke.edu)\\
\textsuperscript{18}\textit{Harvard University}, Cambridge, MA, USA (knmorehouse@gmail.com)\\
\textsuperscript{20}\textit{Microsoft}, Vancouver, BC, Canada (kartikmathur@microsoft.com)\\
\textsuperscript{21}\textit{Independent Researcher}, Tampa, FL, USA (victorjunlu@gmail.com)\\
\textsuperscript{22}\textit{Office of the CTO, Dell Technologies}, Santa Clara, CA, USA (Singh.Tejinder@Dell.com)\\
\textsuperscript{23}\textit{Cornell Tech, Cornell University}, New York, NY, USA (kzm6@cornell.edu)\\
\textsuperscript{24}\textit{Google}, Mountain View, CA, USA (kongtao@google.com)\\
\textsuperscript{25}\textit{Independent Researcher}, San Jose, CA, USA (sasidhar.kunapuli@gmail.com)\\
\textsuperscript{26}\textit{University of Padua}, Padua, Italy (gavinmichael.farrell@phd.unipd.it)\\
\textsuperscript{27}\textit{St. Jude Children's Research Hospital}, Memphis, TN (Renato.Umeton@stjude.org)
\end{tabular}
}
}

%% file: 0010-abstract.tex
\begin{abstract}
 
Benchmarks are one cornerstone of modern machine learning practice, providing standardized evaluations that enable reproducibility, comparison, and scientific progress. 
However, AI benchmarks are becoming increasingly complex, requiring special care, including AI focused dynamic workflows. This is evident by the rapid evolution of 
AI models in architecture, scale, and capability; the evolution of datasets; and deployment contexts continuously change, creating a moving target for evaluation.
Large language models in particular are known for their memorization of static benchmarks, which causes a drastic difference between benchmark results and real-world performance.
Beyond the accepted static benchmarks we know from the traditional computing community, we need to develop and evolve continuous adaptive benchmarking frameworks, as scientific assessment is increasingly misaligned with real-world deployment risks.
This requires the development of skills and education focused on benchmarks in the scientific community: {\em AI Benchmark Carpentry}. 

Drawing on our experience from MLCommons, educational initiatives, and government programs such as the DOE's Trillion Parameter Consortium, we identify key barriers that hinder the broader adoption, utility, and evolution of benchmarking in AI. These include substantial resource demands, limited access to specialized hardware, lack of expertise in benchmark design, and uncertainty among practitioners about how to relate benchmark results to their own application domains. Moreover, current benchmarks often emphasize peak performance on leadership-class hardware, offering limited guidance for more diverse, real-world deployment scenarios. This may include applications to smaller compute resources, but also to larger systems such as LLMs deployed by commercial entities.

We argue that benchmarking itself must become dynamic in order to incorporate evolving models, updated data, and heterogeneous computational platforms while maintaining transparency, reproducibility, and interpretability. Democratizing this process requires not only technical innovation, but also systematic educational efforts as part of AI benchmark carpentry offerings, spanning undergraduate to professional levels, in order to develop sustained expertise in benchmark design and use. Finally, benchmarks should be framed and used to support application-relevant comparisons, enabling both developers and users to make informed, context-sensitive decisions. Advancing dynamic and inclusive benchmarking practices will be essential to ensure that evaluation keeps pace with the evolving AI landscape and supports responsible, reproducible, and accessible AI deployment. Furthermore, we believe that it is timely to provide a solid foundation for designing, using, and evolving benchmarks through community efforts that allows us to enable the concept of {\em AI benchmark carpentry.}

\end{abstract}



\bigskip

%% file: 0020-intro.tex
\section{Introduction}
\label{sec:intro}

Recently, the availability of graphics processing units (GPUs) and the rapid progress in artificial intelligence (AI) -- especially in the area of deep learning -- have brought a revolution to the scientific community. However, the use of these technologies is still in its infancy due to several factors. First, many application scientists are unsure how to leverage these newly available tools and instruments. Second, it remains unclear what level of effort is required to integrate them into their own research. Third, the specific demands these technologies place on infrastructure to be useful for a given scientific problem are not yet well understood.

Some of these challenges can be addressed by providing meaningful benchmarks to the scientific community, which can help researchers assess the usefulness and scalability of AI methods for their own applications. Therefore, it is beneficial to formalize the development of standardized AI benchmarks—not by a few individuals, but by the broader community. Such benchmarks can serve as a critical foundation for the scientific community, enabling rigorous evaluation, comparison, and reproducibility of new models and techniques.

However, as AI systems have become more sophisticated, incorporating complex and dynamic workflows, the traditional static approach to defining benchmarks has proven to be a significant limitation. In addition, to conventional benchmarks that capture key concepts familiar to scientists, we must also account for the continuous evolution of AI models and architectures, the changing nature of datasets, and the diversity of deployment contexts. These factors create a moving target for evaluation, risking a growing misalignment between benchmark results and the actual performance of AI systems in real-world scenarios.

Drawing on insights from our work with MLCommons, educational initiatives, and government-led projects such as the U.S. Department of Energy’s Trillion Parameter Consortium~\cite{trillion-parameter-consortium,Stevens2023-auroraGPT}, we identify a set of fundamental barriers that impede the broader utility and adoption of AI benchmarking.
Beyond the substantial resource demands and limited access to specialized, leadership-class hardware, there exists a pervasive lack of expertise in benchmark design and a growing uncertainty among practitioners regarding how to relate these performance metrics to their specific application domains.
Current benchmarks—by often prioritizing peak performance on elite hardware—offer insufficient guidance for the diverse range of computational platforms encountered in practice, from smaller-scale devices to large, pre-deployed commercial language models.

This paper argues that the practice of AI benchmarking itself must become dynamic and adaptable to keep pace with the rapidly evolving AI landscape.
To achieve this, benchmarks must be designed to transparently incorporate evolving models, updated datasets, and heterogeneous computational platforms, while upholding the core principles of transparency, reproducibility, and interoperability.
We propose that two complementary strategies can advance this goal: first, democratizing the creation of AI benchmarks and expanding the community contributing to them; and second, establishing a robust foundation for the technical execution and innovation of benchmarks through coordinated educational efforts.
Together, these approaches will foster sustained expertise spanning from undergraduate education to professional practice.

We believe it is both timely and necessary to establish a solid foundation for the design, use, and evolution of benchmarks through collaborative community efforts—thereby enabling what we call AI benchmark carpentry.
This paper summarizes the collective perspectives developed through this process within the MLCommons Science \& HPC Working Group.

The paper is organized as follows. In Section~\ref{sec:definition}, we introduce some essential definitions that we use throughout this paper.
Section~\ref{sec:formal} introduces a formal specification for AI benchmarks.
In Section~\ref{sec:benchmarks}, we summarize briefly some existing AI benchmark efforts.
In Section~\ref{sec:share}, we outline how to share benchmarks.
In Section~\ref{sec:edu}, we define activities to be conducted as part of the educational efforts.
In Section~\ref{sec:dem}, we identify what we need to do to conduct democratization efforts.
Lastly, we conclude 
in Section~\ref{sec:conclusion}.

Additionally, we list acronyms and abbreviations used in this paper in the Appendix A.
Contributions of the authors are summarized in the Appendix B. 

%% file: 0030-definitions.tex
\section{Definitions}
\label{sec:definition}

In this section, we introduce some of the definitions and terminology used throughout this work in order to work towards a formal definition of AI benchmarks.

\subsection{What is Benchmarking?}
\label{sec:definition-benchmarking}

In computing and scientific software evaluation, benchmarking is the process of comparing metrics for computer programs, models, or systems in order to assess their relative performance, typically with respect to a baseline. While early benchmarks focused largely on hardware throughput (e.g., the time required to complete a fixed computational task), modern benchmarks increasingly evaluate software, algorithms, and integrated systems. Three dimensions now structure most benchmarking efforts: 1) runtime—the amount of time a system requires to complete a set task; 2) accuracy—the comparative quality or correctness of outcomes for the same task; and 3) efficiency—the ratio between used computational resources and quality of outcomes. 

The goals of benchmarking include identifying performance gaps, establishing baseline expectations, driving innovation, and supporting continuous improvement over both short- and long-term horizons. Benchmarking has been extensively used in computer engineering and science—across both industry and academia—to measure the performance of computing equipment and the applications running on such systems.

In addition to the classical primary outcome metrics (runtime, accuracy, efficiency), today's benchmarks evaluate secondary qualities that are of high importance to the real-world deployment of systems. These include robustness and reliability (stability with respect to distribution shifts and noise, generalization), usability and accessibility (ease of integration with other systems, error transparency, ease of setup), and reproducibility (stability of the results and consistent behavior across versions, seeds, or environments).

\subsection{Lessons Learned from Traditional HPC Benchmarking}

Traditional high-performance computing (HPC) benchmarking includes:  

\begin{enumerate}
\item \textit{synthetic benchmarks} that simulate characteristic community workloads, as exemplified by the TOP500 and Green500 benchmarks;  
\item \textit{application benchmarks} that represent real-world applications to measure end-to-end performance, such as SPEC HPC; and  
\item \textit{scientific application benchmarks} that emphasize the accuracy of computational methods in solving domain-specific scientific problems.
\end{enumerate}

(For a more detailed discussion, see Section \ref{sec:hpc})

Important design and applicability criteria for benchmarks include relevance and representativeness for the field, fairness, repeatability, cost-effectiveness, scalability, and transparency~\cite{wikipedia:benchmarking}. One caveat is that vendors may optimize hardware specifically for these benchmarks, potentially neglecting new real-world problems and emerging challenges not captured by traditional benchmark suites.  

Therefore, it is essential to provide a diverse set of benchmarks so that different communities can evaluate and interpret results in terms of the performance metrics most relevant to their specific needs.  

HPC benchmarking has traditionally focused on supercomputing performance comparisons, targeting compute performance~\cite{Dongarra1989LinpackReport,Dongarra2016HPCG}, as well as memory, communication, and storage performance~\cite{PerfKitBenchmarker,IO500}. With the resurgence of AI and machine learning—including deep learning—it is now appropriate to explore additional lessons for benchmarking drawn from these domains.  

HPC benchmarks are often executed under controlled conditions, such as those maintained by system administrators, to ensure exclusive access to hardware and eliminate interference from other users or applications. This approach allows for measurement of the best achievable performance and is frequently used to guide system procurement decisions. However, such conditions do not reflect the shared nature of most computing environments, which often include factors such as queue wait times and concurrent multi-user workloads sharing hardware resources.

\subsection{What is Democratization?}
\label{sec:definition-democratization}

We believe it is vital not only to allow experts and power users to participate in benchmarking efforts but also to lower barriers to entry — making powerful benchmarks, tools, knowledge, and infrastructure available to everyone, not just those with specialized resources or expertise. For benchmarking, this implies in particular to improve the following:-

\begin{enumerate}
\item[a.] {\bf Accessibility:} Making benchmarks easier to use, enforcing  open-source licensing.

\item[b.] {\bf Open participation:} Encouraging community contributions through open-source development (e.g, on GitHub; shared repositories with transparent governance).

\item [c.] {\bf Knowledge sharing:} Providing tutorials, documentation, and educational resources so that non-experts can effectively use and modify the benchmarks.

\item[d.] {\bf Affordability:} Reducing cost barriers not only by introducing open source benchmarks, but also by allowing benchmarks to be offered at various scales and not only for leadership-class computing resources.

\end{enumerate}

\subsubsection{AI Software Democratization}

One of the major success stories in the field of artificial intelligence is the emergence of AI-specific software libraries such as TensorFlow, PyTorch, and Jupyter Notebooks. These tools have democratized machine learning and data science by making advanced computational capabilities accessible to students, researchers, and small organizations that previously lacked the resources to develop such tools from scratch.

\subsubsection{AI Hardware Democratization}

One must recognize that a significant amount of progress in AI research is conducted on campus computers that are much smaller than hyperscale AI machines or leadership-class government systems. Furthermore, many scientists have begun to use {\em desktop} computers equipped with high-powered graphics cards. Hence, it is important to have meaningful AI benchmarks available that allow for comparisons across different scales.

\subsection{What is Software Carpentry?}
\label{sec:towards-carp}

To set the stage for why we need AI benchmark carpentry, we need to first look at how the term has been introduced and is now commonly associated with software carpentry. After a more detailed analysis of software carpentry, we define the term AI benchmark carpentry.

Software Carpentry \cite{wilson2014software} was initially conceived to teach researchers in scientific fields fundamental computational and software development skills,  analogous to a hammer or level in a tool belt. Thus, non-computer scientists would be able to improve the use and development of the software they need to conduct their own research while benefiting from targeted, short educational tutorials. 

Today, a global community effort has sprung up since 1998~\cite{softwarecarpentry2024} that provides a number of training materials and sessions to the scientific community to we can leverage in some extend. Recently, additional areas beyond software, such as data carpentry.
Together, these efforts includes:

\begin{itemize}
  \item \textbf{Software Carpentry Core Efforts:} 
  Teaches researchers foundational computing skills to enhance their productivity and efficiency in research tasks. This includes lessons in 
  Programming with Python, Version Control with Git, The Unix Shell, Programming with R, Python, and using Git for version control.

 \item \textbf{Data Carpentry Efforts:}
  Teaches researchers skills necessary to work effectively and reproducibly with data in the context of specific domains. This includes lessons in the fiels of Astronomy, Ecology, Genomics, and Social Science with crosscutting topics such as Geospatial and Image Processing. Within those areas, are lessons such as Data Analysis and Visualization in R for Social Scientists, Foundations of Astronomical Data Science, and Introduction to the Command Line for Genomics \cite{datacarpentry2025}.

  \item \textbf{Other Carpentry Efforts:}
   Library Carpentry provides lessons for information scientists, data stewards, and roles in library science, reusing some of the Software Carpentry topics adapted in a curation context. Additional lessons available include High-Performance Computing (HPC Carpentry) \cite{reid2025hpc,HPCcarpentry2025}.

\end{itemize}

From this list, we see that benchmark carpentry is missing.

\subsection{What is Benchmark Carpentry?}
\label{sec:benchmark-carpentry}

Based on our observations in the educational and scientific communities  \cite{las-frontiers-edu}, we find that similar efforts are needed to focus on benchmarking.
This is more important as AI applications consume enormous resources, and properly scaling and using them requires a much deeper understanding of their time and space requirements.
The hope is that, from similar benchmarks, not only can the scientist learn lessons about their own applications, but, if needed, their own benchmarks can be developed to estimate costs and effort more precisely.
In addition, reproducible, portable benchmarks enable the selection and comparison of suitable hardware for the effort.

In general, we distinguish between hardware, software, and application components that significantly impact benchmarks.

On the hardware side, we deal with compute-oriented components such as CPUs, GPUs, and/or AI/neural accelerators (NPUs). Benchmarking them in the traditional way includes processing speed, core utilization, and instruction efficiency of a computer's central processing unit, data movement between xPU and main memory, to name a few. However, for AI, we also need performance in parallel computation, as well as AI workloads derived from AI kernels and applications.

As many AI applications require a large amount of {\em data} to be moved between memory, disks, CPU, and GPU memory, evaluating bandwidth, latency, and throughput is critical to understanding their impact on system performance. Hence, estimating and measuring the impact of, for example, assessing read/write speeds, IOPS, and access latency to identify bottlenecks in data storage systems is important.

Related to this is the {\em Network performance} metric, which measures bandwidth, latency, and packet loss to ensure efficient data transfer across systems, especially when parallel processing is used to address the scale required for good performance.

Benchmark carpentry should also teach 

{\em System Profiling and Monitoring} principles and tools so as to measure real-time system metrics.
{\em Interpreting Results, Analyzing Bottlenecks}, and {\em Optimizing Performance} are essential skills to identify limitations and improve overall performance through iterative strategies.

{\em Benchmark Design and Reproducibility} are similarly essential to allow comparative analyses among heterogeneous and also decentral benchmark runs.
This includes fair, repeatable benchmarks that reflect real-world workloads and enable comparative analysis of the different components involved.

%% file: 0040-formal.tex
\section{Towards a formal specification for AI benchmarks}
\label{sec:formal}

As part of the MLCommons Science Working group meetings, we have identified that ingredients of ML benchmarks include:

\begin{enumerate}
\item Datasets (such as images, application specific scientific data, time series)
\item Tasks to be performed 
\item Methods to perform these tasks (such as machine learning models, language models)
\item Metrics (runtime; accuracy; efficiency computed from the resources required for executing the task, such as  
  space, 
  memory usage, 
  energy efficiency, 
  power draw)
\item ML oriented performance impacts such as Latency impacted by the time per inference, Throughput for the  inferences per second, 
and training time to reach target accuracy.

\item Replication which includes the ability to replicate the experiment while at the same time being able in a structured fashion to compare the results.
\end{enumerate}

\subsection{Formalization}
\label{sec:towards-formal}

To formalize the specification of a benchmark we introduce the following notation

\[  B = (I, D, T~or~W, M, C, R, V) \]

\[
\begin{array}{ll}
B & = \text{Benchmark} \\
I & = \text{Infrastructure} \\
D & = \text{Dataset} \\
T,W & = \text{Scientific Task or Workflow}\\
M & = \text{Metrics} \\
C & = \text{Constraint} \\
R & = \text{Results} \\
V & = \text{Version or Timestamp}
\end{array}
\]

Further we define the task to be executed as an application applied to a set of parameters.

\[ T = (A, P) \]

\[
\begin{array}{ll}
A & = \text{Application} \\
P & = \text{Parameters}
\end{array}
\]

Alternative to a task, a workflow W can be used, if it contains multiple tasks that need to be conducted to achieve the scientific task (see Section \ref{sec:workflow}). 

Each of $B, I, D, T, M, R, A$ can have constraints $C_c$, where 

\[c \in \{B, I, D, T, M, R, A\}\]

In case of static benchmarks, many of the parameters may be fixed. However, when defining dynamic benchmarks, we define a metric that is to be minimized while allowing a predefined set of parameters of the benchmark to be variable. Let $B_i(M)$ denote a benchmark with a fixed metric M and variations in I, D, T, C, R  specified by i. We try to identify the minimum 

\[ min \{ B_i(..., M, ...)(S_j) \mid \forall_j M(S_j)\} \]

where $M(S_j)$ is the value of the solution for the metric and $S_j$ identifies a solution parameter set for the given metric. Please note that due to the statistical nature of the AI algorithms used in the benchmark, multiple solutions exist. However, we are not suggesting to conduct an exhaustive search of all possible solutions.

Let us assume M denotes the scientific accuracy of the benchmark; then, we look for the best scientific solution. Frequently, other restrictions are applied to the benchmark to make it tractable. While it is common to restrict the dataset, variation of the tested algorithm (the function we minimize) is often desired, since the scientific community is often not only interested in comparing hardware, but also in finding the best algorithmic solution. Such a solution can then be further studied with respect to efficiency or cost metrics.

Next, we briefly describe each of the parts that comprise a benchmark in more detail.

\subsection{Infrastructure}
\label{sec:towards-infra}

Infrastructure refers to the computational and software environment required to execute the scientific task.

This includes computational hardware, software libraries, operating systems, and cloud platforms, but also power related infrastructure to operate the resources.
In many cases some of these parameters are targeted by the benchmark for comparison (e.g., different types of GPUs).
As a guiding principle, an attempt should be made for each single benchmark to be clearly described with as many infrastructure parameters as possible. This will foster a clear description, reproducibility, and comparability of the benchmark.

Clearly defined infrastructure will help with (a) reproducibility, as it ensures results can be reproduced across different environments, (b) fairness, as it identifies clearly the differences between different hardware and  software used, (c) scalability as through comparison we can identify various scalability issues and properties, (d) efficiency, as we can assess  resource use in regards to common metrics such as time, space, energy, and cost.

\subsection{Dataset}
\label{sec:towards-dataset}

A dataset or multiple datasets provide the input data for the scientific task to be performed.
Datasets in benchmarking need to be stratified into training data (used to develop a machine learning model by direct interaction with the data), validation data (used to develop a machine learning model by indirect interaction, i.e., hyperparameter tuning), and test data (used to evaluate machine learning model performance after training).
If the benchmark is concerned with hardware performance, not training any machine learning model, only a test dataset might be needed.
In many cases, it is important to provide different sizes of data sets to enable (a) a small set for fast development of the approach, and (b) a larger set that fosters scientific accuracy with longer run-times.
Intermediary sizes are also sometimes needed to adapt to available resource constraints to compare them on different scales.
Data should always be sufficiently described through metadata or documentation so their context within the scientific application can be determined.
Together, these facilitate the establishment of (a) a ground truth that serves as the basis for evaluating scientific accuracy (b) a relevant and representative example that is influential for the scientific application, and (c) the identification of bias for data-driven applications.

We distinguish two different data sets: static and dynamic.
If behavior can be tested statically, this is to be preferred; introduction of hyperparameters into a testing setup results in combinatorial explosion of possibilities, making some benchmarking approaches intractable or prohibitively expensive. In such case, constraints could be posed to restrict the benchmark to the most meaningful hyperparameters. In fact, doing this as part of the workflow could be an integral part of the benchmark. 
For instance, a standard runtime test of a given compute task on different GPUs does not require dynamic datasets, as it is not expected that the results will change over time; the hardware parameters are fully specified.
Recent efforts have shown that, in some cases, we need to consider live data ingestion into benchmarks, for example, in earth science or health care applications, to support real-time predictions.
We term such datasets {\em living datasets}, which are continuously updated with new data, edge cases, or corrections. Such living data sets are a special case of dynamic datasets.
Such living datasets could be real-time data, but they could also be simulated using a static dataset while ingesting the data over time. While modifying the dataset the benchmark could evolve over time as the data available may be growing or become more accurate, supporting the need to identify the most accurate solution.

Living datasets allow us to maintain the relevance of a benchmarking task over time while simultaneously reacting to changes in the benchmarked systems.

It can also be used to adapt the benchmark to issues like over- and underfitting.

One additional aspect is that it can be useful to simulate such datasets and observe the changes of the benchmark when such data sets are utilized. Activities such as developing digital twins promote such approaches.

\subsection{Scientific Task}
\label{sec:towards-task}
\label{sec:workflow}

The scientific task identifies the core challenge being evaluated while precisely identifying the purpose of the evaluated components. 
Typical tasks include classification, translation, reasoning, time series prediction, and planning.
Through its precise definition, it sets the scope of the benchmark and introduces the community to the task to be executed and/or measured.

In more complex situations, the task itself may be a scientific workflow comprised of interacting components. In that case we may use a graph specification of the scientific task that uses subtasks that interact through edges indicating data flows and temporal executions. In that case we can use $W$ instead of $T$ as the specification of a workflow with properly augmented edges. Each task could have its own benchmark.

Formally, $W = (T, E)$, where  $T$ represents the collection of all tasks 

\[
T = \{t_1, t_2, t_3, \dots, t_n\}
\]

where $n$ is the total number of tasks, and 
 $E$ indicates the dependencies between the tasks.
 
\[
E = \{(t_i, t_j) \mid t_i, t_j \in T, t_i \ne t_j\}
\]

where $(t_i, t_j) = (t_j, t_i)$.

The introduction of Workflows into the formal definition is also motivated by the recent introduction of {\em Agentic AI frameworks} to support automation and benchmarking of it.

\subsection{Metrics}
\label{sec:towards-metrics}

Metrics are quantitative measures used to assess the relative performance of the tested system in completing the scientific task.
It has been shown in much previous work that the selection of the metric is the most crucial part of the benchmarking process.

The choice of metric determines many other aspects of the benchmarking purpose.
For instance, by choosing runtime (e.g., wall clock time) as the main metric, it is strongly implied that the benchmark's main purpose is to find the fastest hardware or algorithmic implementation.
By choosing an accuracy metric (e.g., F1 score), it is instead implied that the predictive performance (e.g., in classification tasks) is the target of the benchmark.
Complex metrics can visualize trade-offs between the primitive metrics; for instance, a benchmark for the efficiency of a classification algorithm can weight its F1 score against the runtime (per sample inference speed), model size (in parameters), and energy requirements.

Implemented in this way, metrics can be used to establish a ranking of the benchmarked components, given they were measured in similar circumstances and under similar constraints.

\subsection{Constraints}
\label{sec:towards-constraints}

In many cases, it is necessary to constrain the benchmark to make the comparison tractable.
This may include limits to training, inference, model size, or the amount of data used. 
Introducing constraints can (a) improve fairness while executing the benchmark (b) address operational real-world limitations, and (c) simplify the experimental setup.
Constraints can be applied to any component of the benchmark, e.g., $C_I$, $C_D$, etc.

\subsection{Results}
\label{sec:towards-res}

A benchmark must produce clear easy to comprehend results to allow evaluation of the task performed and to perform unambiguous performance evaluation.
As described above, a major determinant of the informativeness of a benchmark is the choice of metrics.
Performance can be evaluated on main metrics (e.g., accuracy or runtime), but often also includes a grid search of various methods, models, and hyperparameters.
To simplify comparison, metric dashboards with charts and tables, as well as error analysis, are recommended.
This allows (a) analysis of progress over time, (b) informing stakeholders about model capabilities, (c) identifying limitations of the tested methods, and (d) establishing a potential leader board for selecting suitable candidates that may be applicable to similar scientific tasks.

%% file: 0050-benchmark-review.tex
\section{Review of Benchmark Related to this Effort}
\label{sec:benchmarks}

This section provides an overview of key benchmarking efforts that motivated our paper. We start with HPC benchmarks and also address MLCommons benchmark efforts.

\subsection{HPC Benchmarking}
\label{sec:benchmarks-hpc}
\label{sec:hpc}

HPC benchmarking has a great impact on the activities that we report here and we can learn a lot from these efforts. Some of the most known efforts are TOP500 and Green500.

\subsubsection{TOP500}
\label{sec:benchmarks-hpc-top500}

The list of world's largest supercomputers has been released biannually
for nearly 4 decades now and thus offers a number of important lessons
in designing sustainable benchmarks. At the heart of the TOP500 scoring procedure, which yields a ranked list of 500 supercomputing installations, is the LINPACK benchmark~\cite{dongarra2003hpl}, which bears the name of the
namesake software library~\cite{dongarra1979linpack} for
solving systems of linear equations. This linear solver package was
designed in the 1970s and implemented in FORTRAN. The user guide for the
library was published in 1979 and included a list of only 24
computers~\cite{dongarra1979linpack}. The following decades brought in various aspects of scaling into the
software, the list sizes, and the machines submitted for inclusion in the ranking as well as data and reporting information.

\subsubsection{Green500}
\label{sec:benchmarks-hpc-green500}

Power and energy play a dominant role in the modern world of high-performance and distributed computing, with multi-megawatt data centers and computing facilities abound in many locations across the globe.
The issues of excessive power draw and energy consumption
data in the mid-2000s~\cite{feng2005pwrprofsciapps,
cameron2005hpcpowerdistcompsciapps} culminated in a special working
group of cross-industry members~\cite{specpower2008, specpower}, combining the TOP500 ranking with the available power
draw information from the supercomputers to yield the ranking
called Green500~\cite{green500}. Since then, it is published
alongside the TOP500 ranking and continues to underscore the
importance of efficient energy use at large HPC installations.

\subsubsection{HPC innovation}
\label{sec:benchmarks-hpc-innov}

Besides the recognition of development of tools and software to
facilitate the use of HPC systems and foster democratization, power
consumption monitoring has been integrated at the various levels of HPC
facilities, from the processing and networking elements to the data
center level infrastructure. Also, by utilizing different floating-point
precisions~\cite{abdelfattah2021mxpsurvey} the applications improve
their efficiency and benefit from a great impact on the system
performance due to direct targeting of the specific architectural
designs.

The creation of leaderboards has led to a better understanding of the overall HPC system, but insights can be limited by misalignment of algorithm scaling and leaderboard projections.
To counter misalignment, benchmarks should closely resemble the scientific task to be benchmarked.
In some cases, it is informative to include end-to-end performance, including data storage limitations.

\subsection{Machine Learning Benchmarks}
\label{sec:ai-benchmakrs}
\label{sec:benchmarks-mlcommons}

Benchmarking in scientific machine learning (ML) has emerged as a critical area to guide algorithm development, enable fair comparisons towards progress and innovation, and facilitate reproducibility. The development of ML benchmarks for science is especially critical because of the multi-disciplinary nature of the development, often including domain experts, computing hardware developers, and ML researchers.  That, coupled with the variety of tasks and workloads, makes {\it high quality} benchmarking critical to making progress. 

To obtain an overview how many academic benchmarks have been published in well known public domain archives, so we queried arXiv~\cite{www-arXiv} and Google Scholar~\cite{www-google-scholar}. Note that according to Google, Google Scholar does not include all entries from arXiv, but it does include most of them. However, it also includes many more resources, so we expect a larger number from Google Scholer. As of Oct 1, 2025, we find 106 entries on arXiv when searching for the topic {\em ``AI benchmark''}.
executing equivalent queries in Google Scholar yields 2,490 entries for {\em ``AI benchmark''}. It is evident from this that a complete survey of these papers is difficult to achieve through manual inspection. In an upcoming effort, we plan to explore how to automatically categorize these entries using LLMs while implementing an agentic AI framework for it.

The vast number and diversity of scientific tasks poses challenges to finding a well-defined, high-quality benchmark for any given task.
To improve discoverability, we have cataloged in this paper all MLCommons benchmarks that have a result submission.
Secondly, we have developed an ontology \cite{www-las-mlcommons-benchmark-collection,www-mlcommons-science-benchmarks-paper} that allows users to identify suitable benchmarks.

\subsubsection{MLCommons}

MLCommons \cite{www-mlcommons} provides one of the most comprehensive and standardized ecosystems of AI benchmarking. It addresses training, inference, scientific computing, and domain-specific benchmarks. Most prominently, the MLPerf benchmark suite—covering datacenter, edge, mobile, and training applications—establishes industry-wide baselines for performance, accuracy, power efficiency, and quality of service across diverse model classes such as computer vision, language, recommendation, speech, and reinforcement learning. Additionally, it offers specialized evaluations including MLPerf Tiny for microcontroller-class devices, MLPerf Storage for I/O workloads, and MLPerf Science for large-scale scientific AI. Furthermore, MLCommons promotes the reproducibility through initiatives such as Croissant ML, a standardized metadata schema for datasets, and MLCube, a portable container-based model packaging standard. Additional domain-specific working groups in medical AI, multilingual speech, and responsible AI have recently expanded the targeted domains. 

We have provided a comprehensive list of benchmarks in Tables \ref{tab:benchmarks-mlcommons} and \ref{tab:llm_benchmarks_long}.  The tables contain information about the benchmark name, model, task, domain, model type, metrics, hardware, and a brief note. The evaluations of the AILuminate benchmarks can be found on the MLCommons Web pages and include (a) Safety / Jailbreak Tests, (b) LLM Safety Evaluation, (c)  Responsible AI / Alignment (d)  LLM (Decoder) (e)  Safety Rate, Toxicity Score (f) Cloud LLM APIs (g) Robustness and Alignment.

\subsubsection{Ontology}

To improve discoverability of suitable benchmarks for a given task, we introduce a definition and AI Benchmark ontology of scientific machine learning benchmarks, where benchmarks are classified and mapped to their scientific domain and machine learning task type in~\cite{www-mlcommons-science-benchmarks-paper}. This work grew out of the Web page created at  \cite{www-las-mlcommons-benchmark-collection}, \cite{www-mlcommons-benchmarks} and provides an easy to use interactive mechanism to query the cataloged benchmarks.

New AI benchmarks are added through an open submission workflow overseen by the MLCommons Science Working Group. Each submission is evaluated against a rubric of currently six categories (Software Environment, Problem Specification, Dataset, Performance Metrics, Reference Solution, Documentation) that assigns an overall rating and potential endorsement. The scoring framework enables stakeholders, researchers, domain scientists, and hardware vendors to identify representative subsets of benchmarks that align with their specific priorities. The ontology supports adding new scientific domains, AI/ML motifs, and computing motifs.   

A subset of information collected by the Web page is shown in Table \ref{tab:ontology}. It not only includes some elementary information about the benchmarks but also a perceived rating displayed as a radar chart. Such radar charts include ratings from 1-5, where 5 is the best rating. Ratings are identified for documentation, specification, software, metrics, dataset, and reference solution. The Web page not only includes an automatically generated report of all benchmarks in PDF format, but also a convenient online publication of the benchmarks with convenient search capabilities.

\input{0055-table-mlcommons-5}


\input{ontology/selected-new}

\subsection{Technical aspects of AI Benchmarks}

In addition to discoverability challenges, there are also technical issues that need to be addressed in dealing with democratization and AI benchmark carpentry.

\subsubsection{Workflows}
\label{sec:benchmarks-mlcommons-desktop}

There are many workflow frameworks that can support the AI Benchmark Workflow. Two of them 
are the Compute Coordinator and the Experiment Executer; they can be used in conjunction or separately \cite{las-2022-templated}. The Compute Coordinator allows hybrid infrastructure access from the benchmark application, while the Experiment Executor allows the repeated execution of templated benchmarks. Both produce results in a structured fashion so they can be combined from multiple experiments and multiple infrastructures in order to support the FAIR principles.

\subsubsection{Containerization}
\label{sec:benchmarks-mlcommons-hpc}

Benchmarking on HPC and even smaller machines can be simplified by providing containerized environments which not only enable easy deployment, but also can harmonize execution by providing stable operating system and software environments. In addition to portable makefiles, the uniform generation of containers can be leveraged between applications.
Although docker is today widely used to containerize applications, on HPC systems we find that limited root access on many HPC systems led to the development of apptainers.
Hence, AI benchmarking carpentry should include the development of software in apptainers directly or converting Docker containers to apptainers.

\subsubsection{System-Dependent Software and Deployment Variability}

Benchmarking can be complex if the software, libraries and infrastructure differ across systems.
To support coordinated  benchmarking across different machines, we have introduced a 
templates hybrid reusable computational analytics workflow management framework with cloudmesh. This framework has been be applied to multiple Deep Learning MLCommons Applications. The details are explained in  \cite{las-2022-templated}. Utilizing such workflow systems promotes adaptation as deployment and execution is typically included in the workflow specifications. However, it can also address adaptation and modifications to future improvements and porting to different hardware as a working template is already provided..
    
\subsubsection{Logging and Monitoring}
\label{sec:benchmarks-mlcommons-logging}

A variety of logging frameworks exist for AI Benchmark logging. This includes logging tools such as MLPerf logging. While such tools provide elementary logging features, their outputs are not human readable and require post processing. This is also an issue when running applications in interactive mode during debugging phases. For this reason, we have provided Cloudmesh-stopwatch that not only allows human readable format, but also allows automatic MLPerf logging (if desired) with a single line change in the code. Cloudmesh stopwatch supports Python, shell, and batch script execution, and employs a consistent log format across all three. 

In general, we distinguish between four types of monitoring: (a) Infrastructure Monitoring, (b) Application  Monitoring, (c) Training Monitoring, and (d) Model-Level Monitoring. A wide range of tools exists for each type, making it essential to identify those that provide effective functionality while remaining easy to use. TensorBoard is one example.

%% file: 0055-table-mlcommons-5.tex
\setlength{\tabcolsep}{2pt}
\renewcommand{\arraystretch}{1.1}

{\tiny
\onecolumn
\begin{landscape} 
\begin{longtable}
{|p{
0.1\textwidth}|p{
0.09\textwidth}|p{
0.1\textwidth}|p{
0.2\textwidth}|p{
0.2\textwidth}|p{
0.2\textwidth}|p{
0.2\textwidth}|p{
0.2\textwidth}|}
\caption{MLCommons Benchmarks}
\label{tab:benchmarks-mlcommons}
\\ \hline
\rowcolor{blue!30}
\textbf{Benchmark Name} & \textbf{Model} & \textbf{Task} & \textbf{Application Domain / Use Case} & \textbf{Model Type / Architecture} & \textbf{Metrics / KPIs} & \textbf{Hardware} & \textbf{Notes / Description} \\
\hline
\endfirsthead
\caption{MLCommons Benchmarks (Cont.)} \\
\hline
\rowcolor{blue!30}
\textbf{Benchmark Name} & 
\textbf{Model} & 
\textbf{Task} & 
\textbf{Application Domain / Use Case} & 
\textbf{Model Type / Architecture} & 
\textbf{Metrics / KPIs} & 
\textbf{Hardware} & 
\textbf{Notes / Description} \\
\hline
\endhead
\hline
\multicolumn{8}{|r|}{{\footnotesize Continued on next page}} \\
\endfoot
\hline
\endlastfoot
\hline
%
%
\rowcolor{gray!20} \multicolumn{8}{|l|}{\textbf{MLPerf Inference: Datacenter}} \\ \hline
 deepseek-r1 & DeepSeek R1 (671B params) & Reasoning / Code Generation & Knowledge \& Reasoning, Complex Problem Solving, Step-by-Step Planning & Large Language Model (LLM), Reasoning LLM, High context/output length (up to 20K tokens) & Accuracy: Exact Match, Code Evaluation; Latency: TTFT (Time to First Token), TPOT (Time Per Output Token)& Data Center GPUs (NVIDIA H100/H200) with massive VRAM, optimized for $671$B parameters. &The model's large output length emphasizes its use in complex reasoning chains. Requires powerful systems (e.g., multiple H100 GPUs). \\ \hline
 dlrm-v2-99 & DLRM-v2 & Recommendation & Personalized product/content recommendation (e.g., e-commerce, social media feeds) & Deep Learning Recommendation Model (DLRM), Sparse/Dense Architecture & Throughput: Queries Per Second (QPS); Latency: 99th Percentile Latency & Data Center CPUs and GPUs (NVIDIA B200/GB200/B300), prioritizing high I/O and memory bandwidth for massive embedding tables. &Tests high-throughput, low-latency deployment for online services with a 99\% latency constraint. \\ \hline
 dlrm-v2-99.9 & DLRM-v2 & Recommendation & Personalized product/content recommendation (e-commerce, social media feeds) & Deep Learning Recommendation Model (DLRM), Sparse/Dense Architecture & Throughput: Queries Per Second (QPS); Latency: 99.9th Percentile Latency &Data Center CPUs and GPUs (NVIDIA H200), often using higher precision to ensure quality target is met. &Tests high-throughput, very low-latency deployment for critical online services with a strict 99.9\% latency constraint. \\ \hline
 llama2-70b-99 & Llama 2 (70B params) & Large Language Model (LLM) Inference & General text generation, chat, summarization, and understanding & LLM, Transformer-based & Throughput: Tokens Per Second (TPS); Latency: TTFT, TPOT (99th Percentile) & Data Center GPUs (e.g., AMD MI300X/MI325X, NVIDIA B200/GB200/H100/H200/L40S, MS-Intel Arc Pro B60) in multi-GPU configurations, focused on high throughput and low latency. &Represents a larger LLM workload, measuring performance under a 99\% latency constraint. \\ \hline
 llama2-70b-99.9 & Llama 2 (70B params) & Large Language Model (LLM) Inference & General text generation, chat, summarization, and understanding & LLM, Transformer-based & Throughput: Tokens Per Second (TPS); Latency: TTFT, TPOT (99.9th Percentile) & Data Center GPUs (AMD MI300X/MI325X, NVIDIA B200/GB200/H100/H200/L40S, MS-Intel Arc Pro B60), often testing the limits of precision vs. speed trade-offs. &Represents a larger LLM workload, measuring performance under a stricter 99.9\% latency constraint. \\ \hline
 llama3.1-8b-datacenter & Llama 3.1 (8B params) & Summarization / Text Generation & Low-cost, high-volume LLM services, interactive code assistants & LLM, Transformer-based & Accuracy: ROUGE metrics (1, 2, L); Latency: TTFT $\le$2s, TPOT $\le$100ms (Server) & Single-node systems or smaller GPU clusters, used to lower the entry barrier for the MLPerf Training suite. &Benchmarks a smaller LLM for efficient deployment in both Data Center and Edge scenarios. \\ \hline
 llama3.1-405b & Llama 3.1 (405B params) & Large Language Model (LLM) Inference & Generative AI, high-capability models & LLM, Transformer-based & Throughput: Output Tokens per second; Latency: TTFT, TPOT &Large-scale AI Clusters and Supercomputers (requires hundreds of GPUs (NVIDIA B200/GB200/GB300/H100/H200) with high-speed interconnects). &One of the largest LLMs in the suite, demonstrating the need for advanced parallelism (tensor, pipeline) on high-end systems (e.g., NVIDIA H200). \\ \hline
 mixtral-8x7b & Mixtral (46.7B total params) & Large Language Model (LLM) Inference & generative AI, multilingual tasks & Mixture-of-Experts (MoE) LLM (activates $\approx$13B params per token) & Throughput: Tokens Per Second (TPS); Latency &Data Center GPUs (AMD MI300X/MI325X, NVIDIA H200/RTX PRO 6000), optimizing MoE architecture for low active compute per token. &Showcases the efficiency of MoE architecture, offering high quality with lower active compute cost than dense models. \\ \hline
 retinanet & Retinanet-ResNext50 & Object Detection & Identifying and localizing objects in images & Object Detection Model, often with ResNext backbone and FPN & Accuracy: mAP (mean Average Precision); Throughput: Samples Per Second &Data Center and Edge GPUs (NVIDIA GeForce RTX 4090/H200/L4-PCIe/L40S), measuring both throughput and latency under a $100$ms constraint. &A standard computer vision benchmark using the OpenImages dataset. \\ \hline
 rgat & Relational Graph Attention Network & Node Classification & Graph data analysis, social network processing, knowledge graphs & Graph Neural Network (GNN), Graph Attention Network (GAT) variant & Accuracy (on node classification); Throughput: Samples Per Second & Data Center GPUs (NVIDIA B200), specifically testing performance on irregular, graph-structured data.&Addresses graph-structured data and multi-relational graphs, testing system efficiency for complex graph workloads. \\ \hline
 stable-diffusion-xl & Stable Diffusion XL (SDXL) & Text-to-Image Generation & Generative AI for creating high-quality images from text prompts & Diffusion Model (Latent Diffusion) & Throughput: Images Per Second; Latency & Data Center and Professional GPUs (AMD MI325X,NVIDIA B200/H100/H200/L4-PCI/L40S/NVIDIA RTX PRO 6000), focusing on the speed of image generation (samples/second).&Represents the Text-to-Image Generative AI domain, measuring the speed of image synthesis. \\ \hline
 whisper & Whisper-Large-V3 & Automatic Speech Recognition (ASR) & Converting spoken audio to text & Encoder-Decoder Transformer, Speech-to-Text Model & Accuracy: WER (Word Error Rate), Word Accuracy (Acc); Latency & Data Center GPUs (NVIDIA B200/GB200/GeForce RTX 4090/H100/H200/L4-PCIe/L40S), measuring performance on a complex sequence-to-sequence model for speech.&An ASR benchmark on multilingual audio, measuring both encoder (audio feature) and decoder (token generation) performance. \\ \hline
%
%
\rowcolor{gray!20} \multicolumn{8}{|l|}{\textbf{MLPerf HPC}} \\ \hline
 CosmoFlow & CosmoFlow 3D CNN & Regression & Astrophysics, Cosmology (predicting properties of the universe from simulation data) & 3D Convolutional Neural Network (3D CNN) & Time to Quality (TTQ) (e.g., Time to reach validation MAE $\le 0.124$) & Supercomputers \& Large HPC Clusters (e.g., Fugaku, Perlmutter). Stresses distributed training, 3D data handling, and fast data I/O for massive volumetric datasets ($\approx 5$ TB) GPUs used for running this benchmark: NVIDIA A100/V100.&Uses massive 3D volumetric data ($\approx 5.1$ TB). Stresses memory bandwidth and interconnect.  \\
\hline
 DeepCAM & DeepCAM Encoder-Decoder & Semantic Segmentation & Climate Science, Extreme Weather Prediction (identifying atmospheric rivers, tropical cyclones) & Convolutional Encoder-Decoder (e.g., U-Net or DeepLab-like) & Time to Quality (TTQ) (e.g., Time to reach validation IoU $\ge 0.82$) & Supercomputers \& Large HPC Clusters. Stresses large-scale image processing, high-dimensional data (many channels), and efficient communication on systems with thousands of GPUs (A100/P100/V100). &Trained on massive, high-resolution 2D image data ($\approx 8.8$ TB). Stresses I/O and communication efficiency. \\
\hline
 OpenCatalyst & DimeNet++ & Regression & Computational Chemistry, Materials Science (discovering new catalysts for energy storage) & Graph Neural Network (GNN) & Time to Quality (TTQ) (Time to reach target energy/force prediction error) & Supercomputers \& Large HPC Clusters. Stresses performance on graph-structured data (atomic systems) and complex GNN operations that require high GPU utilization. GPUs used for running this benchmark: NVIDIA A100/P100/V100. & Models atoms and bonds as a graph structure. Benchmarks complex, irregular GNN workloads at scale.  \\
\hline
%
%
\rowcolor{gray!20}\multicolumn{8}{|l|}{\textbf{MLPerf Training}} \\ \hline
 BERT (Bidirectional Encoder Representations from Transformers) & NLP - Question Answering & General NLP, Text Understanding & Transformer (Encoder) & Time to Quality (TTQ) (F1 Score on SQuAD) & Data Center GPUs, Accelerators & CPU, Single GPU (e.g., NVIDIA A100/H100), or moderate clusters. &A foundational benchmark for Natural Language Processing tasks. \\
\hline
 DLRM-dcnv2 (Deep Learning Recommendation Model - DCNv2)& Recommendation Systems & E-commerce, Content Streaming, Personalized Ads & Deep Learning Recommendation Model w/ DCNv2 & Time to Quality (TTQ) ($\text{AUC}$ on Criteo 4TB) & Data Center GPUs, Specialized Accelerators &Large-scale GPU clusters with high-speed interconnects (e.g., InfiniBand) for distributed training. This benchmark was running on GPUs: NVIDIA B300/B200/GB200/H200/H100/H200.&Stresses memory bandwidth and communication for massive embedding tables. \\
\hline
 llama2-70b-lora & LLM Fine-Tuning & Customizing LLMs for specific enterprise tasks & Transformer with LoRA & Time to Quality (TTQ) (ROUGE Score) & Multi-GPU servers, Mid-size GPU clusters & High-end Multi-GPU servers or small clusters (e.g., systems with AMD MI300X/MI325X/MI350X/MI355X, NVIDIA B200/B300/H100/H200). &Measures the efficiency of Low-Rank Adaptation (LoRA) on a $\approx 70\text{B}$ parameter model. \\
\hline
 llama3.1-405b & LLM Pretraining & Generative AI, Foundational Model Development & Transformer-based LLM ($\approx 405\text{B}$ params) & Time to Quality (TTQ) (Log Perplexity) & Large-scale, Multi-node GPU clusters & Single Node or small GPU systems (e.g., a few GPUs per node) to keep the benchmark accessible.Benchmark running on GPUs: NVIDIA B200/B300/H200.&The largest, most compute-intensive benchmark for pretraining state-of-the-art LLMs.  \\
\hline
 RetinaNet & Object Detection & Autonomous Vehicles, Surveillance, Image Analysis & One-stage Object Detector (ResNet, FPN) & Time to Quality (TTQ) ($\text{mAP}$ on COCO) & Data Center GPUs, Cloud Instances &Single or multi-GPU systems (NVIDIA B200/H200/RTX Pro 6000), often used in both Datacenter and Edge devices for inference.&Measures performance for a core computer vision task: localizing and classifying objects. \\
\hline
 RGAT (Relational Graph Attention Network) & GNN - Node Classification & Drug Discovery, Social Network Analysis, Fraud Detection & Relational Graph Attention Network (R-GAT) & Time to Quality (TTQ) (Accuracy on IGBH) & Systems optimized for high-bandwidth interconnects & GPU-based systems (NVIDIA B200/B300/H100), optimized for workloads with complex, sparse data structures like graphs.&Focuses on the irregular memory access and communication patterns of Graph Neural Networks. \\
\hline
 Flux1 (stable-diffusion) & Text-to-Image Generation & Generative AI, Digital Art, Content Creation & Latent Diffusion Model (U-Net, Transformer) & Time to Quality (TTQ) ($\text{FID}$ and $\text{CLIP}$ Scores) & Multi-GPU servers, Cloud Instances & High-performance Single or Multi-GPU systems (especially for fast inference or training). This benchmark was running on: NVIDIA B200/GB200/GB300.&Benchmarks the training of a major generative model in the AI industry. \\
\hline
%
%
\rowcolor{gray!20}\multicolumn{8}{|l|}{\textbf{MLPerf Inference: Edge}} \\ \hline
 3D U-Net (99\%) & 3D U-Net & Medical Image Segmentation & Healthcare, Volumetric Imaging (e.g., MRI/CT) & 3D Convolutional Encoder-Decoder CNN & Accuracy (Dice Score), Latency, Throughput (QPS) & Data Center GPUs (e.g., NVIDIA A100/H100), high-performance computing (HPC) systems, specialized accelerators. &99\% of reference accuracy target. Typically runs in Offline scenario for batch processing of medical scans. \\
\hline
 3D U-Net (99.9\%) & 3D U-Net & Medical Image Segmentation &Healthcare, High-Fidelity Imaging & 3D Convolutional Encoder-Decoder CNN & Accuracy (Dice Score), Latency, Throughput (QPS) & Data Center GPUs (e.g., NVIDIA A100/H100), high-performance computing (HPC) systems, specialized accelerators. &99\% of reference accuracy target. Represents a stricter quality constraint, often requiring higher-precision compute (e.g., FP16 vs. INT8). \\
\hline
 llama3.1-8b-edge & Llama 3.1 (8B params) & Text Generation / Summarization & Edge AI, On-device LLMs, Interactive Assistants & Quantized Transformer (Decoder-only LLM) & Tokens Per Second (TPS), Latency (TTFT, TPOT), Power & Edge devices, mobile SoCs (System-on-Chips), smaller GPUs (MS-Intel Arc Pro B60), high-end CPUs. &Benchmarks a modern, smaller LLM variant optimized for performance and low-latency on resource-constrained Edge devices. \\
\hline
 resnet & ResNet50-v1.5 & Image Classification & Vision, Quality Control, Surveillance	 & CNN (Residual Network) & Accuracy (Top-1), Latency, Throughput (QPS) & Data Center GPUs (NVIDIA GeForce RTX 4090/RTX-2000E), Edge devices, Mobile SoCs, CPUs, specialized accelerators. &The foundational computer vision benchmark, often used as a baseline for measuring performance and efficiency across all MLPerf tiers. \\
\hline
 retinanet & RetinaNet-ResNext50 & Object Detection & Autonomous Vehicles, Advanced Security Systems	 & One-stage Object Detection (often with FPN) & Accuracy (mAP - mean Average Precision), Latency, Throughput (SPS) & Data Center GPUs (NVIDIA GeForce RTX 4090/4000/2000E), Edge devices, specialized detection accelerators.&Measures the system's ability to find and localize multiple objects in images. Uses the OpenImages dataset. \\
\hline
 stable-diffusion-xl & Stable Diffusion XL (SDXL) & Text-to-Image Generation & Generative AI, Digital Content Creation	 & Diffusion Model (Latent Diffusion with U-Net) & Images Per Second, Latency (Time to generate an image) & Data Center GPUs (e.g., NVIDIA H100/H200, AMD MI300 series), powerful consumer-grade GPUs.&Represents the high-compute generative AI domain. Measures the speed of synthesizing high-resolution images from text prompts. \\
\hline
 whisper & Whisper-Large-V3 & Automatic Speech Recognition (ASR) & Speech-to-Text Services, Live Transcription	 & Encoder-Decoder Transformer & Accuracy (WER - Word Error Rate, Word Acc), Tokens Per Second & Data Center GPUs (NVIDIA GeForce RTX 4090), Edge/Client devices for real-time transcription.&A modern, high-accuracy ASR benchmark, using a Transformer architecture that handles both audio encoding and token generation. \\
\hline
%
%
\rowcolor{gray!20}\multicolumn{8}{|l|}{\textbf{MLPerf Inference: Mobile}} \\ \hline
 MLPerf Mobile/Edge & MobileNetV4-Conv-L & Image Classification, Object Detection & Edge/Mobile AI, low-latency on-device vision tasks. & CNN / MobileNet Family (V4) & Latency (ms), Throughput (Inferences/sec), Top-1/Top-5 Accuracy, Average Precision (AP). & Mobile SoCs, Specialized Mobile Accelerators (e.g., Apple Neural Engine, Edge TPUs, dedicated DSPs) &The largest convolutional-only variant of MobileNetV4. Optimized via Neural Architecture Search (NAS) for better latency-accuracy trade-offs on mobile and embedded hardware. \\
\hline
 MLPerf Mobile/Edge & Mobile SSD Variants & Object Detection & Edge/Mobile AI, real-time detection on resource-constrained devices. & Single Shot Detector (SSD) + Mobile Backbone & Average Precision (AP) (e.g., COCO AP), Latency (ms), FPS. & Mobile SoCs (CPU, GPU, NPU/DSP), Edge AI Accelerators& Refers to models like SSD-MobileNet V1/V2/V3 which are standard mobile benchmarks.\\
\hline
 MLPerf Edge & SSD-MobileNet & Object Detection (Small) & Edge/Mobile AI, detection for systems with tight latency/power budgets. & Single Shot Detector (SSD) + MobileNet Backbone & Average Precision (AP), Latency (ms). & Mobile SoCs (CPU, GPU, NPU/DSP), Edge AI Accelerators &A specific variant that is an original, primary benchmark for MLPerf Inference: Edge.\\
\hline
 MLPerf Mobile/Edge & MobileNet V1–V4 & Image Classification, Feature Extractor & Efficient Vision Models, low-power and low-latency inference. & CNN (V1: Depthwise Separable Convs, V2: Inverted Residuals, V3: Squeeze-and-Excitation, V4: UIB/Mobile MQA) & MACs/FLOPs, Latency (ms).& Mobile SoCs (CPU, GPU, NPU/DSP, e.g., Qualcomm Snapdragon, Apple A-series), Microcontrollers (MCUs), Edge AI Accelerators (e.g., Google Edge TPU) &A progression of architectures from Google, all focused on minimal computational cost while maintaining high accuracy, crucial for all MLPerf Edge divisions. \\
\hline
 MLPerf Mobile & MobileNet V4 & Image Classification, Object Detection & Universally Efficient AI, aiming for state-of-the-art accuracy-latency trade-offs. & Hybrid (Convolutional + Attention - Mobile MQA) & Latency (ms)& Mobile SoCs (CPU, GPU, NPU/DSP, e.g., Qualcomm Snapdragon) &The latest generation, featuring the Universal Inverted Bottleneck (UIB) and Mobile MQA. \\
\hline
 MLPerf Mobile & MOSAIC & Image Segmentation & Mobile Image Segmentation, on-device image processing. & U-Net variant with a MobileNet-style backbone. & Mean Intersection over Union (mIoU), Latency (ms).& Mobile SoCs (CPU, GPU, NPU) &A common model used for segmentation tasks in the MLPerf Mobile suite. \\
\hline
 MLPerf Mobile & MobileDETs & Object Detection & Edge/Mobile AI, high-speed detection for mobile chips. & Model Family derived from Neural Architecture Search (NAS) & Average Precision (AP), Latency (ms).& Mobile SoCs (NPU/DSP emphasized), Edge AI Accelerators &A family of detectors specifically optimized for latency on mobile SoCs. \\
\hline
 MLPerf Tiny\/Mobile & BERT-Tiny\/ DistilBERT & Natural Language Processing (NLP) Tasks (e.g., Q\&A) & Mobile\/Edge NLP, faster, smaller language understanding on local devices. & Transformer \/ Distillation Models & Latency (ms), F1 Score (SQuAD), GLUE Score. & CPUs, GPUs, Edge AI Accelerators, Mobile SoCs (optimized for low-latency) &Smaller, compressed versions of BERT achieved through knowledge distillation for resource-constrained environments. \\
\hline
 MLPerf Mobile & Mobile-BERT & Natural Language Processing (NLP) Tasks & Edge/Mobile NLP, task-agnostic BERT for resource-limited devices. & Compressed Transformer (Bottleneck structures, Knowledge Distillation) &Latency (ms), F1 Score (SQuAD), GLUE Score.& CPUs, GPUs, Edge AI Accelerators, Mobile SoCs (optimized for low-latency) &Achieves competitive results to BERT-Base with much higher speed and smaller size. \\
\hline
 MLPerf Mobile & EDSR F32B5 & Image Super-Resolution (SR) & Image Enhancement, upscaling low-resolution images for improved quality.& Enhanced Deep Super-Resolution (EDSR) Network &Latency (ms), PSNR, SSIM. & GPUs, Custom Hardware/FPGAs, specialized ISP (Image Signal Processor) components. &A common, high-quality reference model for measuring performance on image enhancement\/quality tasks.\\
\hline
 MLPerf Mobile & Stable Diffusion & Text-to-Image Generation & Generative AI, creating high-resolution images from text prompts. & Latent Diffusion Model (LDM) (U-Net, VAE, CLIP Text Encoder) & Images/Query Per Second (Throughput), Latency (Time-to-Image), FID/CLIP Scores. &High-end GPUs (e.g., NVIDIA A100/H100, RTX series), high-power Workstations and Data Center Accelerators. &A critical benchmark for measuring performance on large, complex generative workloads.\\
\hline
%
%
\rowcolor{gray!20}\multicolumn{8}{|l|}{\textbf{MLPerf Inference: Tiny}} \\ \hline
 MLPerf Tiny v 0.5 & Keyword Spotting Model & Audio Classification & TinyML/MCU, always-on voice assistant, device wake-word detection. &Small CNN (e.g., DS-CNN) or RNN. & Latency (ms), Energy (Joules), Area Under the ROC Curve (AUC). & Microcontrollers (MCUs) (e.g., Arm Cortex-M4\/M7), Digital Signal Processors (DSPs), Tiny Neural Network Accelerators. &Detects a specific word (e.g., "Hey Google") from a stream of audio, running on a highly constrained power budget. \\
\hline
 MLPerf Tiny  v 0.5 & Visual Wake Words (VWW) Model & Image Classification (Binary) & TinyML/MCU, low-power sensing, person detection, motion-activated cameras. & Small CNN (e.g., MobileNet V1/V2 variant). & Latency (ms), Energy (Joules), AUC. &MCUs, low-power vision processors, small-scale embedded systems. & Determines if a person is present in the image (person/not-person). Much simpler and smaller than general ImageNet classification. \\
\hline
 MLPerf Tiny  v 0.5 & Image Classification Model & Image Classification (Multi-class) & TinyML/MCU, general object recognition on ultra-low-power sensors. & Very small CNN (e.g., ResNet-8 or Micro-CNN). & Latency (ms), Energy (Joules), Top-1 Accuracy (e.g., on CIFAR-10). & MCUs with limited RAM and Flash storage. &A more complex classification task than VWW, but still constrained to a very small model size. \\
\hline
 MLPerf Tiny  v 0.5 & Anomaly Detection (AD) Model & Time Series Anomaly Detection &TinyML/MCU, industrial predictive maintenance, system health monitoring. & Small Autoencoder or similar lightweight model. & Latency (ms), Energy (Joules), AUC. & MCUs, industrial IoT sensors, devices monitoring vibration or sound. &Learns a baseline of normal sensor data (e.g., machine vibrations) and flags deviations as anomalies. \\
\hline
%
%
\rowcolor{gray!20}\multicolumn{8}{|l|}{\textbf{MLPerf Client}} \\ \hline
 MLPerf Client & Llama 2 7B Chat & Code analysis, Content generation, Creative writing, Summarization (various lengths). &General-purpose AI, Dialogue/Chatbots, Client-side LLM inference on PCs. & Transformer, Decoder-Only, Instruction-Tuned (SFT + RLHF), 7 Billion parameters. & Time-to-First Token (TTFT), Tokens/Second (Throughput). & Client GPUs (e.g., AMD Radeon, Intel Arc), Integrated NPUs (e.g., Intel Core Ultra, AMD Ryzen AI), Data Center GPUs (e.g., NVIDIA A100/H100) for server-side inference. &A foundational model in the benchmark for measuring core client-side LLM performance.\\
\hline
 MLPerf Client & Llama 3.1 8B Instruct (8B parameters) & Generative AI workloads: Code analysis, Content generation, Creative writing, Summarization. &General-purpose AI, Instruction Following, Client-side LLM inference on PCs. & Transformer, Decoder-Only, Instruction-Tuned, 8 Billion parameters. & Time-to-First Token (TTFT) (Latency), Tokens/Second (Throughput). & Client PCs and Data Center/Cloud-based GPUs (optimized for both low-latency "Time to First Token" and high-throughput "Tokens Per Second"). &An updated and highly capable open-weight model, demonstrating improved performance and alignment over Llama 2. \\
\hline
 MLPerf Client & Phi 3.5 Mini Instruct & Reasoning (Math, Code, Logic), Long Context Query \& Summarization (up to 128K tokens).&Memory/Compute Constrained Environments, Low-Latency Applications, On-device deployment (AI PCs, mobile). & Dense Decoder-Only Transformer, Instruction-Tuned, 3.8 Billion parameters. & Time-to-First Token (TTFT) (Latency), Tokens/Second (Throughput). & Client GPUs, NPUs, and potentially high-end mobile/edge processors (optimized for on-device deployment). &A highly efficient and lightweight model optimized for speed and strong reasoning despite its small size.\\
\hline
 MLPerf Client & Phi 4 Reasoning 14B & Complex Reasoning (multi-step math, scientific, coding, planning), Generating detailed chain-of-thought traces. & Agentic applications, High-accuracy problem-solving, Applications requiring explainability. & Dense Decoder-Only Transformer, Reasoning-Focused SFT (and possible RLHF for Plus variant), 14 Billion parameters. & Time-to-First Token (TTFT) (Latency), Tokens/Second (Throughput), Accuracy on reasoning tasks. & High-performance Client PCs (Workstations) and Data Center GPUs (due to its larger size and focus on complex, token-intensive reasoning). &Included as an experimental model in the benchmark, specifically designed to emphasize logical and complex problem-solving.\\
\hline
%
%
\rowcolor{gray!20}\multicolumn{8}{|l|}{\textbf{MLPerf Storage}} \\ \hline
 MLPerf Storage & ResNet-50 & I/O Workload for Image Classification Training &General-purpose computer vision, low-latency image processing. & Convolutional Neural Network (CNN) & Max Supported Accelerators, Aggregate Throughput (MiB/s), Accelerator Utilization ($\ge 90\%$ required). & Data Center GPUs (NVIDIA A100/H100), Edge AI Accelerators, and high-end CPUs (widely used across all MLPerf divisions: Data Center, Edge, Tiny). &High IOPS Demand. Characterized by highly concurrent, random reads of many small data samples ($\approx 150 \text{ KB}$ each), stressing metadata and IOPS capability.\\
\hline
 MLPerf Storage & 3D U-Net & I/O Workload for Medical Image Segmentation Training & Healthcare/Radiology, Medical Image Analysis, 3D data processing. &3D U-Net (3D CNN) &Max Supported Accelerators, Aggregate Throughput (GiB/s), Accelerator Utilization ($\ge 90\%$ required). &High-end Data Center GPUs (NVIDIA A100/H100) and specialized high-throughput storage systems (MLPerf Storage benchmark focus).&High Bandwidth Demand. Characterized by concurrent random reads of very large data files ($\approx 140 \text{ MB}$ each), stressing sustained data throughput.\\
\hline
 MLPerf Storage & CosmoFlow & I/O Workload for Scientific Parameter Prediction Training & Scientific High-Performance Computing (HPC), Astrophysics. &3D Convolutional Neural Network (3D CNN) &Max Supported Accelerators, Aggregate Throughput (GiB/s), Accelerator Utilization ($\ge 70\%$ required). & Supercomputers \& HPC Clusters: Requires massive scale distributed training across hundreds or thousands of GPUs (e.g., utilizing NVIDIA H100s, Intel Gaudi, and specialized high-speed interconnects like InfiniBand). &CPU-Intensive Workload. Uses medium-sized samples ($\approx 2 \text{ MB}$), but the client-side processing is more CPU-heavy, leading to a slightly lower required accelerator utilization threshold.\\
\hline
%
%
\rowcolor{gray!20}\multicolumn{8}{|l|}{\textbf{MLPerf Automotive}} \\ \hline
\hline
 MLPerf Automotive & SSD-ResNet50 & 2D Object Recognition and Segmentation &
ADAS / Collision Avoidance, Lane Departure & Single Shot Detector (SSD) with ResNet-50 Backbone &
Latency, Throughput, $mAP$ (Accuracy) &Edge AI Accelerators, Embedded GPUs, and Automotive System-on-Chips (SoCs).& Baseline benchmark for camera-based detection on high-res (8MP) images. Used in v0.5. \\
\hline
 MLPerf Automotive & BEVFormer-Tiny & Camera-based 3D Object Detection & Autonomous Driving (L2+ to L4), Environmental Perception & Bird's Eye View (BEV) Transformer-based Network &
Latency, Throughput, $\text{mAP}$ (Accuracy) & High-compute Automotive SoCs, next-generation AI accelerators (specifically targeting transformer and multi-sensor fusion capabilities). &Represents state-of-the-art camera-only 3D perception. Used in MLPerf Auto v0.5. \\
\hline
 MLPerf Automotive & DeepLabV3Plus / PointPainting & Semantic Segmentation (as a component of 3D Detection) & Lidar-Camera Sensor Fusion, 3D Perception & DeepLabV3+ (for Segmentation) + PointPillars (for 3D Detection) & Latency ($p99.9$ percentile), Throughput, Accuracy) &Safety-critical Automotive SoCs, purpose-built AI processors for ADAS/AV, often requiring high-reliability and low-latency performance.&
DeepLabV3+ is the 2D segmentation part of the PointPainting sensor fusion pipeline. Used in MLPerf Inference v5.0 Automotive. \\
\hline
%
%
\rowcolor{gray!20}\multicolumn{8}{|l|}{\textbf{MLPerf Training:HPC}} \\ \hline
\hline
MLPerf Training:HPC & CosmoFlow &Prediction of Cosmological Parameters ($\Omega_m, \sigma_8, n_s, H$) &
Astrophysics, Cosmology, Scientific Simulation Parameter Prediction & 3D Convolutional Neural Network (3D CNN)&
Time-to-Train (Total time to reach a target quality metric), Aggregate Throughput (Models trained per unit of time in weak scaling). &Supercomputers \& Large Clusters (e.g., NVIDIA Selene, Perlmutter, Fugaku), utilizing thousands of interconnected High-Performance GPUs (e.g., NVIDIA A100/H100) and high-speed parallel file systems.&Trained on 3D volumetric data (dark matter distributions) from N-body simulations. The large, volumetric data introduces significant I/O challenges and stresses high-bandwidth interconnects and storage. \\
\hline
MLPerf Training:HPC & DeepCAM & Semantic Segmentation of Extreme Weather Events (e.g., atmospheric rivers, tropical cyclones) &
Climate Science, Weather Forecasting, Earth System Modeling & Convolutional Encoder-Decoder (U-Net variant) & Time-to-Train (Total time to reach a target quality metric), Aggregate Throughput (Models trained per unit of time in weak scaling). & Supercomputers \& Large Clusters, demanding high I/O bandwidth to handle the massive 8.8 TB climate datasets and requiring excellent strong-scaling performance. This benchmark was running on NVIDIA V100/A100.&Trained on massive, high-resolution, multi-channel images (e.g., $768 \times 1152$ pixels with $16$ channels). Features high computational intensity and large memory footprint per sample. \\
\hline
MLPerf Training:HPC & OpenCatalyst & Prediction of energy and forces for molecular systems (AI for materials science) & Catalyst Discovery, Computational Chemistry, Materials Science, Energy Storage &
Graph Neural Network (GNN), specifically DimeNet++ & Time-to-Train (Total time to reach a target quality metric), Aggregate Throughput (Models trained per unit of time in weak scaling). & Supercomputers \& Large Clusters, typically emphasizing the performance of GNNs, which stress different aspects of the system, like memory access patterns and graph-specific operations. This benchmark was running on NVIDIA V100/A100.& Predicts quantum mechanical properties of catalyst systems. Stresses complex data structures (graphs) and large-scale parallel processing. Uses the massive OC20 dataset. \\
\hline
%
%
\rowcolor{gray!20}\multicolumn{8}{|l|}{\textbf{MLCommons Science}} \\ \hline
\hline
MLCommons Science & Cloud Mask &  image processing / segmentation & Earth Observation, Segmentation model for the pixel classification in satellite images & U-Net deep neural network  &training and inference timing and scalability on the training across a number of GPUs;runtime of training and inference. & HPC Clusters \& High-Performance GPUs (e.g., NVIDIA A100/V100) running distributed training frameworks like PyTorch or TensorFlow, often benchmarked for large-scale data I/O. & Focuses on identifying and isolating cloud cover in high-resolution satellite imagery for subsequent analysis. \\ 
\hline
MLCommons Science & STEMDL & A universal classifier for the space group of solid-state materials.
 & Scientific Machine Learning (General benchmark suite) &CNN: ResNet, VGG, DenseNet &  top1 accuracy and F1 score (Macro)& HPC Systems of all sizes, used for general performance comparison across different hardware architectures and scaling tests. This benchmrak was running NVIDIA A100/V100.& The goals of this benchmark are to: (1) explore the suitability of machine learning algorithms in the advanced analysis of Convergent beam electron diffraction (CBED) and (2) produce a machine learning algorithm capable of overcoming intrinsic difficulties posed by scientific datasets. \\
\hline
MLCommons Science & CANDLE UNO & Cancer Drug Response Prediction &Life Sciences / Personalized Medicine   & Neural Networks(MLP)& TTT, Prediction Accuracy & HPC Systems (e.g., Summit, Polaris) and Cloud Environments, stressing both compute performance and workflow management for parameter sweep tasks.  This benchmark was running NVIDIA A100.& Benchmarks deep learning models for predicting the response of various cancer cell lines to different therapeutic compounds. \\
\hline
MLCommons Science & Earthquake & TEvolOp Earthquake Forecasting Model &Earthquake Science & Neural Networks(MLP)- recurrent neural networks and transformers &Nash Sutcliffe efficiency &HPC \& Big Data Systems, requiring efficient handling of large, continuous time-series datasets and high-throughput data processing. This benchmark was running on NVIDIA V100.& Benchmarks deep learning models for predicting the response of various cancer cell lines to different therapeutic compounds. \\
\hline 
%
%
\rowcolor{gray!20}\multicolumn{7}{|l|}{\textbf{MLCommons AlgoPerf}} \\ \hline
\hline
AlgoPerf &  Criteo 1TB & Click-Through Rate (CTR) Prediction & Large-scale Recommender Systems, Digital Advertising & DLRM-Small (Deep Learning Recommendation Model) &Time-to-Result (Time to reach a target AUC) & Datacenter CPUs/GPUs with high memory bandwidth (HBM) due to massive embedding tables, and highly optimized network I/O. & Stresses memory access and sparse feature embedding computations due to the large, sparse Criteo 1TB dataset. Represents a common commercial workload. \\
\hline 
AlgoPerf &  FastMRI & k-space MRI Reconstruction & Medical Imaging, Healthcare Diagnostics & U-Net (Convolutional Encoder-Decoder) &Time-to-Result (Time to reach a target PSNR / SSIM)&High-Performance GPUs and dedicated AI accelerators, as the model must run with high accuracy and low latency for clinical use.&Focuses on accelerating the image formation process from raw MRI data. U-Net is a standard model for semantic segmentation and image-to-image translation tasks.\\
\hline 
AlgoPerf &  ImageNet & Image Classification & General-purpose Computer Vision & ResNet-50 and Vision Transformer (ViT) variants & Time-to-Result (Time to reach a target Top-1 Accuracy & General-Purpose GPUs (Training/Inference), Edge Devices, and Mobile SoCs, as it is a widely-used test across all compute scales. &The quintessential computer vision workload. Includes two major architecture types (CNN and Transformer) to test algorithm generalizability.\\
\hline
AlgoPerf & LibriSpeech  &Speech Recognition / ASR (Automatic Speech Recognition)& Voice Assistants, Transcription Services & Conformer and DeepSpeech variants & Time-to-Result (Time to reach a target Word Error Rate (WER)) & Datacenter/Cloud GPUs (for large-scale ASR), Edge/Mobile Processors (for on-device assistants). &Tests algorithms on sequential data. Conformer is a hybrid CNN/Transformer architecture common in modern ASR.\\
\hline
AlgoPerf & OGBG  &Graph Property Prediction & Scientific Machine Learning, Drug Discovery, Social Networks & GNN (Graph Neural Network) & Time-to-Result (Time to reach a target ROC-AUC) &Datacenter CPUs/GPUs with high-speed interconnects due to the irregular, sparse nature of graph-structured data. &Uses the Open Graph Benchmark (OGB) dataset. This workload stresses algorithms in domains that rely on non-Euclidean data structures.\\
\hline
AlgoPerf & WMT  & Machine Translation (En-De) &Natural Language Processing (NLP), Global Communication & Transformer (Base Architecture) & Time-to-Result (Time to reach a target BLEU Score) &Datacenter CPUs/GPUs with specialized Tensor Cores for efficient processing of the Transformer's self-attention mechanism.& A standard, large-scale sequence-to-sequence task, famous for being the original domain of the Transformer architecture.\\
\hline
%
%
\rowcolor{gray!20}\multicolumn{7}{|l|}{\textbf{MLCommons AILuminate}} \\ \hline
AILuminate Safety v1.0 & System Under Test (SUT) (Any LLM-based general-purpose chat system)  & Assess Baseline AI Safety and Reliability &Pre-deployment Validation, Regulatory Compliance, Vendor Comparison & LLMs and AI Chat Systems (Text-to-Text), potentially with guardrails/filters &Overall Safety Grade (5-tier scale: Poor to Excellent), Violation Rate (\% of unsafe responses), Per-Hazard Performance &The AI System itself (typically hosted in a Datacenter/Cloud) is the system under test (SUT). The evaluation is performed by a separate, specialized Safety Evaluator Model (often a tuned LLM ensemble). &Assesses safety against 12 Hazard Categories (e.g., Violent Crimes, Hate, Suicide \& Self-Harm). Uses a tuned ensemble of safety evaluator models for grading. Focuses on single-turn, content-only hazards.\\
\hline
AILuminate Jailbreak Benchmark v0.5 & System Under Test (SUT) (Any LLM-based general-purpose chat system)  & Quantify Resilience to Adversarial "Jailbreak" Attacks & AI Security, Robustness Testing, Defense Mechanism Comparison & LLMs (Text-to-Text) and Vision-Language Models (VLMs) (Text+Image-to-Text) & Resilience Gap (Drop in safety performance from baseline to under-attack), Jailbreak Success Rate &The AI System (SUT) is tested in a Datacenter/Cloud environment. The benchmark focuses on the input (adversarial prompts) and the system's subsequent failure rate under attack conditions. & v0.5 is an initial release establishing the framework. It specifically measures the degradation of safety when a system is subjected to prompts designed to bypass its safety filters ("jailbreaks").\\
\hline
\end{longtable}
\end{landscape} 
\twocolumn

}
\clearpage

%
%

\onecolumn
\begin{landscape}

{\tiny
\begin{longtable}
{|p{
0.1\textwidth}|p{
0.1\textwidth}|p{
0.1\textwidth}|p{
0.2\textwidth}|p{
0.2\textwidth}|p{
0.2\textwidth}|p{
0.2\textwidth}|}
\caption{Large Language Model Benchmark Details}
\label{tab:llm_benchmarks_long} \\
\hline
\rowcolor{blue!30}
\textbf{Benchmark Name} & \textbf{Model} & \textbf{Task} & \textbf{Application Domain / Use Case} & \textbf{Model Type / Architecture} & \textbf{Metrics / KPIs} & \textbf{Notes / Description} \\
\hline
\endfirsthead

\rowcolor{blue!30}\multicolumn{7}{c}%
{{\bfseries \tablename\ \thetable{} -- Continued from previous page}} \\
\toprule
\textbf{Benchmark Name} &
\textbf{Model} &
\textbf{Task} &
\textbf{Application Domain / Use Case} &
\textbf{Model Type / Architecture} &
\textbf{Metrics / KPIs} &
\textbf{Notes / Description} \\
\midrule
\endhead

\midrule
\multicolumn{7}{r}{{\footnotesize Continued on next page}} \\
\endfoot

\bottomrule
\endlastfoot
\hline
\rowcolor{gray!20}\multicolumn{7}{|l|}{\textbf{Commercial/Proprietary LLMs (API/Systems)}} \\
\hline
LLM Inference & Claude 3.5 Haiku 20241022 & Generative AI & General Purpose, Light Reasoning & Large Transformer (Proprietary) & TTFT, TPOT, Throughput, MMLU (Quality) & A faster, smaller version in the Claude 3.5 family. \\
\hline
LLM Inference & Claude 3.5 Sonnet 20241022 & Generative AI & Complex Reasoning, Data Processing & Large Transformer (Proprietary) & TTFT, TPOT, Throughput, MMLU (Quality) & Mid-tier model focusing on balance of speed and intelligence. \\
\hline
LLM Inference & Mistral Large 2402 Moderated & Generative AI & Enterprise Chatbots, Content Moderation & MoE/Dense Transformer (Proprietary) & TTFT, TPOT, Throughput, Safety Index & Flagged as moderated; emphasis on safety and reliable output. \\
\hline
LLM Inference & Amazon Nova Lite v1.0 & Generative AI & AWS Services, Embedded Use Cases & Large Transformer (Proprietary) & Latency, Throughput, Cost/Token & Lightweight, cloud-optimized model. \\
\hline
LLM Inference & Gemini 1.5 Pro (API, with option) & Generative AI / Multimodal & Long Context, Multi-Source Reasoning & MoE/Dense Transformer (Proprietary, Multimodal) & TTFT, Throughput, Latency, RAG/Context Recall & Known for its massive context window. \\
\hline
LLM Inference & Gemini 2.0 Flash 001 & Generative AI / Multimodal & High-Speed Chat, Real-time Tasks & Dense Transformer (Proprietary, Multimodal) & p99 Latency, Throughput & Focuses on speed and efficiency for low-latency needs. \\
\hline
LLM Inference & Gemini 2.0 Flash Lite & Generative AI & Edge/Client-Side Inference & Dense Transformer (Proprietary, Small) & Energy Efficiency, Latency & Highly optimized for resource-constrained environments. \\
\hline
LLM Inference & GPT-4o & Generative AI / Multimodal & Real-time Conversation, Vision Integration & Dense Transformer (Proprietary, Multimodal) & TTFT, TPOT, Low-Latency Response & All-in-one model for low-latency multimodal interactions. \\
\hline
LLM Inference & GPT-4o mini & Generative AI & Quick, Cost-Effective Tasks & Dense Transformer (Proprietary, Small) & Cost/Token, Throughput & Optimized for efficiency and scaling simple tasks. \\
\hline
LLM Inference & Minustral 8B 24.10 (API) & Generative AI & General Text Generation & MoE/Dense Transformer (Proprietary) & Latency, Throughput & Represents a competitive, smaller model in a commercial API. \\
\hline
\rowcolor{gray!20}\multicolumn{7}{|l|}{\textbf{Open-Source/Bare Models (Used for Training or Deployment)}} \\
\hline
LLM Inference & Minustral 8B 24.10 Moderation & Generative AI & General Text Generation, Safety Research & MoE/Dense Transformer (Open-weights) & Latency, Safety Compliance & Open-weight version with a focus on safety. \\
\hline
LLM Inference & Gemma 2 9b & Generative AI & Fine-tuning, Edge Deployment & Dense Transformer (Open-weights) & Perplexity, MMLU, Throughput & Smaller model from the Gemma family, good for fine-tuning. \\
\hline
LLM Inference & Phi 3.5 MoE Instruct & Generative AI & Instruction Following, Small Scale Reasoning & MoE (Open-weights, Small) & MMLU, HumanEval (Code) & Instruction-tuned, likely using a small Mixture-of-Experts. \\
\hline
LLM Inference & Phi 4 & Generative AI & Research, Prototyping & Dense Transformer (Open-weights, Small) & Perplexity, BLEU (Generation) & Successor in the Phi family, typically very small. \\
\hline
LLM Inference & Athene V2 Chat Hf & Generative AI & Open Chatbot Deployment & Dense Transformer (Open-weights, Fine-tuned) & TTFT, TPOT, Chat Metrics & An instruction-tuned model from the Hugging Face ecosystem. \\
\hline
LLM Inference & Aya Expanse 8B Hf & Generative AI & Multilingual Tasks, Text Translation & Dense Transformer (Open-weights) & BLEU (Translation), Accuracy & Focused on broad language coverage. \\
\hline
LLM Inference & Cohere C4Ai Command A 03 2025 Hf & Generative AI & Enterprise RAG, Instruction Following & Dense Transformer (Open-weights) & Contextual Recall, RAG Latency & Cohere model variant used in the Hugging Face ecosystem. \\
\hline
LLM Inference & Llama 3.1 405B Instruct & Generative AI & State-of-the-Art Reasoning, Long Context & Dense Transformer (Open-weights) & TTFT, Throughput, MMLU & An extremely large, cutting-edge open-weight model (used in MLPerf). \\
\hline
LLM Inference & Llama 3.1 8b Instruct FP8 & Generative AI & Edge/Quantized Deployment & Dense Transformer (Quantized) & Inference Accuracy, Memory Footprint & Highly optimized for efficient computation using 8-bit precision. \\
\hline
LLM Inference & Llama 3 1 Tulu 3 8B Hf & Generative AI & General Chat, Fine-tuning Research & Dense Transformer (Open-weights, Fine-tuned) & Alpaca Eval, Human Preference & A variant of Llama tuned for instruction following. \\
\hline
LLM Inference & Mistralai Mistral Large 2402 & Generative AI & Complex Reasoning, RAG & MoE/Dense Transformer (Open-weights) & TTFT, TPOT, MMLU & Open-weight version of Mistral's flagship model. \\
\hline
LLM Inference & Olmo 2 0325 32b Instruct & Generative AI & Research, Reproducible AI & Dense Transformer (Open-weights) & Perplexity, Training Speed & High-parameter model focused on openness and research. \\
\hline
LLM Inference & Olmo 2 1124 13B Instruct Hf & Generative AI & Instruction Following, General Chat & Dense Transformer (Open-weights) & TTFT, Throughput & Smaller, instruction-tuned version of the Olmo family. \\
\hline
LLM Inference & Phi 3.5 Mini Instruct & Generative AI & Mobile/Edge Inference, Simple Tasks & Dense Transformer (Open-weights, Small) & Latency, MMLU & Ultra-small model optimized for fast responses. \\
\hline
LLM Inference & Qwen1 5 110B Chat Hf & Generative AI & Multi-Language Chat, High Accuracy & Dense Transformer (Open-weights) & C-Eval, MMLU, Throughput & High-parameter model known for strong Chinese/general performance. \\
\hline
LLM Inference & Yi 1 5 34B Chat Hf & Generative AI & General Purpose, Instruction Following & Dense Transformer (Open-weights) & MMLU, C-Eval, Latency & Mid-to-large size model focusing on quality chat performance. \\
\hline
LLM Inference & Ai21Labs Ai21 Jamba Large 1.5 Azure & Generative AI & Cloud Deployment, Enterprise Apps & Hybrid MoE/Dense Transformer & Throughput, Latency & A large model known for its hybrid architecture, deployed via Azure. \\
\hline
LLM Inference & Google Gemma 3 27B It Hf Nebius & Generative AI & Cloud Deployment, Fine-tuning & Dense Transformer (Open-weights, Fine-tuned) & TTFT, TPOT, Cloud Efficiency & Gemma model deployed on the Nebius cloud platform. \\
\hline
LLM Inference & Llama 3.3 70B Instruct Turbo Together & Generative AI & Fast, High-Quality Instruction Following & Dense Transformer (Open-weights) & Latency, Throughput, Cost & A large model optimized for speed via the Together API. \\
\hline
LLM Inference & Mistral Large 24.11 & Generative AI & Enterprise AI, High Performance & MoE/Dense Transformer (Open-weights) & Throughput, MMLU, Reasoning & A very recent high-performance model. \\
\hline
LLM Inference & Qwq 32B Hf & Generative AI & General Purpose, Instruction Following & Dense Transformer (Open-weights) & Latency, Throughput & A mid-sized model in the open-weight ecosystem. \\
\hline
LLM Inference & OLMo 7b 0724 Instruct & Generative AI & Research, Instruction Following & Dense Transformer (Open-weights) & Perplexity, Speed & Smaller, instruction-tuned model for general tasks. \\

\end{longtable}
}
    
\end{landscape}
\twocolumn

\clearpage

%% file: ontology/selected-new.tex
\newlength{\ratingswidth}
\setlength{\ratingswidth}{0.11\textwidth}

\newlength{\imageratingswidth}
\setlength{\imageratingswidth}{0.06\textwidth}

\setlength\LTpre{0pt}
\setlength\LTpost{0pt}
\setlength{\parskip}{0pt}
\setlength{\parindent}{0pt}
\renewcommand{\arraystretch}{0.90}
\setlength{\tabcolsep}{2pt}
\newcolumntype{M}[1]{>{\centering\arraybackslash}m{#1}}

\onecolumn
\begin{landscape}
{\footnotesize
\begin{longtable}{
|M{0.08\textwidth}
|M{0.30\textwidth}
|M{0.20\textwidth}
|M{0.20\textwidth}
|M{0.30\textwidth}
|M{0.11\textwidth}
|}
\caption{Ontology Table for Selected AI Science Benchmarks. \\ 
{\tiny(For detailed view of the Radar Charts, see \cite{www-las-mlcommons-benchmark-collection}.)} } 
\label{tab:ontology} 
\\ \hline
\rowcolor{blue!30} \textbf{Ratings} & \textbf{Name} & \textbf{Domain} & \textbf{Models} & \textbf{Metrics} & \textbf{Citation} \\ \hline
\endfirsthead
\caption{Ontology Table for Selected AI Science Benchmarks (cont.). }\\
\hline
\rowcolor{blue!30} \textbf{Ratings} & \textbf{Name} & \textbf{Domain} & \textbf{Models} & \textbf{Metrics} & \textbf{Citation}  \\ \hline
\endhead
\hline
\multicolumn{6}{r}{Continued on next page} \\ \hline
\endfoot
\hline
\endlastfoot
\includegraphics[width=0.05\textwidth]{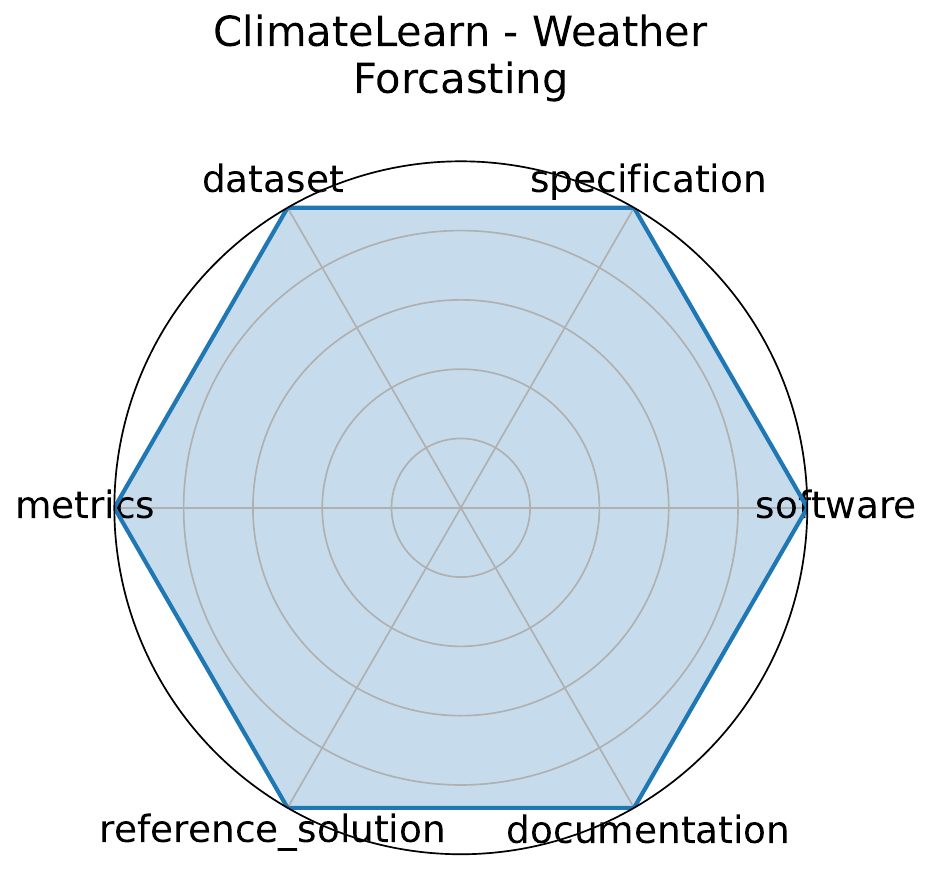} & ClimateLearn - Weather Forcasting & Climate \& Earth Science & CNN baselines, ResNet variants & RMSE, Anomaly correlation & \cite{nguyen2023climatelearnbenchmarkingmachinelearning} \\ \hline
\includegraphics[width=0.05\textwidth]{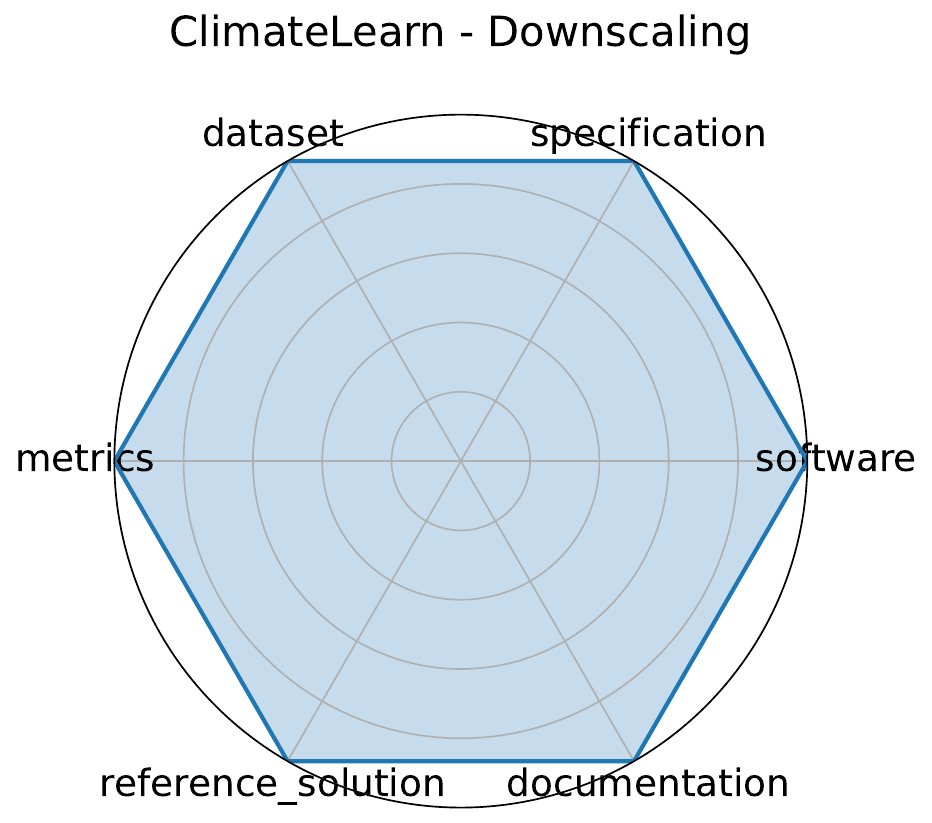} & ClimateLearn - Downscaling & Climate \& Earth Science & CNN baselines, ResNet variants & RMSE, Anomaly correlation & \cite{nguyen2023climatelearnbenchmarkingmachinelearning} \\ \hline
\includegraphics[width=0.05\textwidth]{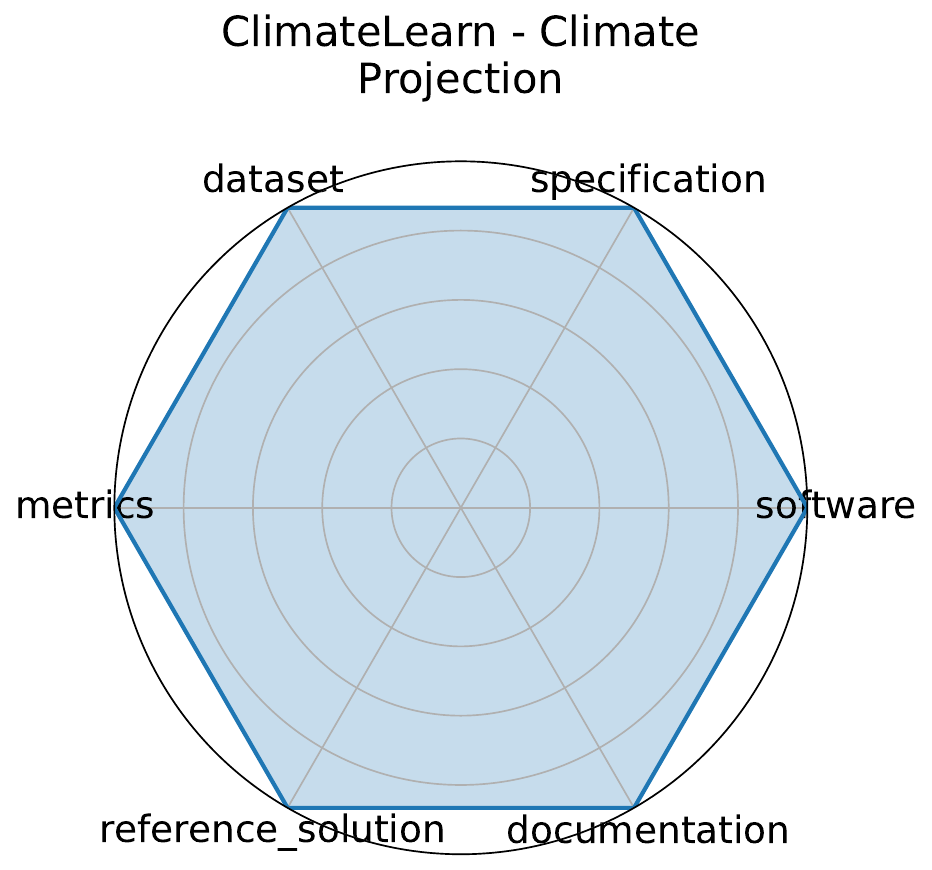} & ClimateLearn - Climate Projection & Climate \& Earth Science & CNN baselines, ResNet variants & RMSE, Anomaly correlation & \cite{nguyen2023climatelearnbenchmarkingmachinelearning} \\ \hline
\includegraphics[width=0.05\textwidth]{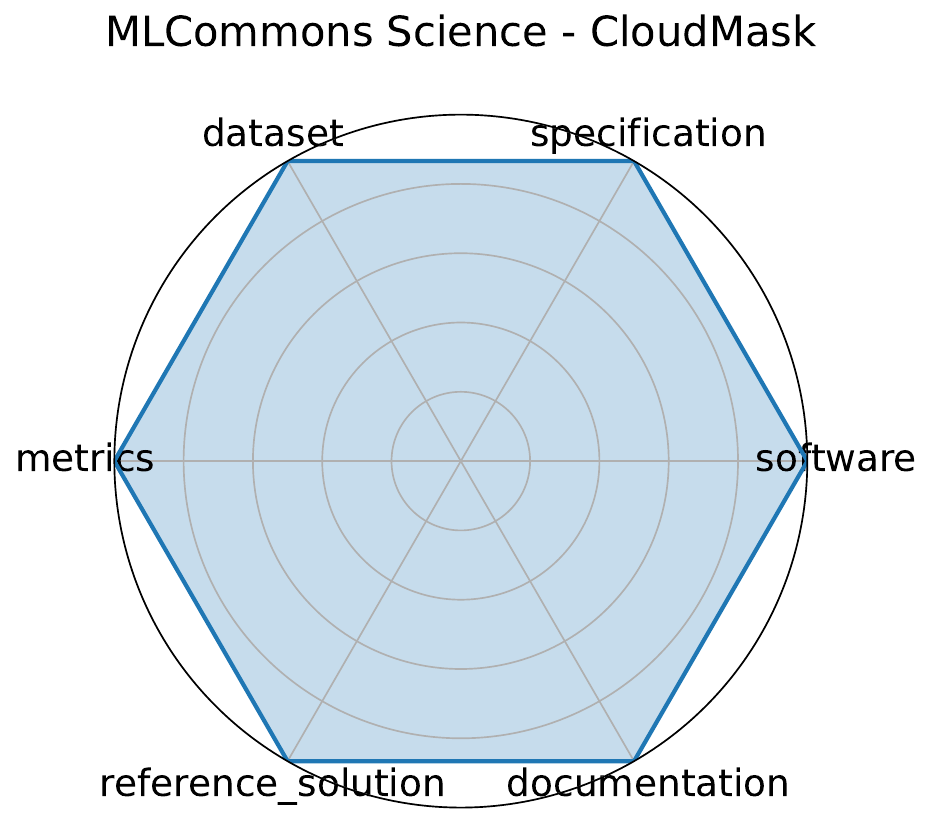} & MLCommons Science - CloudMask & Climate \& Earth Science & CNN, GNN, Transformer & MAE, Accuracy, Speedup vs simulation & \cite{10.1007/978-3-031-23220-6_4} \\ \hline
\includegraphics[width=0.05\textwidth]{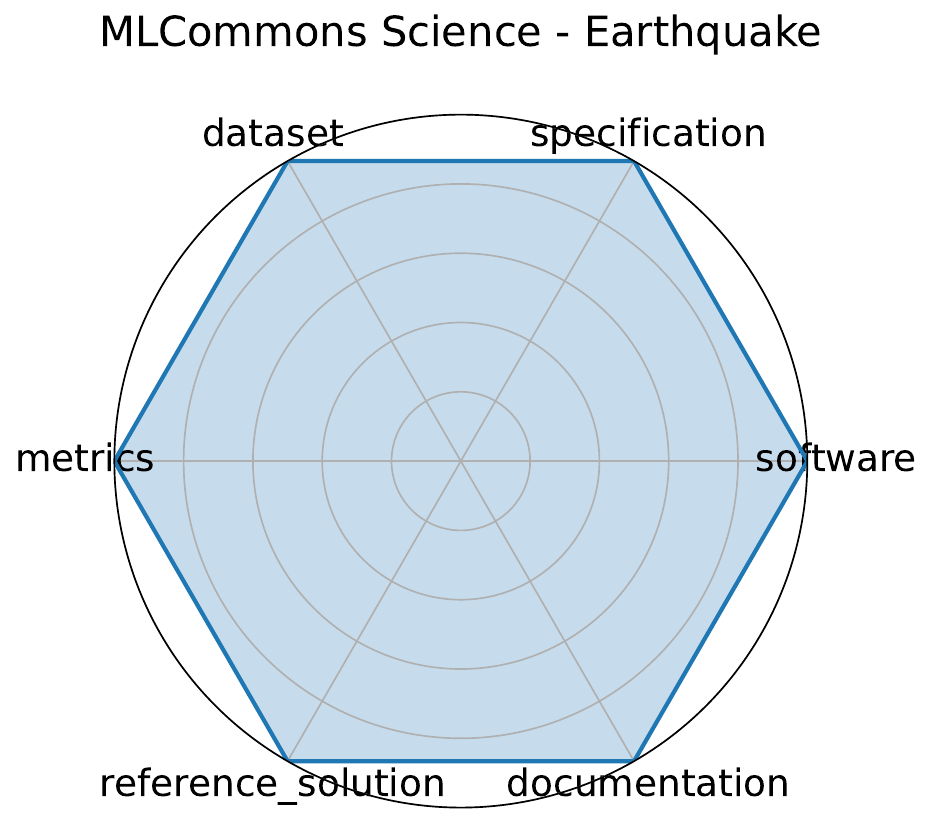} & MLCommons Science - Earthquake & Climate \& Earth Science & CNN, GNN, Transformer & MAE, Accuracy, Speedup vs simulation & \cite{10.1007/978-3-031-23220-6_4} \\ \hline
\includegraphics[width=0.05\textwidth]{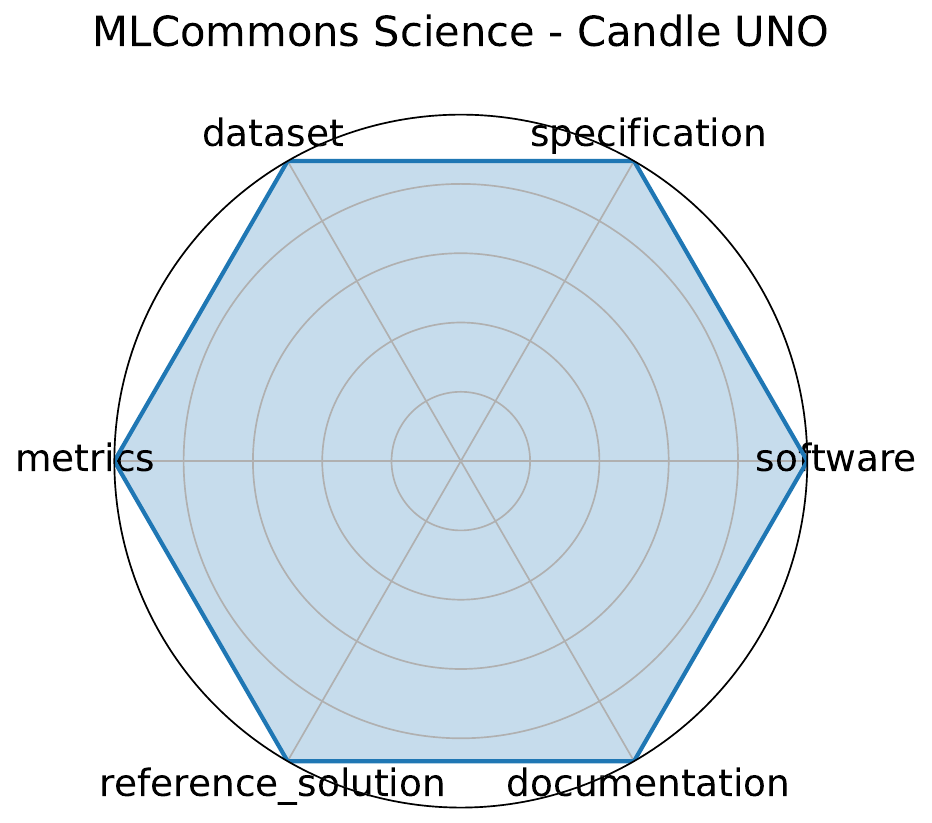} & MLCommons Science - Candle UNO & Biology \& Medicine & CNN, GNN, Transformer & MAE, Accuracy, Speedup vs simulation & \cite{10.1007/978-3-031-23220-6_4} \\ \hline
\includegraphics[width=0.05\textwidth]{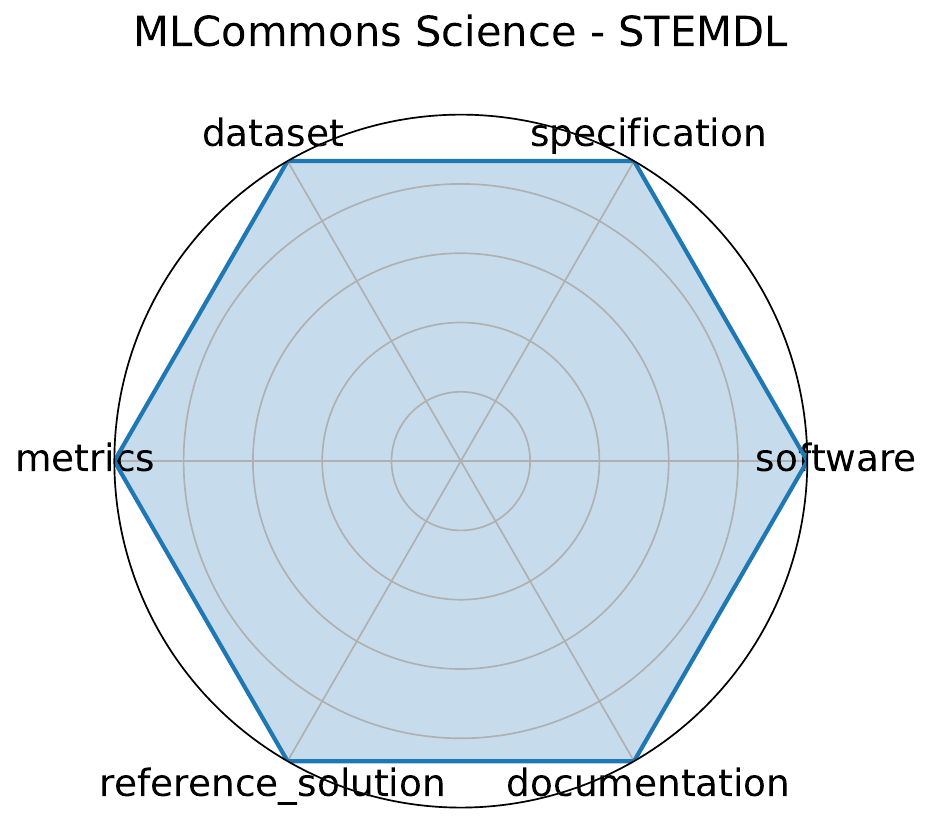} & MLCommons Science - STEMDL & Materials Science & CNN, GNN, Transformer & MAE, Accuracy, Speedup vs simulation & \cite{10.1007/978-3-031-23220-6_4} \\ \hline
\includegraphics[width=0.05\textwidth]{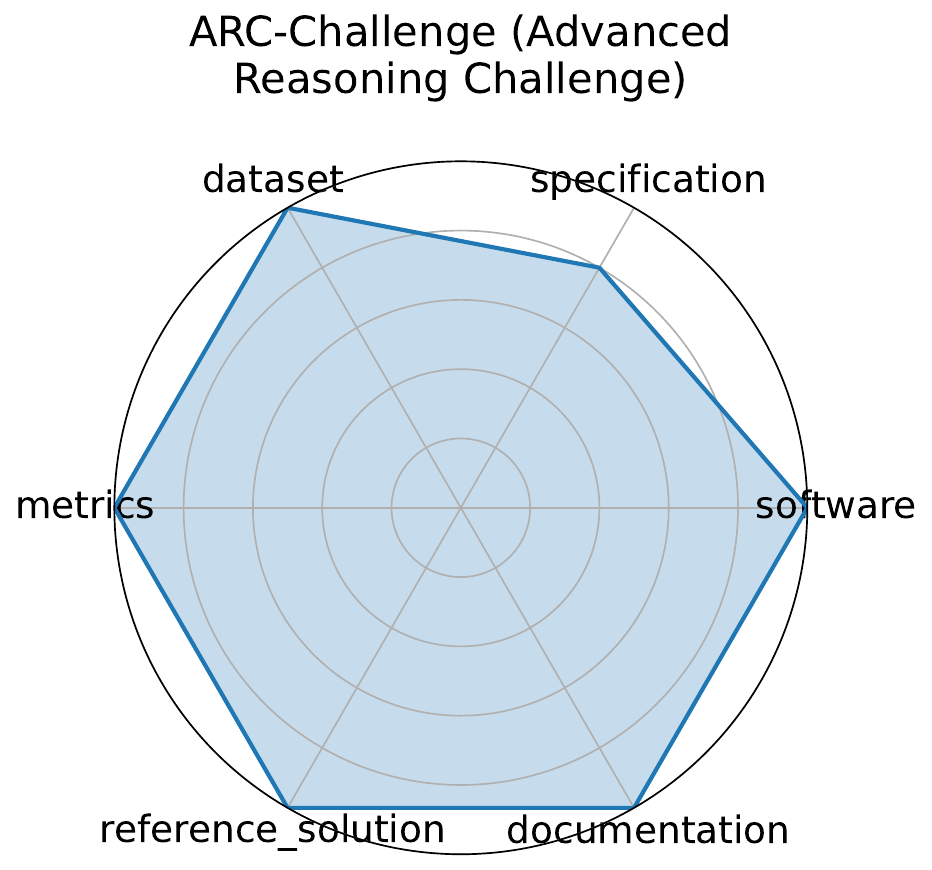} & ARC-Challenge (Advanced Reasoning Challenge) & Computational Science \& AI & GPT-4, Claude & Accuracy & \cite{allenai:arc} \\ \hline
\includegraphics[width=0.05\textwidth]{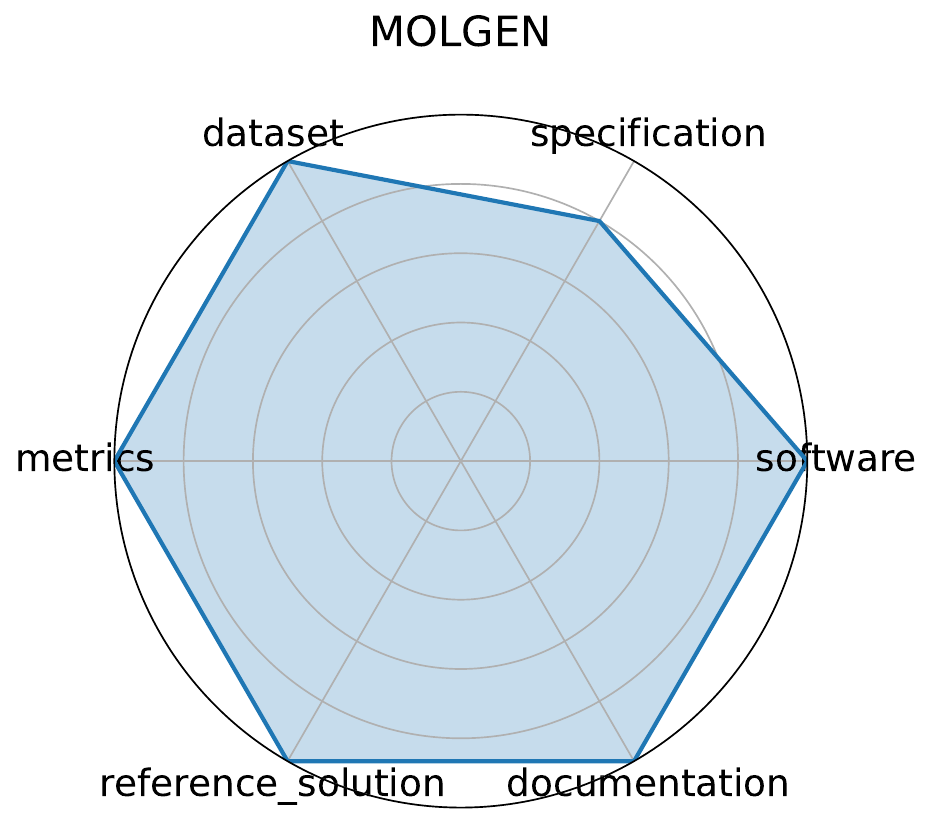} & MOLGEN & Chemistry & MolGen & Validity\%, Novelty\%, QED, Docking score, penalized logP & \cite{fang2024domainagnosticmoleculargenerationchemical} \\ \hline
\includegraphics[width=0.05\textwidth]{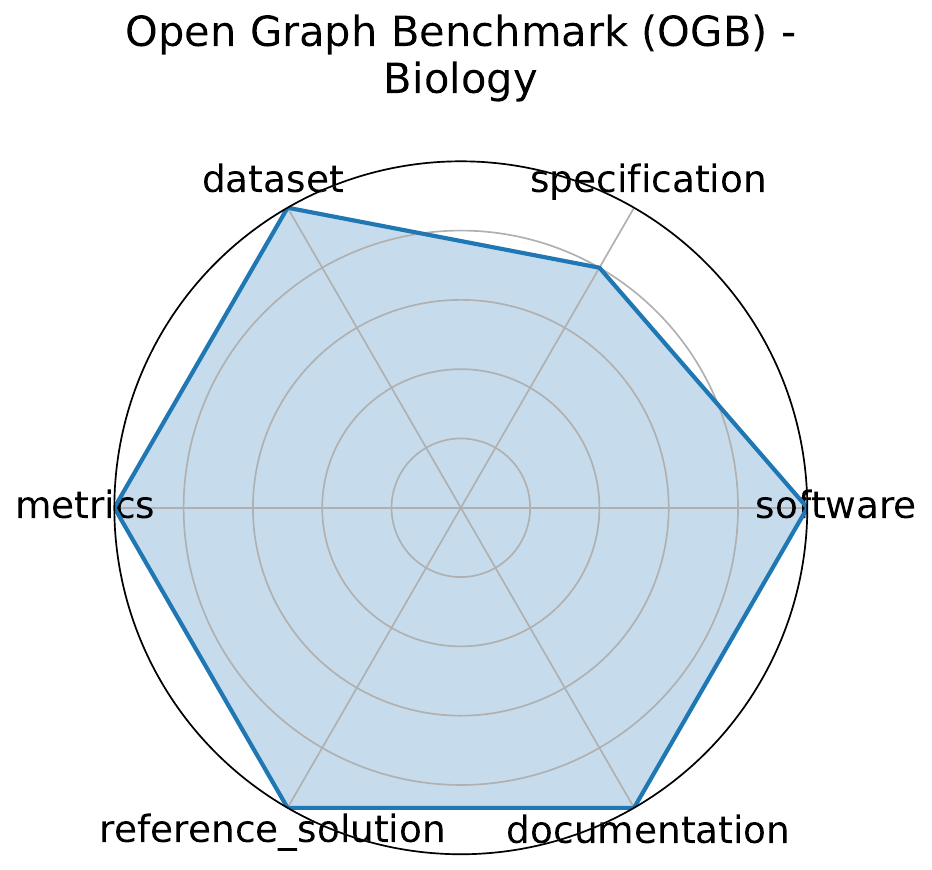} & Open Graph Benchmark (OGB) - Biology & Biology \& Medicine & GCN, GraphSAGE, GAT & Accuracy, ROC-AUC & \cite{hu2021opengraphbenchmarkdatasets} \\ \hline
\includegraphics[width=0.05\textwidth]{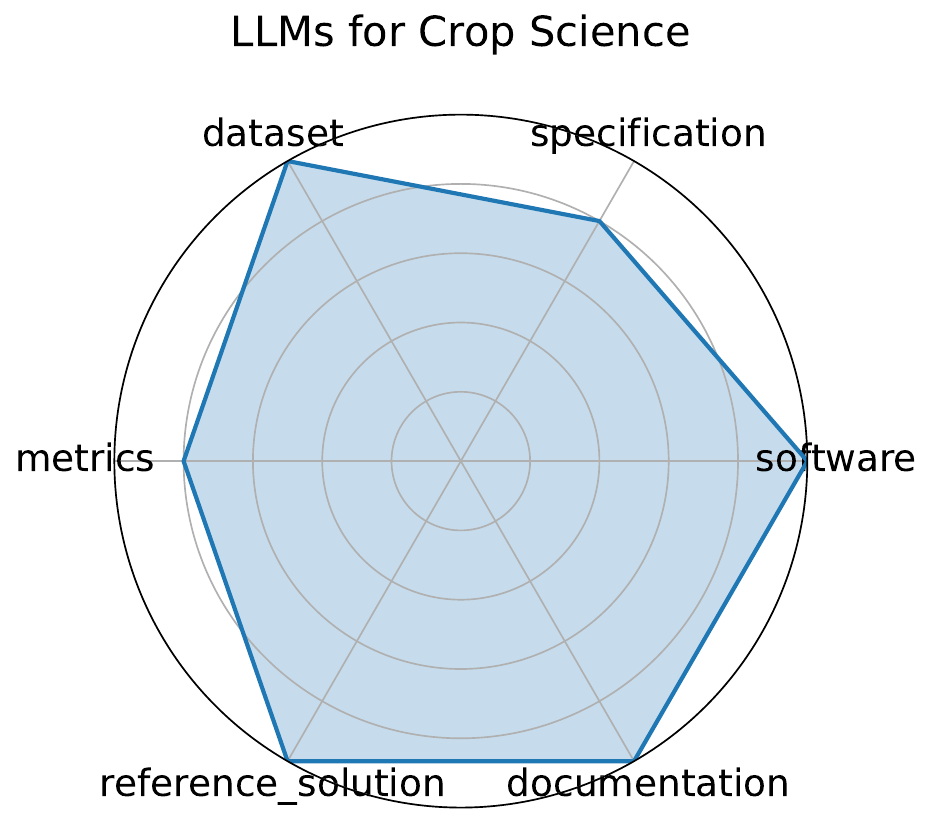} & LLMs for Crop Science & Climate \& Earth Science & GPT-3.5, GPT-4, Claude-3-opus, Qwen-max, LLama3-8B, InternLM2-7B, Qwen1.5-7B & Accuracy, F1 score & \cite{zhang2024empowering} \\ \hline
\includegraphics[width=0.05\textwidth]{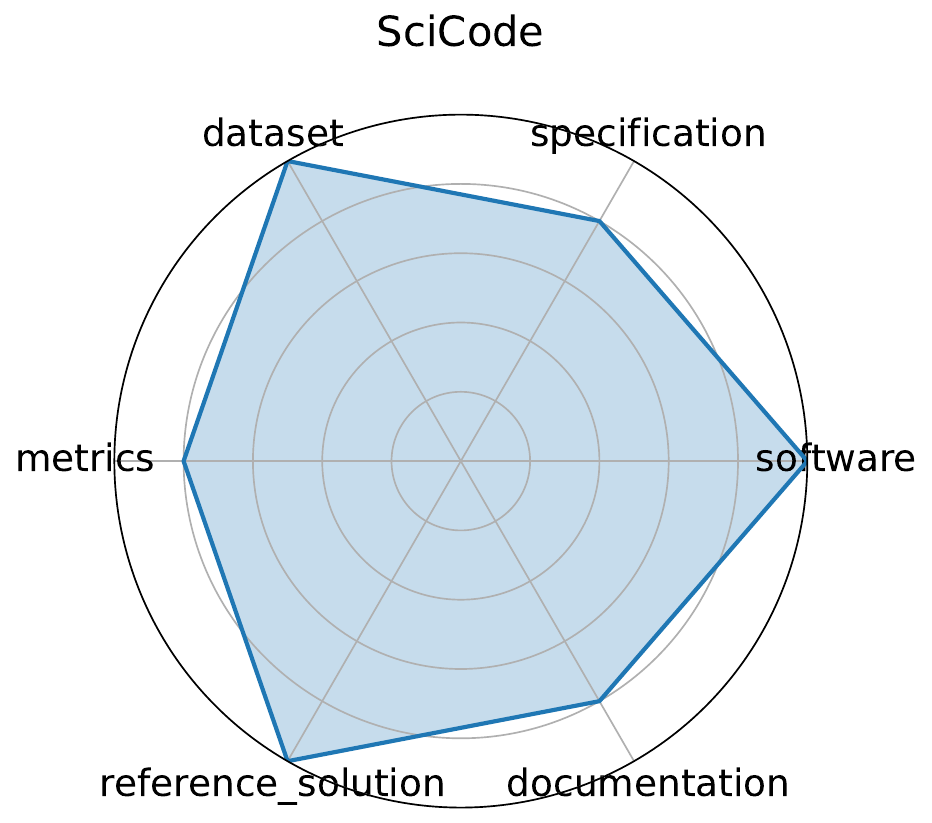} & SciCode & Computational Science \& AI & Claude3.5-Sonnet & Solve rate (\%) & \cite{tian2024scicoderesearchcodingbenchmark} \\ \hline
\includegraphics[width=0.05\textwidth]{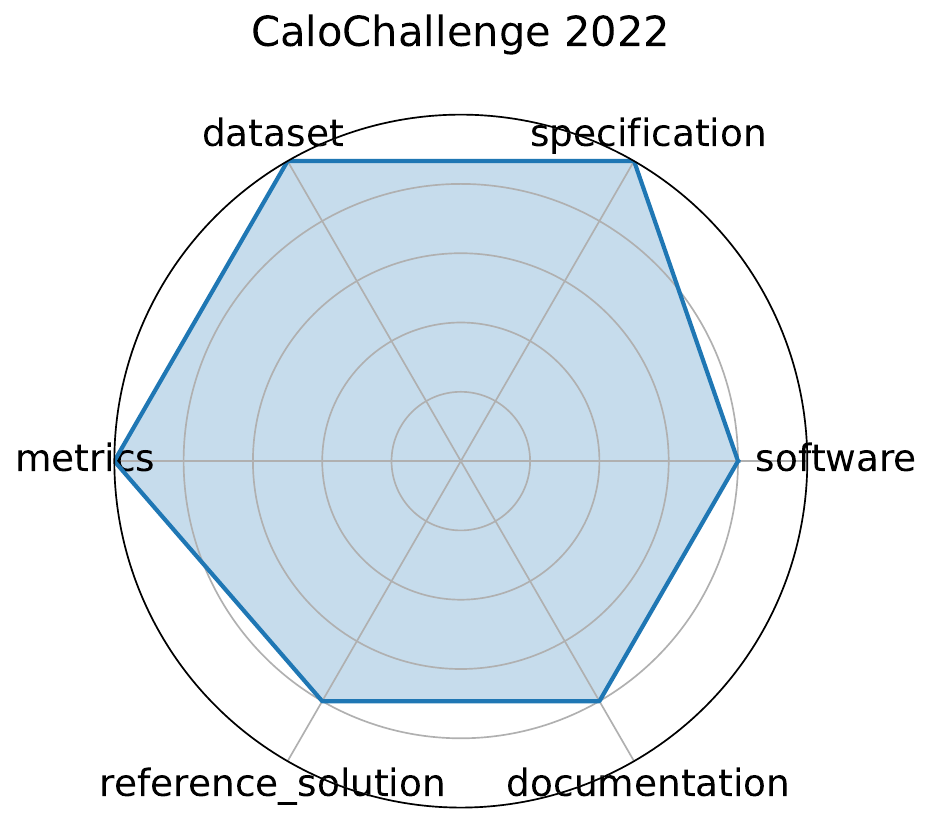} & CaloChallenge 2022 & High Energy Physics & VAE variants, GAN variants, Normalizing flows, Diffusion models & Histogram similarity, Classifier AUC, Generation latency & \cite{krause2024calochallenge2022communitychallenge} \\ \hline
\includegraphics[width=0.05\textwidth]{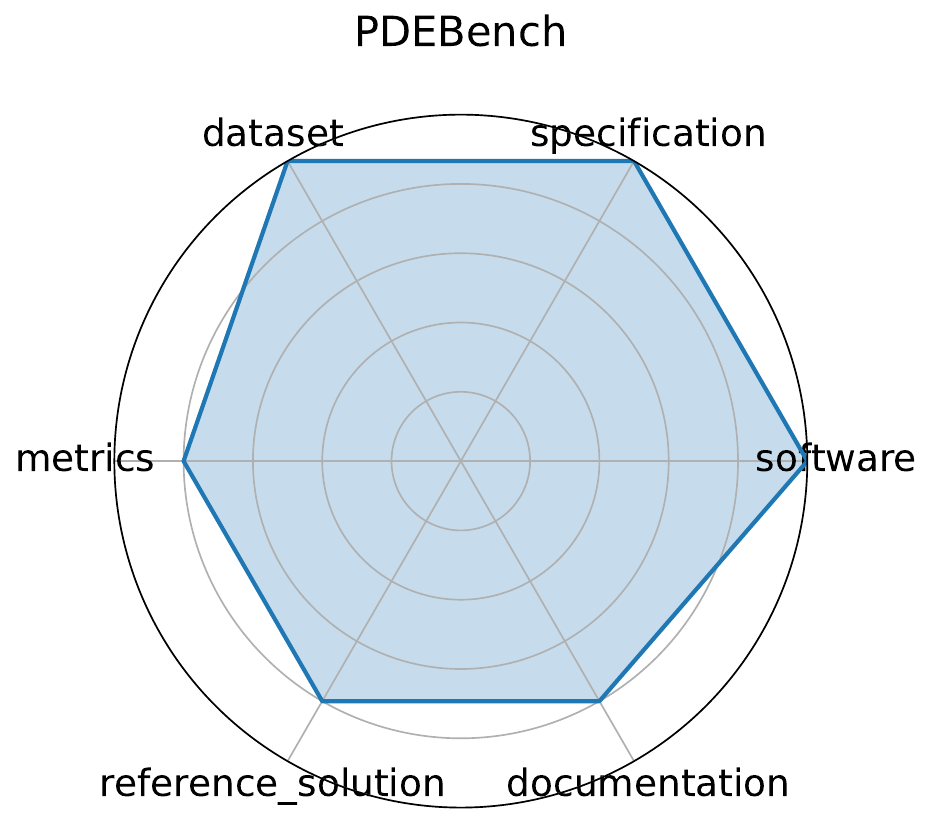} & PDEBench & Computational Science \& AI, Climate \& Earth Science, Mathematics & FNO, U-Net, PINN, Gradient-Based inverse methods & RMSE, boundary RMSE, Fourier RMSE & \cite{takamoto2024pdebenchextensivebenchmarkscientific} \\ \hline
\includegraphics[width=0.05\textwidth]{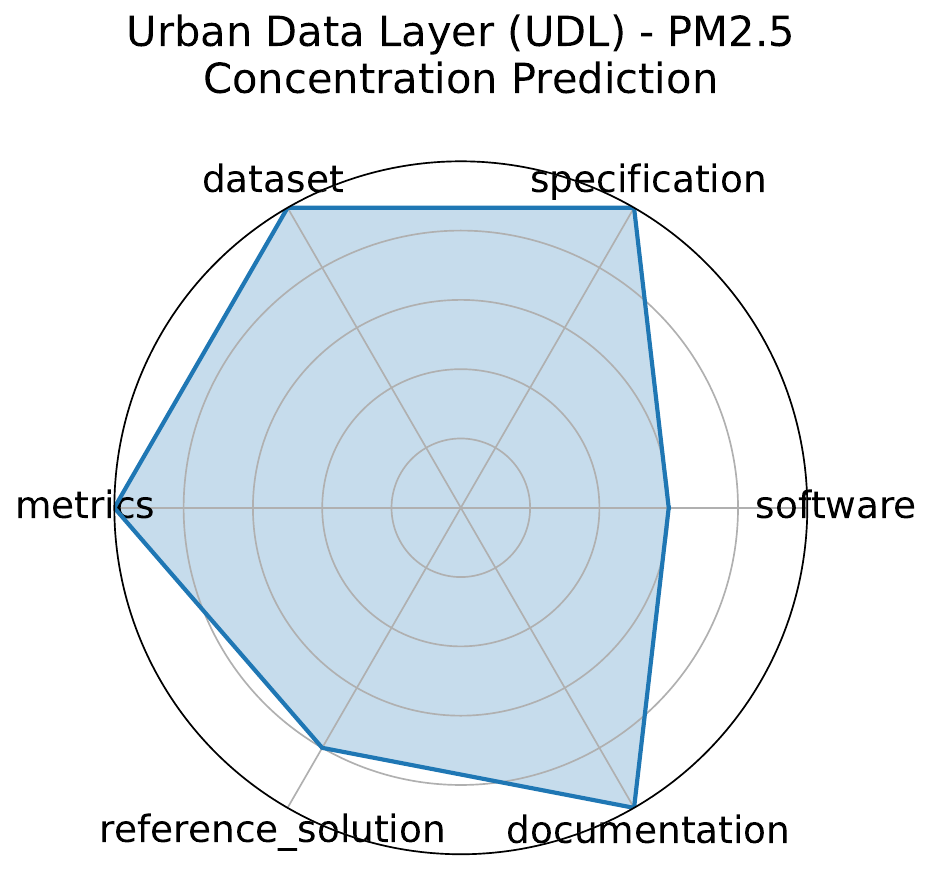} & Urban Data Layer (UDL) - PM2.5 Concentration Prediction & Climate \& Earth Science & Baseline regression/classification pipelines & Task-specific accuracy or RMSE & \cite{neurips2024_0db7f135} \\ \hline
\includegraphics[width=0.05\textwidth]{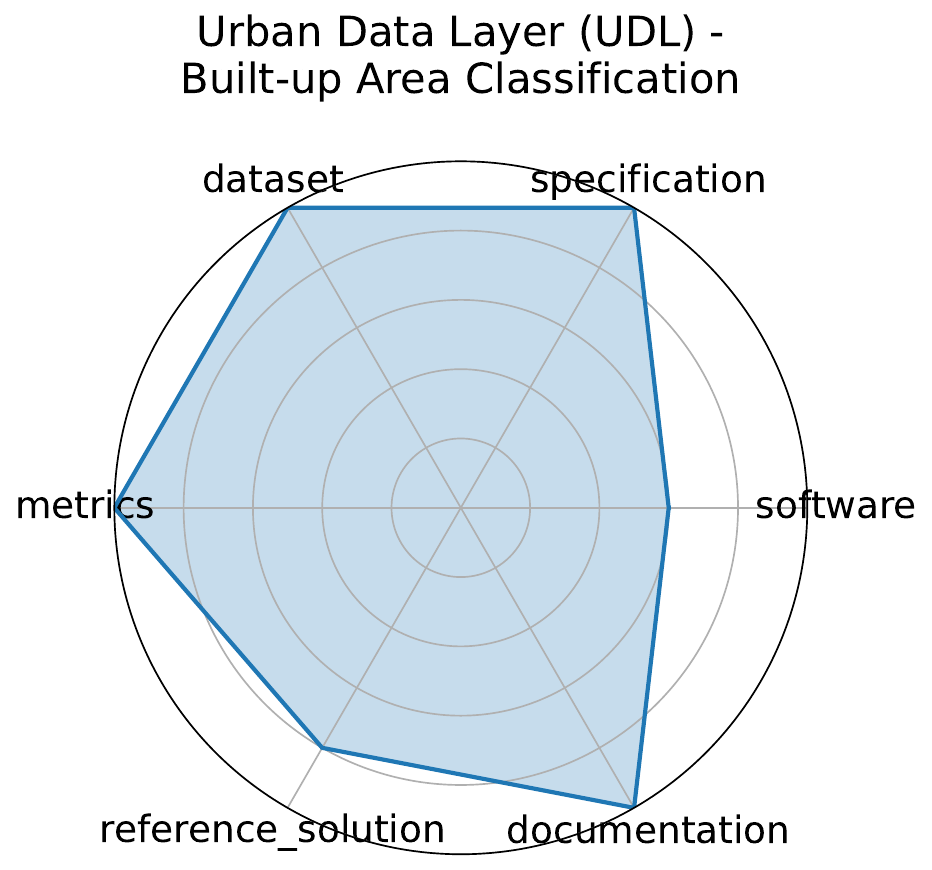} & Urban Data Layer (UDL) - Built-up Area Classification & Climate \& Earth Science & Baseline regression/classification pipelines & Task-specific accuracy or RMSE & \cite{neurips2024_0db7f135} \\ \hline
\includegraphics[width=0.05\textwidth]{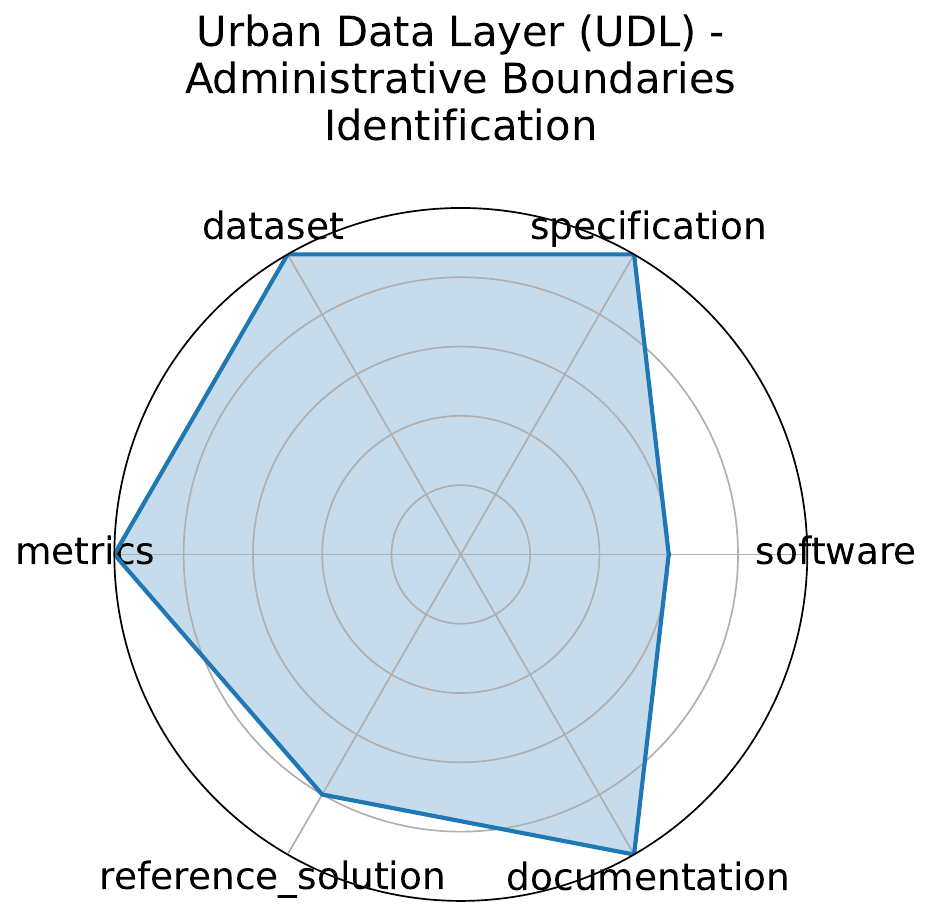} & Urban Data Layer (UDL) - Administrative Boundaries Identification & Climate \& Earth Science & Baseline regression/classification pipelines & Task-specific accuracy or RMSE & \cite{neurips2024_0db7f135} \\ \hline
\includegraphics[width=0.05\textwidth]{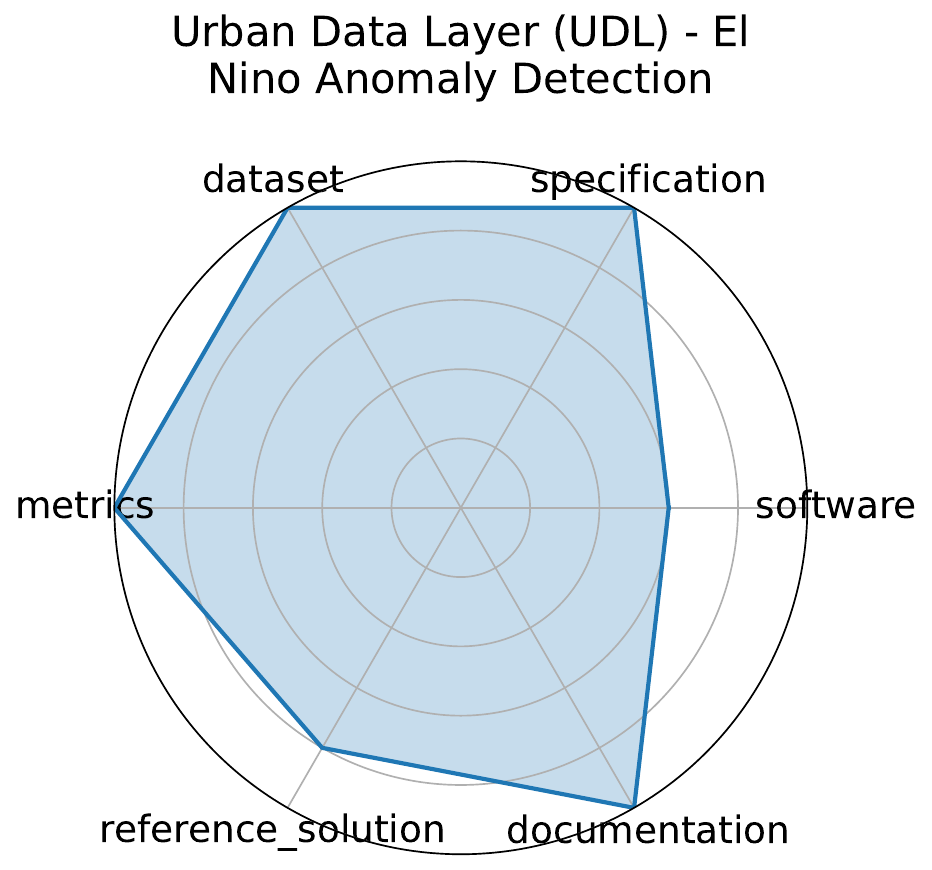} & Urban Data Layer (UDL) - El Nino Anomaly Detection & Climate \& Earth Science & Baseline regression/classification pipelines & Task-specific accuracy or RMSE & \cite{neurips2024_0db7f135} \\ \hline
\includegraphics[width=0.05\textwidth]{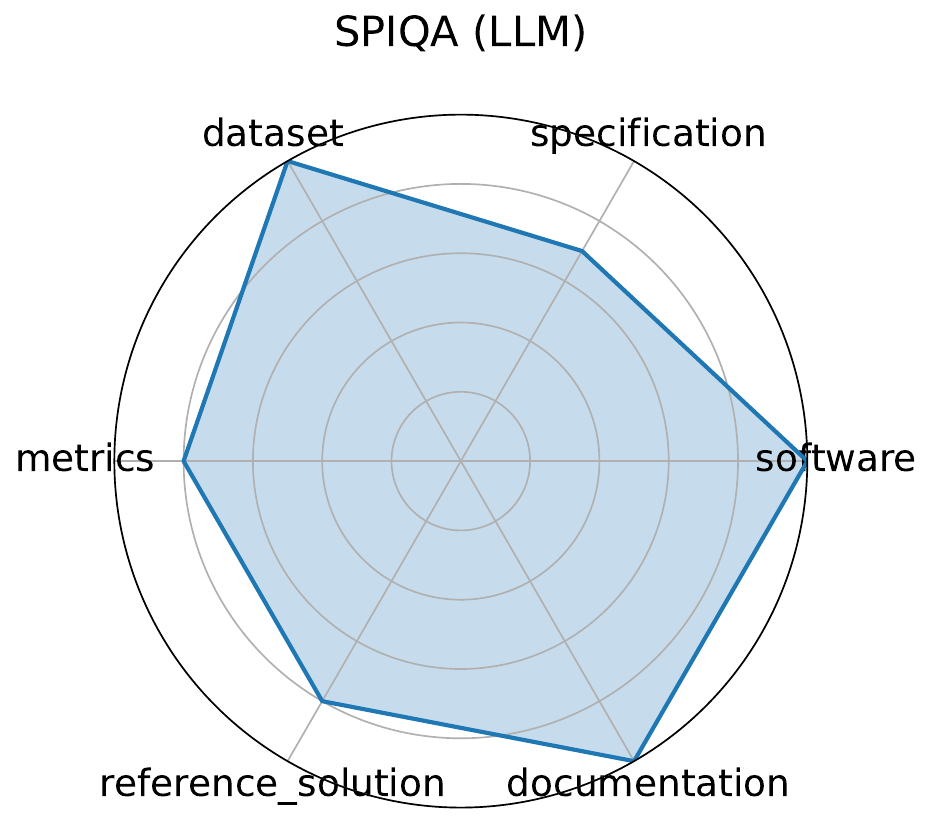} & SPIQA (LLM) & Computational Science \& AI & LLaVA, MiniGPT-4, Owl-LLM adapter variants & Accuracy, F1 score & \cite{pramanick2025spiqadatasetmultimodalquestion} \\ \hline
\includegraphics[width=0.05\textwidth]{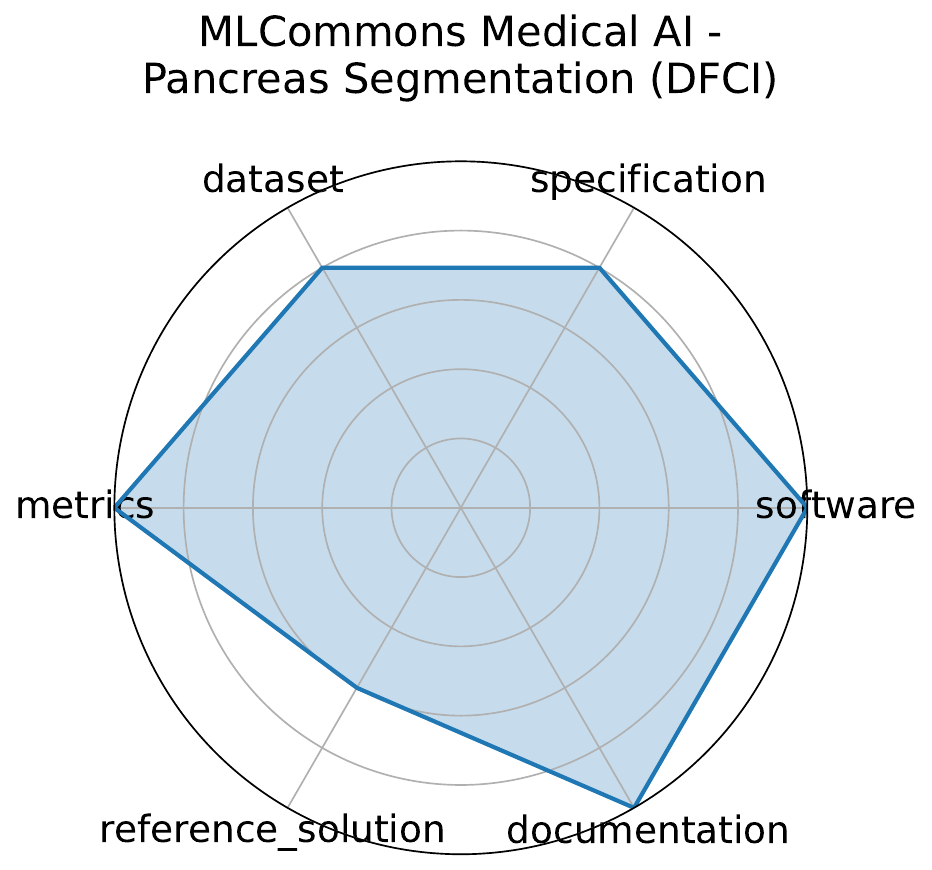} & MLCommons Medical AI - Pancreas Segmentation (DFCI) & Biology \& Medicine & MedPerf-validated CNNs, GaNDLF workflows & ROC AUC, Accuracy, Fairness metrics & \cite{karargyris2023federated} \\ \hline
\includegraphics[width=0.05\textwidth]{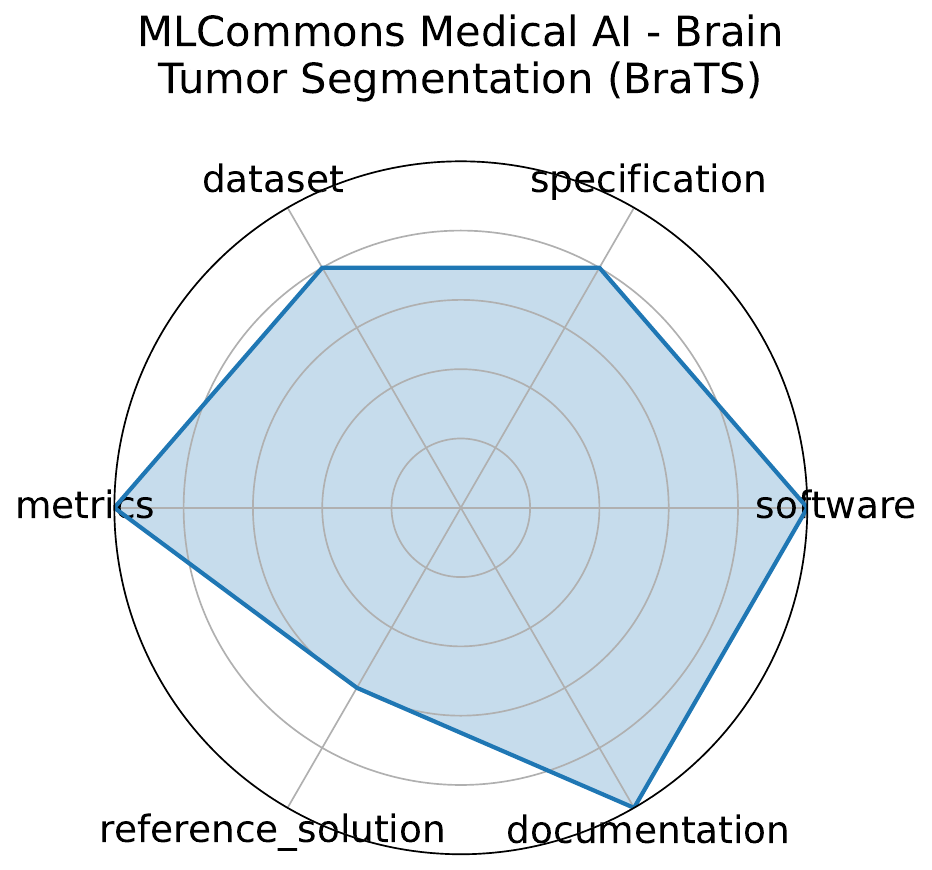} & MLCommons Medical AI - Brain Tumor Segmentation (BraTS) & Biology \& Medicine & MedPerf-validated CNNs, GaNDLF workflows & ROC AUC, Accuracy, Fairness metrics & \cite{karargyris2023federated} \\ \hline
\includegraphics[width=0.05\textwidth]{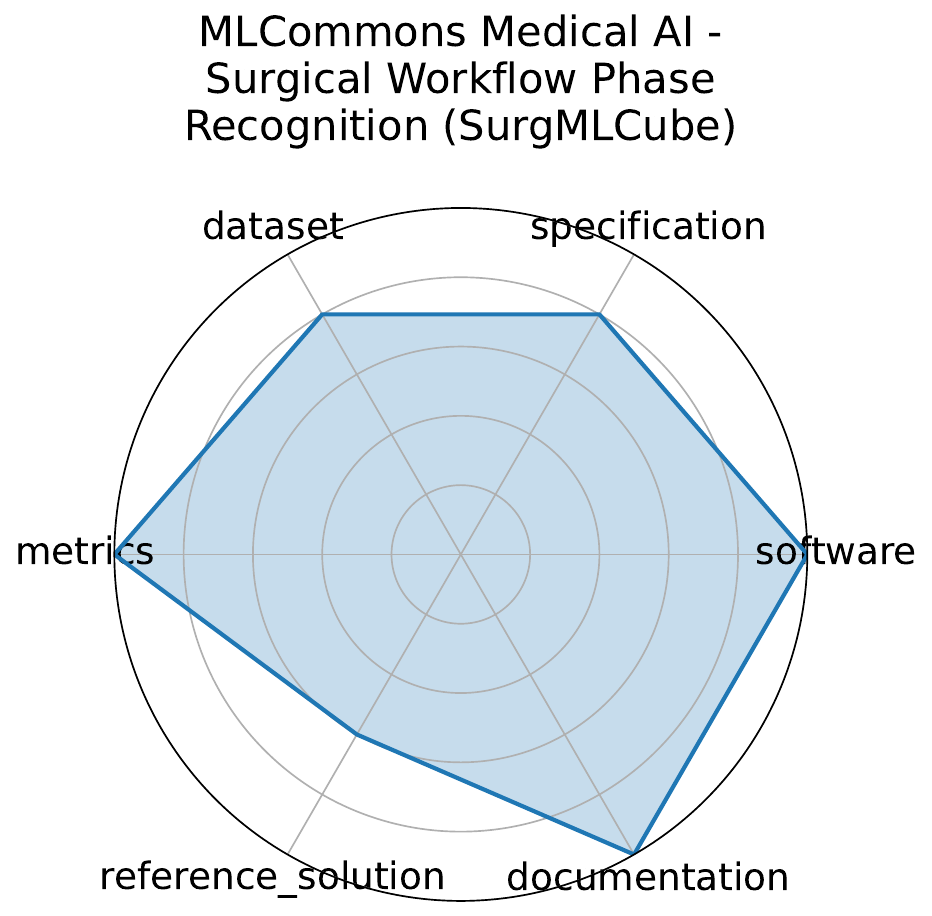} & MLCommons Medical AI -  Surgical Workflow Phase Recognition (SurgMLCube) & Biology \& Medicine & MedPerf-validated CNNs, GaNDLF workflows & ROC AUC, Accuracy, Fairness metrics & \cite{karargyris2023federated} \\ \hline
\includegraphics[width=0.05\textwidth]{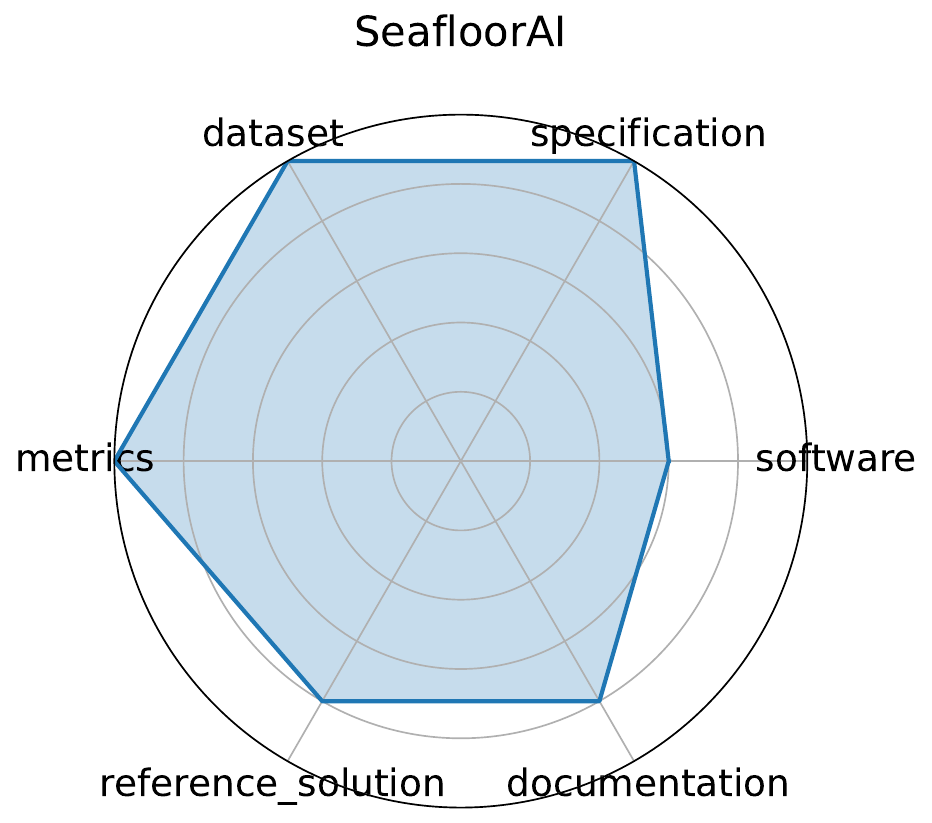} & SeafloorAI & Climate \& Earth Science & SegFormer, ViLT-style multimodal models & Segmentation pixel accuracy, QA accuracy & \cite{nguyen2024seafloor} \\ \hline
\includegraphics[width=0.05\textwidth]{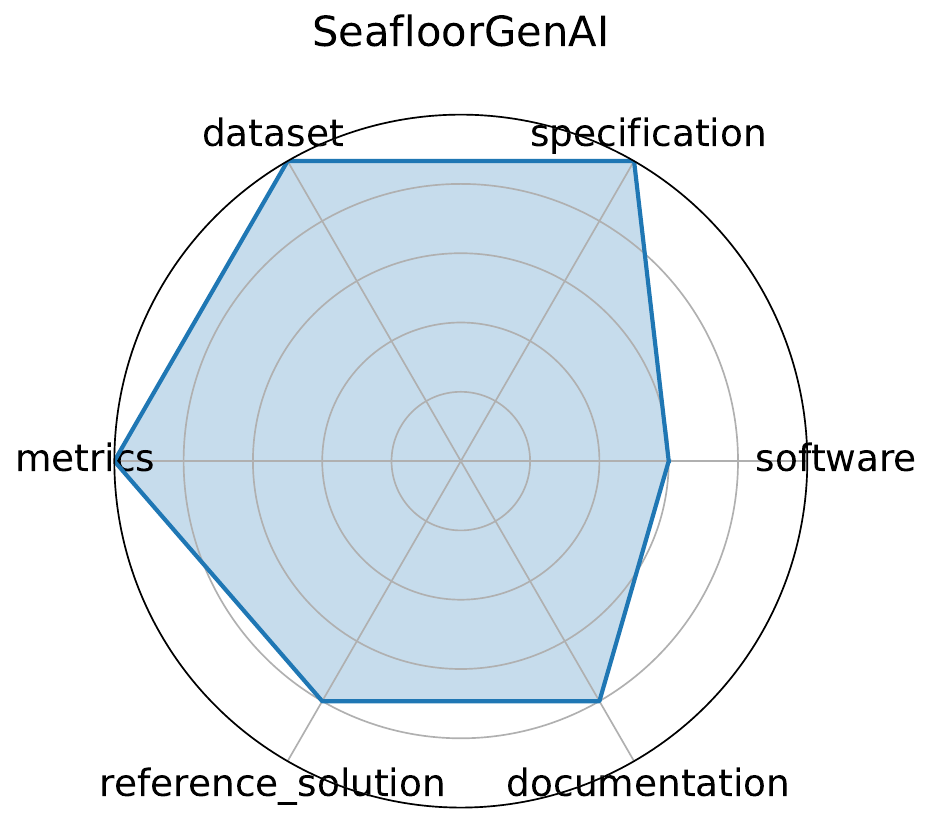} & SeafloorGenAI & Climate \& Earth Science & SegFormer, ViLT-style multimodal models & Segmentation pixel accuracy, QA accuracy & \cite{nguyen2024seafloor} \\ \hline
\includegraphics[width=0.05\textwidth]{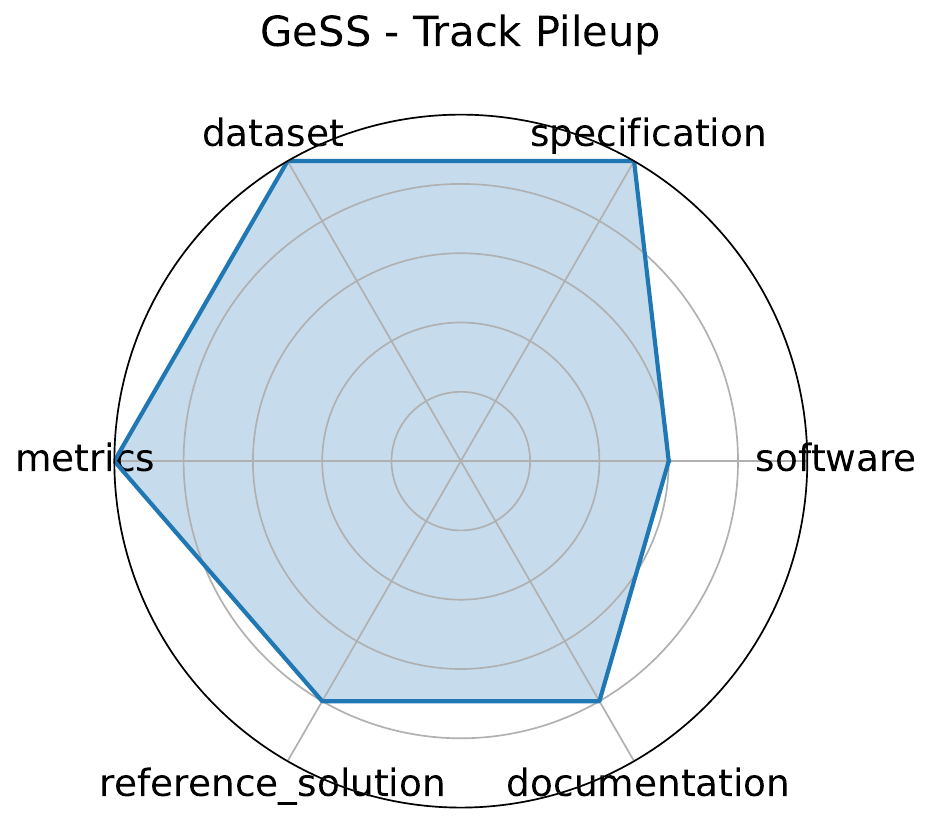} & GeSS - Track Pileup & High Energy Physics & GCN, EGNN, DimeNet++ & Accuracy, RMSE, OOD robustness delta & \cite{neurips2024_a8063075} \\ \hline
\includegraphics[width=0.05\textwidth]{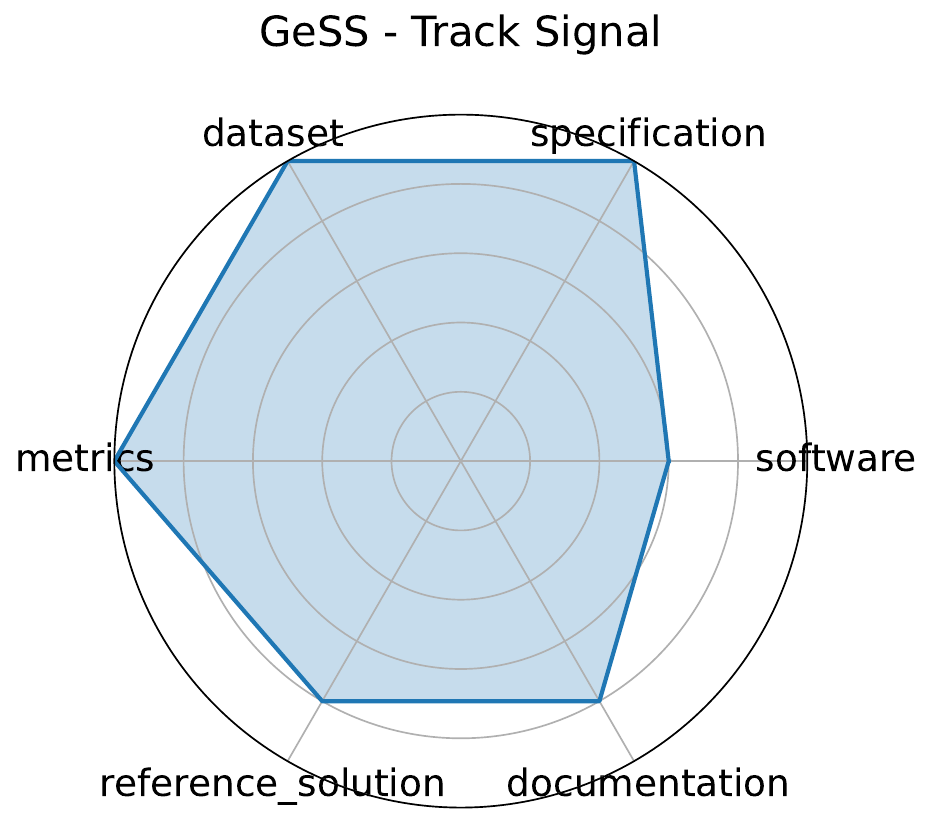} & GeSS - Track Signal & High Energy Physics & GCN, EGNN, DimeNet++ & Accuracy, RMSE, OOD robustness delta & \cite{neurips2024_a8063075} \\ \hline
\includegraphics[width=0.05\textwidth]{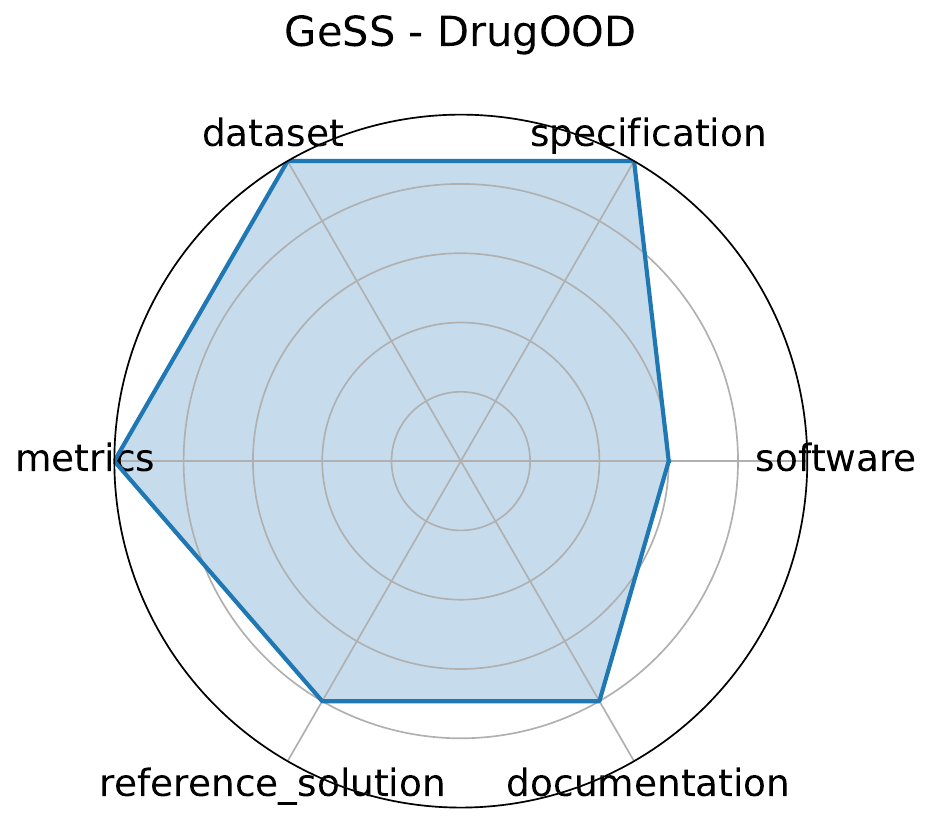} & GeSS - DrugOOD & Biology \& Medicine & GCN, EGNN, DimeNet++ & Accuracy, RMSE, OOD robustness delta & \cite{neurips2024_a8063075} \\ \hline
\includegraphics[width=0.05\textwidth]{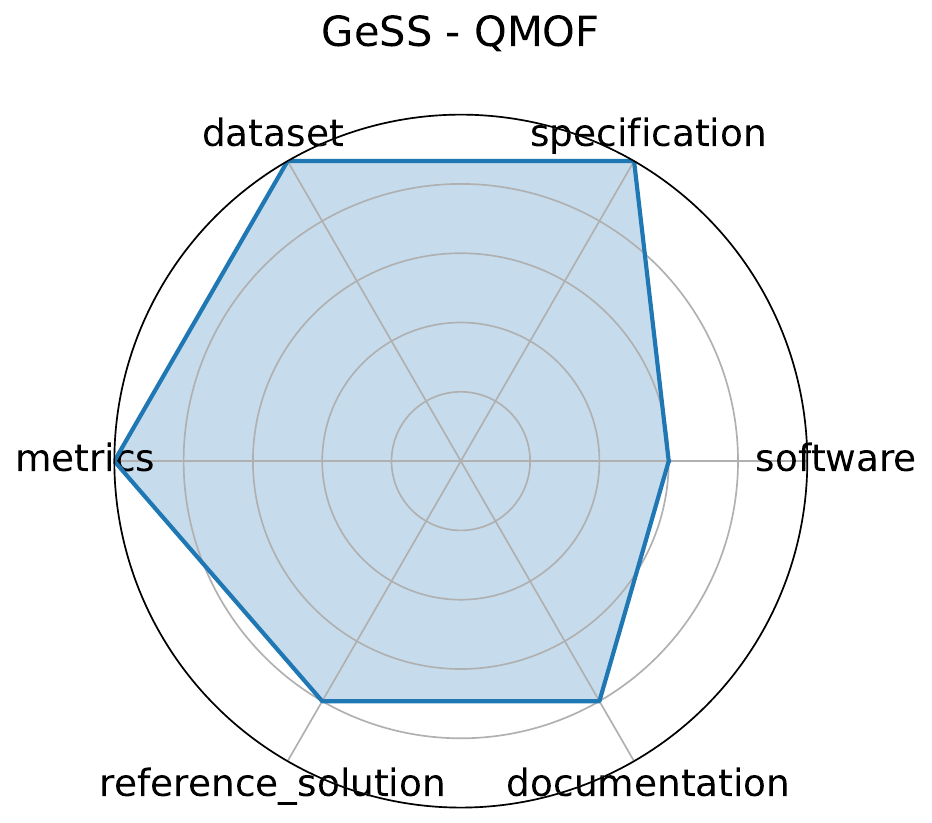} & GeSS - QMOF & Materials Science & GCN, EGNN, DimeNet++ & Accuracy, RMSE, OOD robustness delta & \cite{neurips2024_a8063075} \\ \hline
\includegraphics[width=0.05\textwidth]{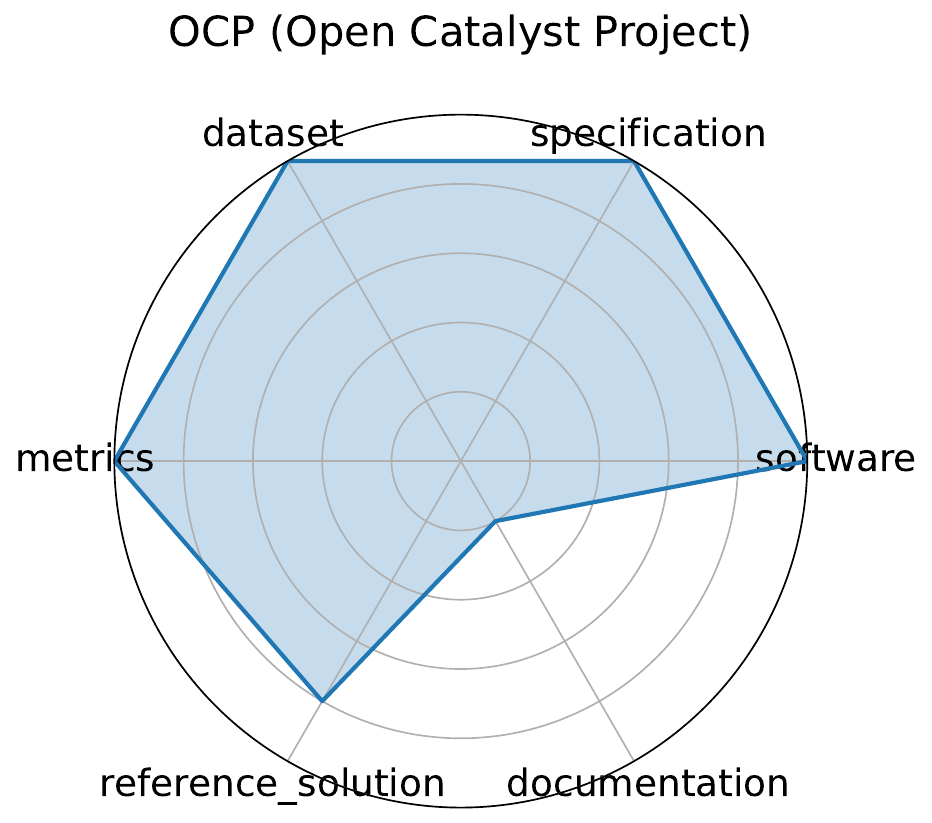} & OCP (Open Catalyst Project) & Chemistry, Materials Science & CGCNN, SchNet, DimeNet++, GemNet-OC & MAE (energy), MAE (force) & \cite{chanussot2021oc20,tran2023oc22} \\ \hline
\includegraphics[width=0.05\textwidth]{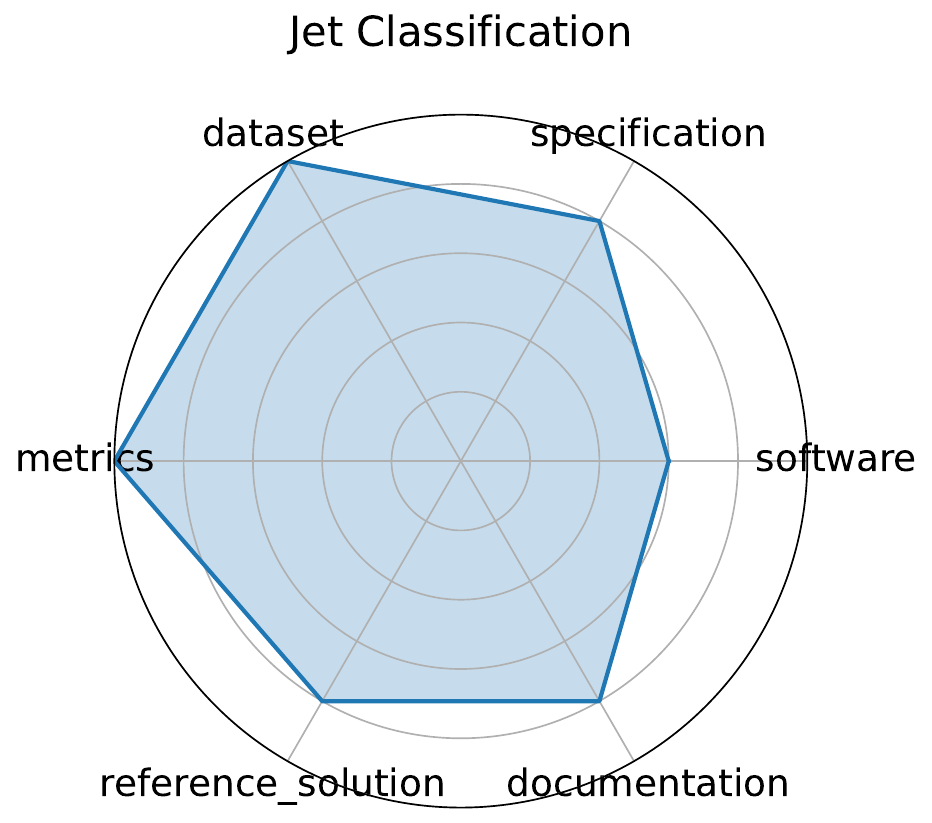} & Jet Classification & High Energy Physics & Keras DNN, QKeras quantized DNN & Accuracy, AUC & \cite{duarte2022fastml} \\ \hline
\includegraphics[width=0.05\textwidth]{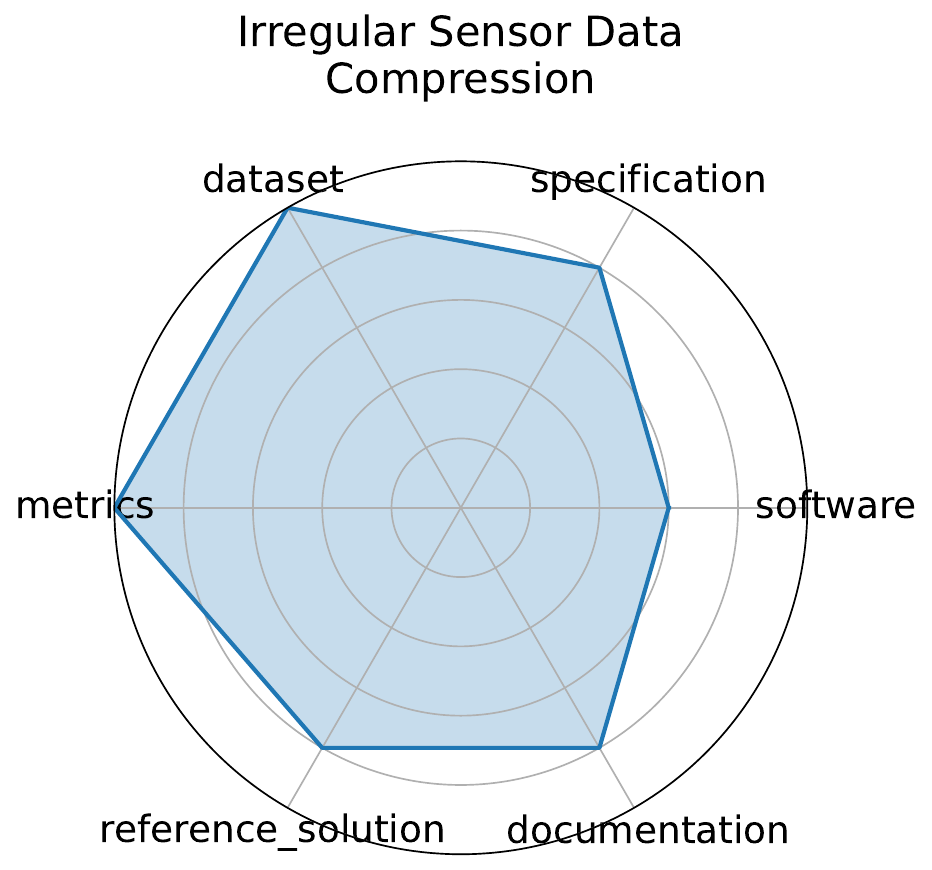} & Irregular Sensor Data Compression & High Energy Physics & Autoencoder, Quantized autoencoder & MSE, Compression ratio & \cite{duarte2022fastml} \\ \hline
\includegraphics[width=0.05\textwidth]{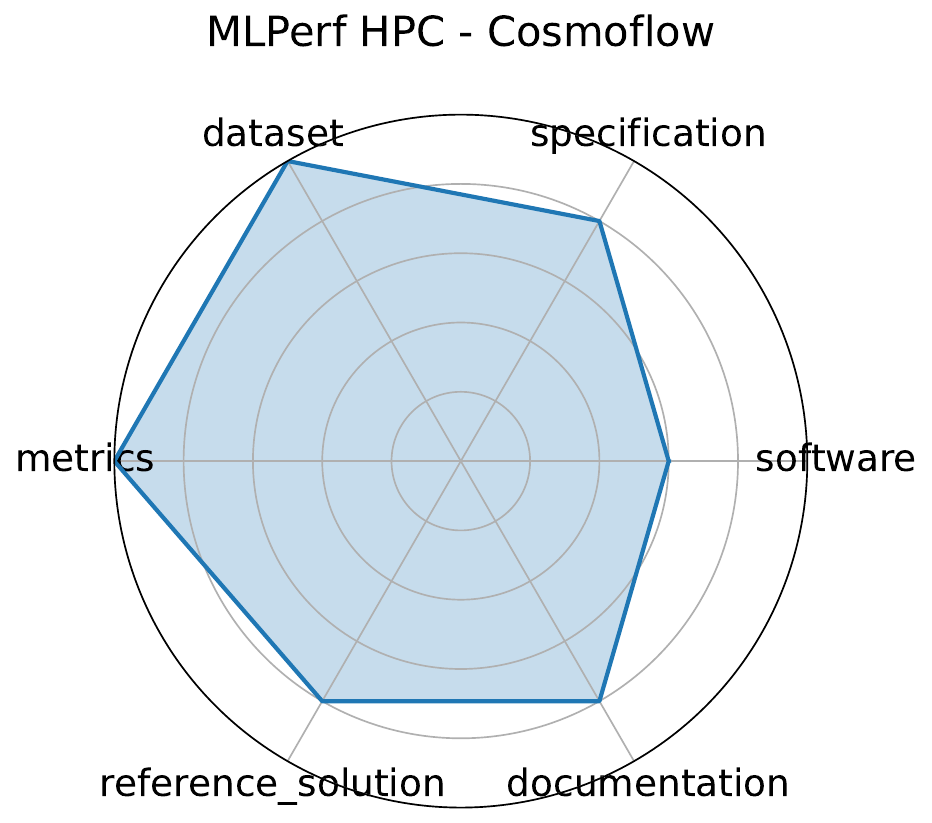} & MLPerf HPC - Cosmoflow & High Energy Physics & CosmoFlow, DeepCAM, OpenCatalyst & Training time, Accuracy, GPU utilization & \cite{farrell2021mlperfhpcholisticbenchmark} \\ \hline
\includegraphics[width=0.05\textwidth]{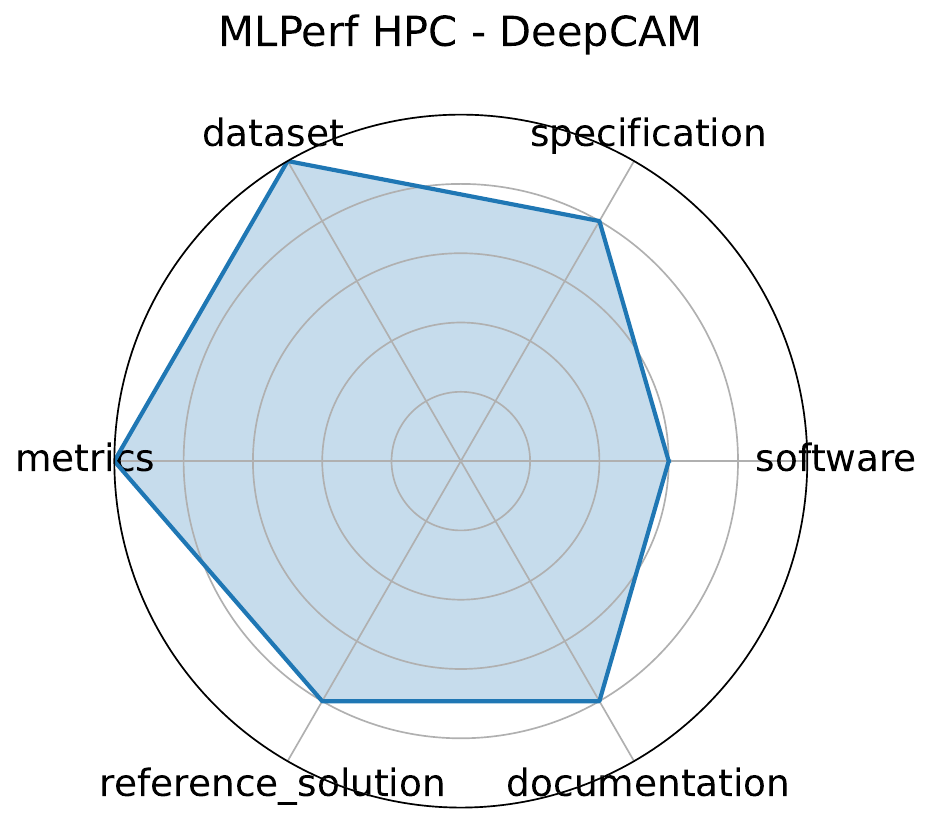} & MLPerf HPC - DeepCAM & Climate \& Earth Science & DeepCAM & Training time, Accuracy, GPU utilization & \cite{farrell2021mlperfhpcholisticbenchmark} \\ \hline
\includegraphics[width=0.05\textwidth]{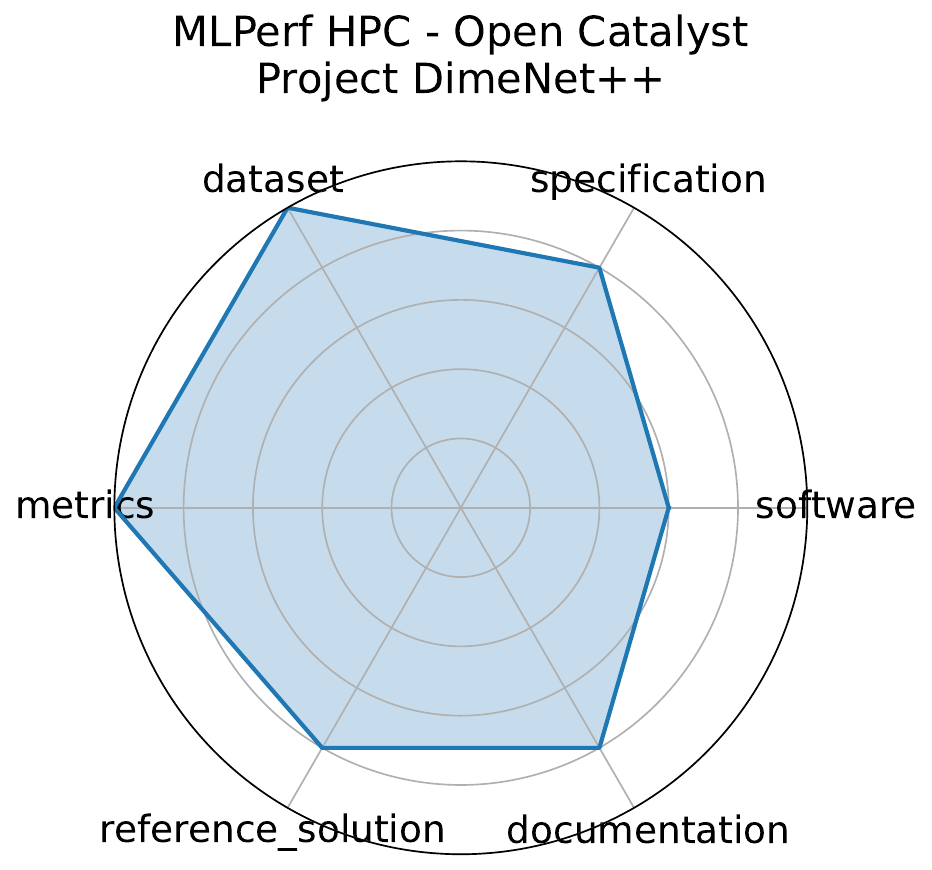} & MLPerf HPC - Open Catalyst Project DimeNet++  & Chemistry & DeepCAM & Training time, Accuracy, GPU utilization & \cite{farrell2021mlperfhpcholisticbenchmark} \\ \hline
\includegraphics[width=0.05\textwidth]{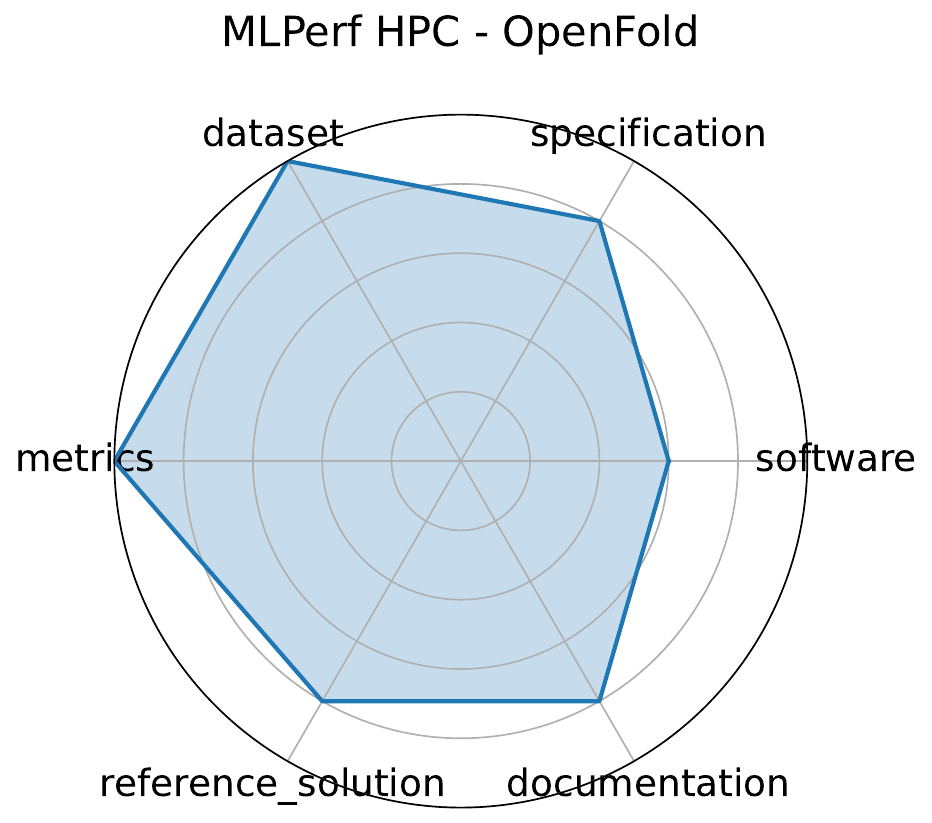} & MLPerf HPC - OpenFold & Biology \& Medicine & DeepCAM & Training time, Accuracy, GPU utilization & \cite{farrell2021mlperfhpcholisticbenchmark} \\ \hline
\includegraphics[width=0.05\textwidth]{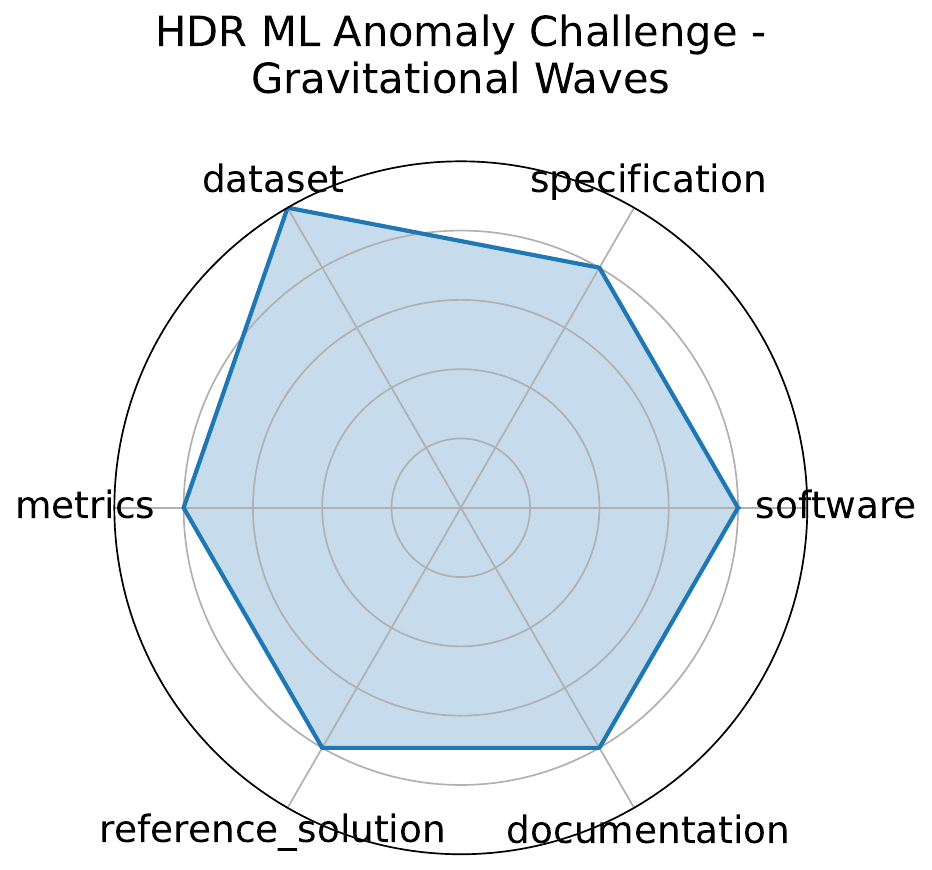} & HDR ML Anomaly Challenge - Gravitational Waves & High Energy Physics & Deep latent CNNs, Autoencoders & ROC-AUC, Precision/Recall & \cite{campolongo2025buildingmachinelearningchallenges} \\ \hline
\includegraphics[width=0.05\textwidth]{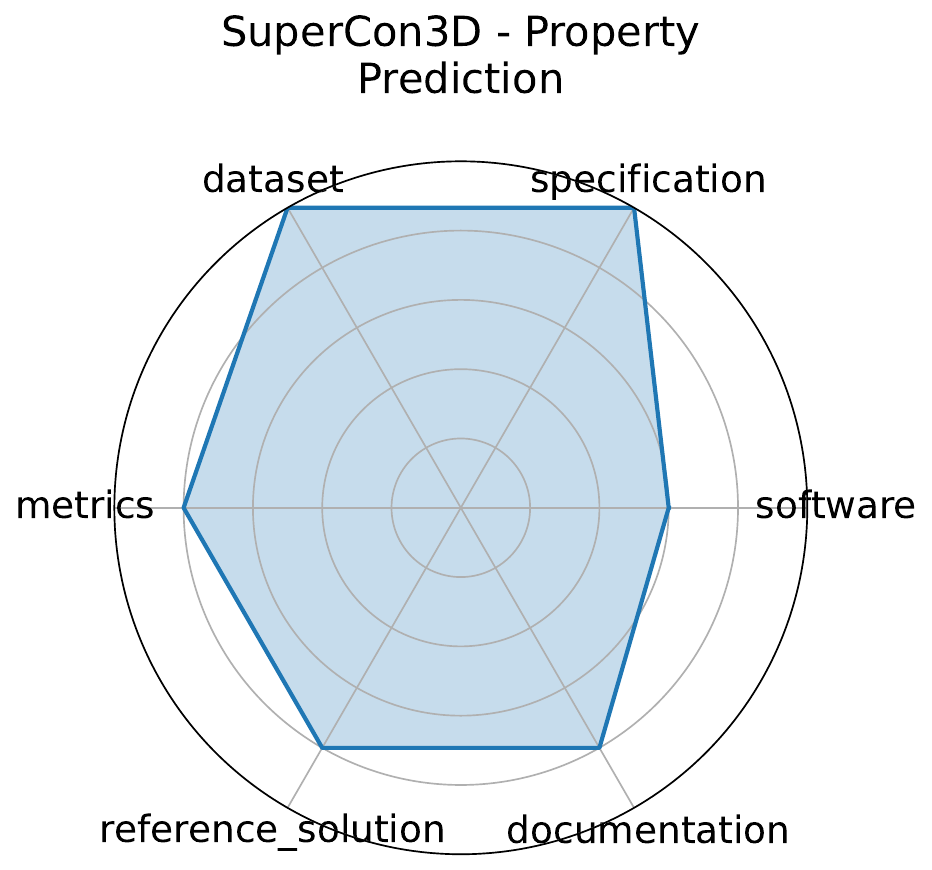} & SuperCon3D - Property Prediction & Materials Science & SODNet, DiffCSP-SC & MAE (Tc), Validity of generated structures & \cite{neurips2024_c4e3b55e} \\ \hline
\includegraphics[width=0.05\textwidth]{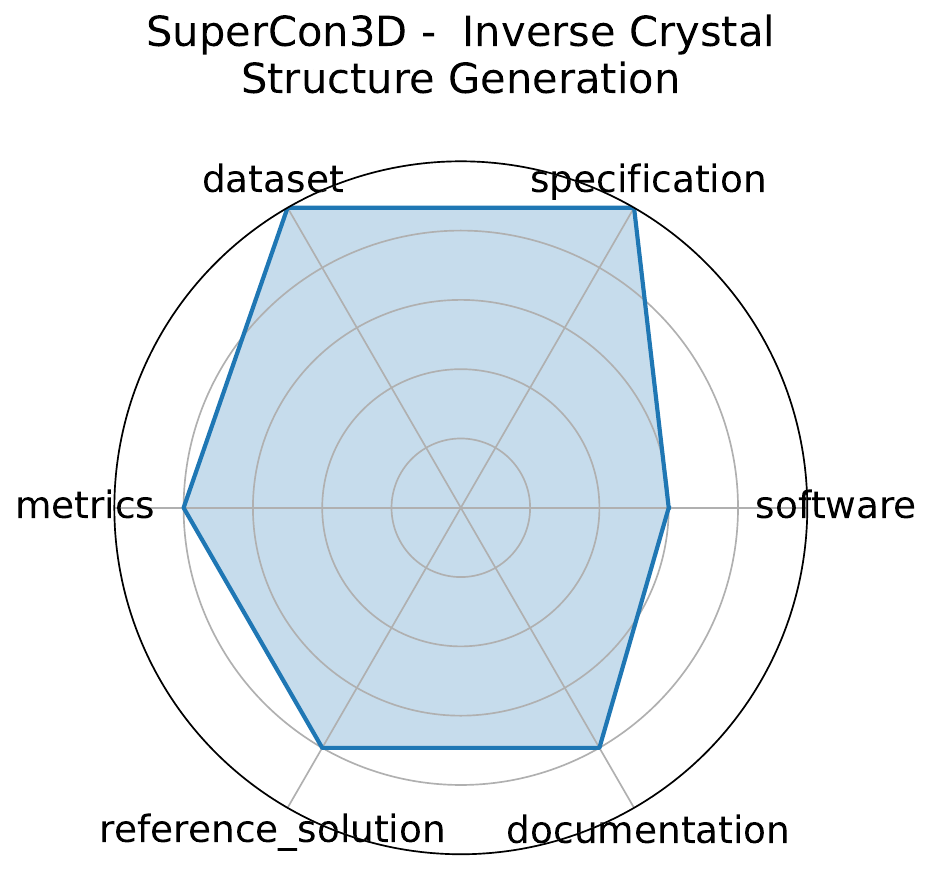} & SuperCon3D -  Inverse Crystal Structure Generation & Materials Science & SODNet, DiffCSP-SC & MAE (Tc), Validity of generated structures & \cite{neurips2024_c4e3b55e} \\ \hline
\includegraphics[width=0.05\textwidth]{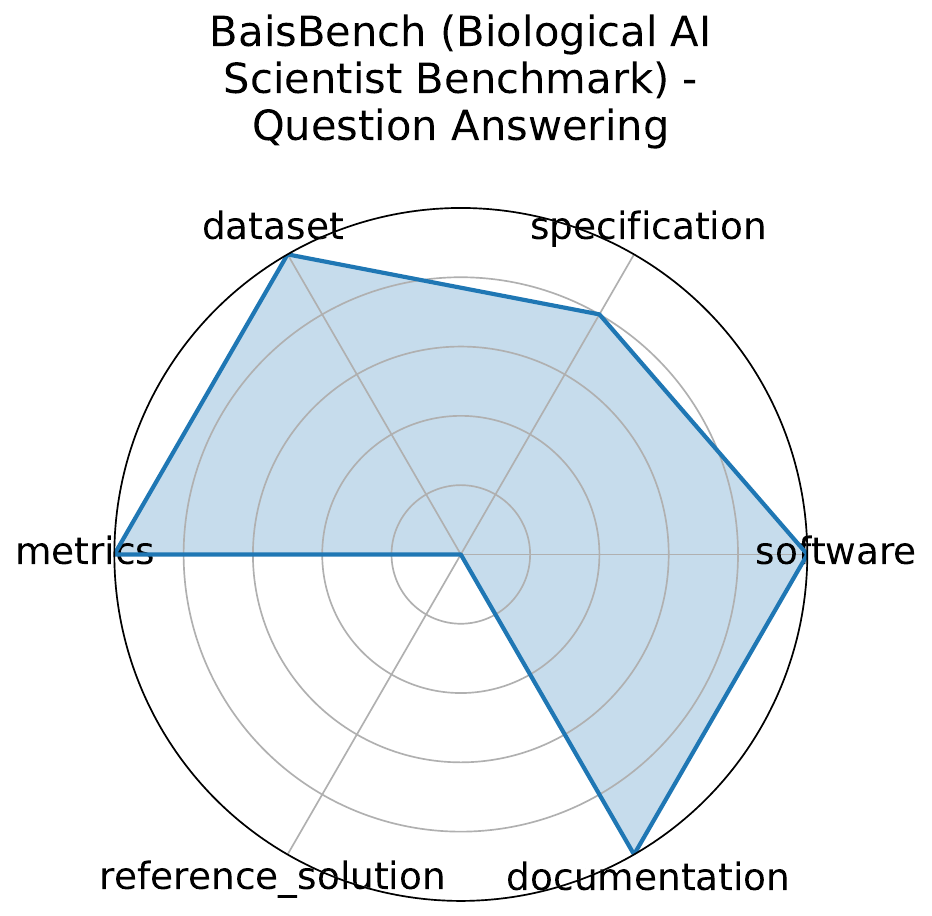} & BaisBench (Biological AI Scientist Benchmark) - Question Answering & Biology \& Medicine & LLM-based AI scientist agents & Annotation accuracy, QA accuracy & \cite{luo2025benchmarkingaiscientistsomics} \\ \hline
\includegraphics[width=0.05\textwidth]{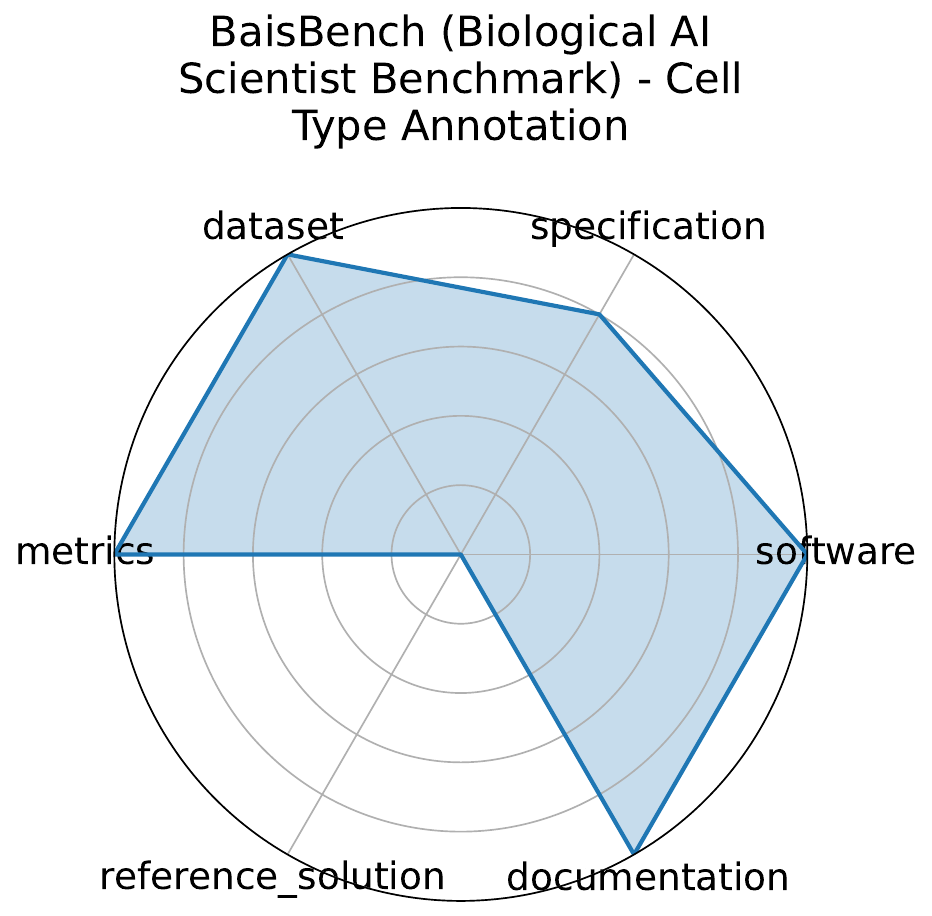} & BaisBench (Biological AI Scientist Benchmark) - Cell Type Annotation & Biology \& Medicine & LLM-based AI scientist agents & Annotation accuracy, QA accuracy & \cite{luo2025benchmarkingaiscientistsomics} \\ \hline
\includegraphics[width=0.05\textwidth]{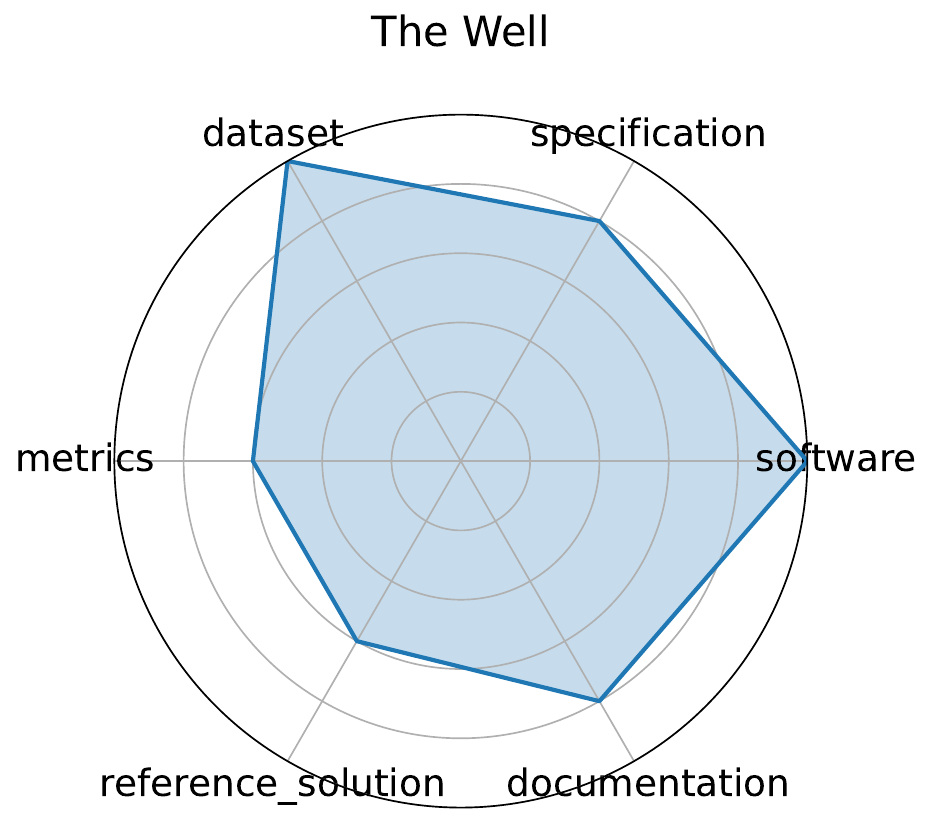} & The Well & Biology \& Medicine, Computational Science \& AI, High Energy Physics & FNO baselines, U-Net baselines & Dataset size, Domain breadth & \cite{neurips2024_4f9a5acd} \\ \hline
\includegraphics[width=0.05\textwidth]{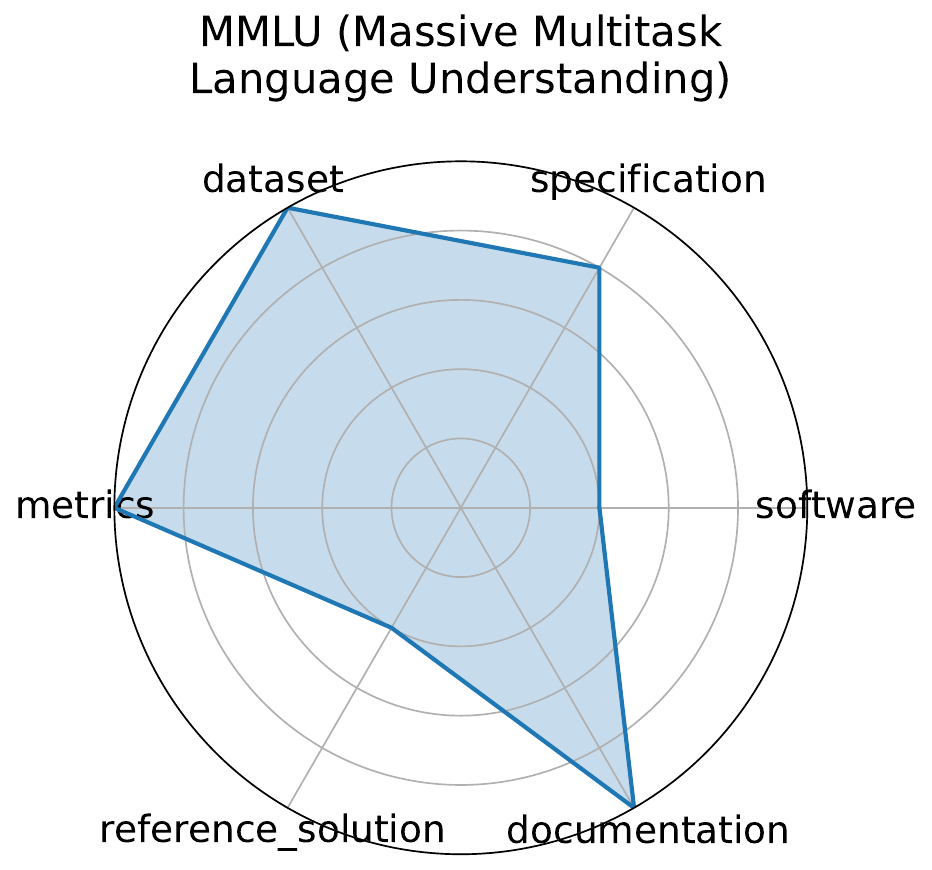} & MMLU (Massive Multitask Language Understanding) & Computational Science \& AI & GPT-4o, Gemini 1.5 Pro, o1, DeepSeek-R1 & Accuracy & \cite{hendrycks2021measuring} \\ \hline
\includegraphics[width=0.05\textwidth]{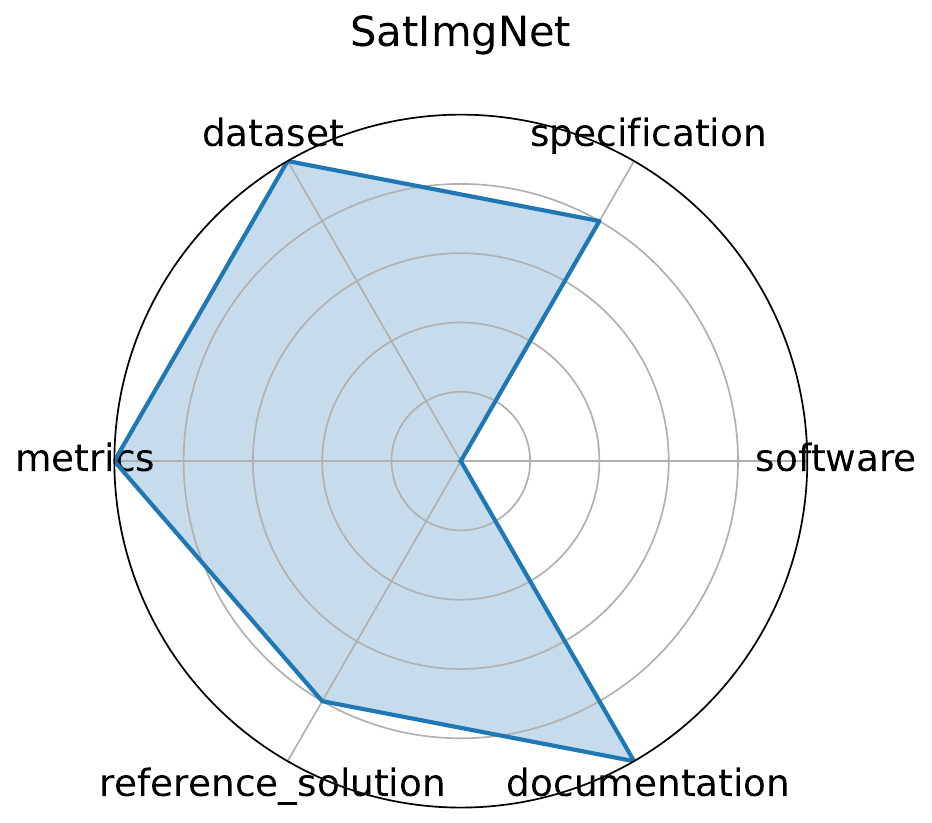} & SatImgNet & Climate \& Earth Science & CLIP, BLIP, ALBEF & Accuracy & \cite{roberts2023satin} \\ \hline
\includegraphics[width=0.05\textwidth]{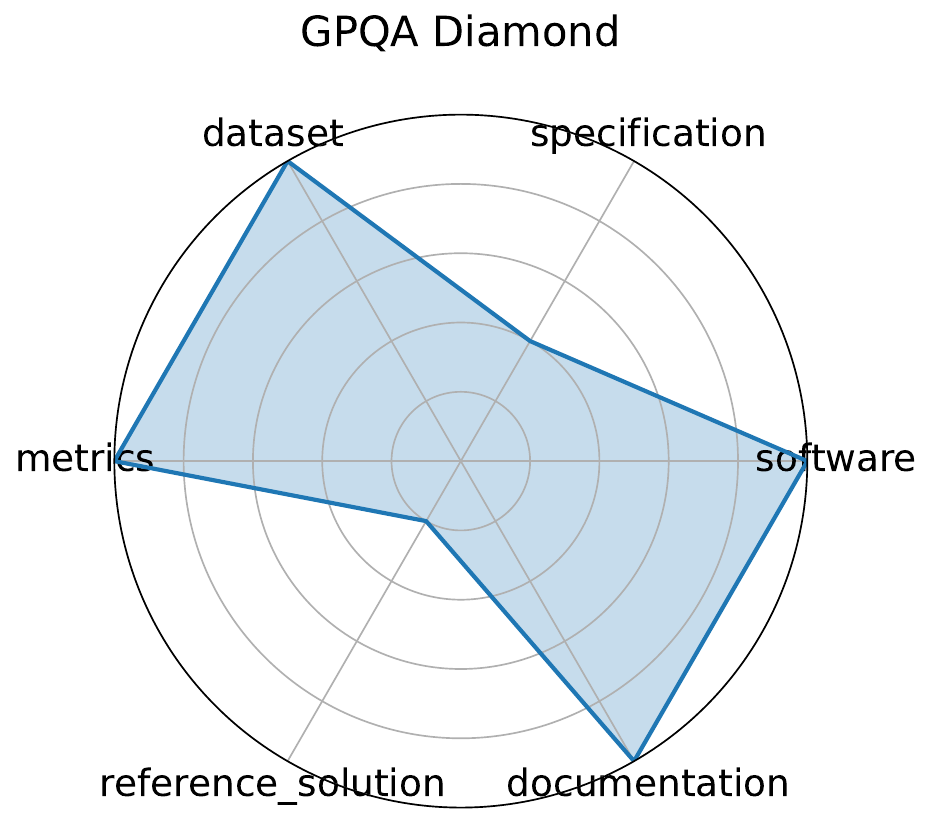} & GPQA Diamond & Biology \& Medicine, Chemistry, High Energy Physics & o1, DeepSeek-R1 & Accuracy & \cite{rein2023gpqagraduatelevelgoogleproofqa} \\ \hline
\includegraphics[width=0.05\textwidth]{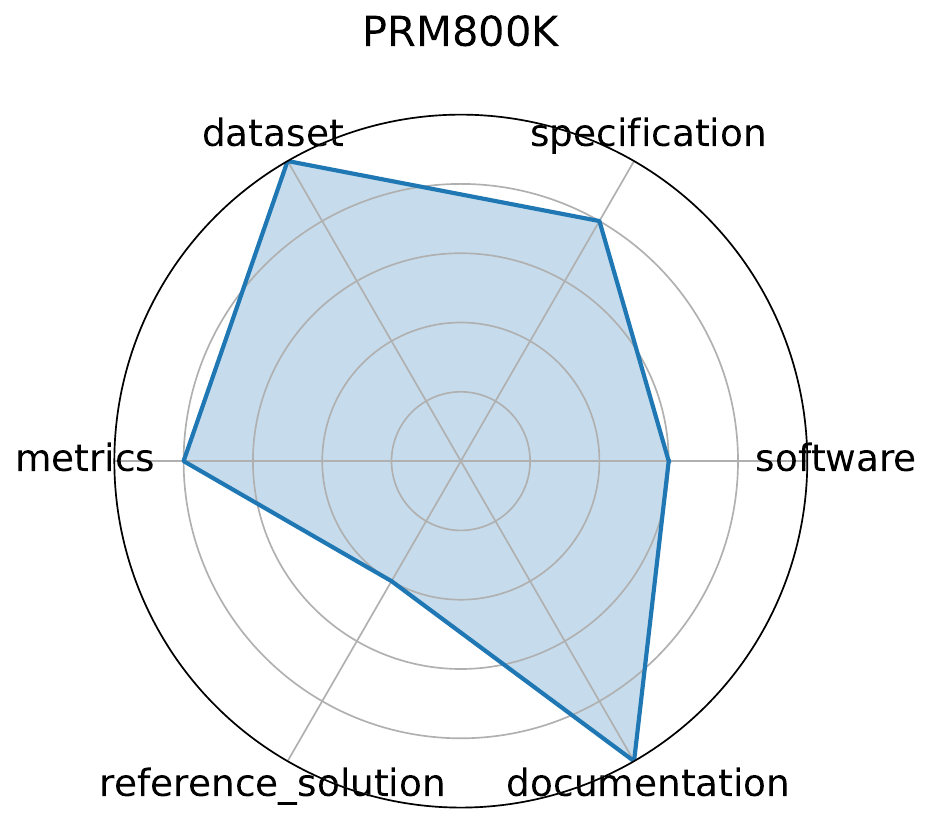} & PRM800K & Mathematics & GPT-4 & Accuracy & \cite{lightman2023lets} \\ \hline
\includegraphics[width=0.05\textwidth]{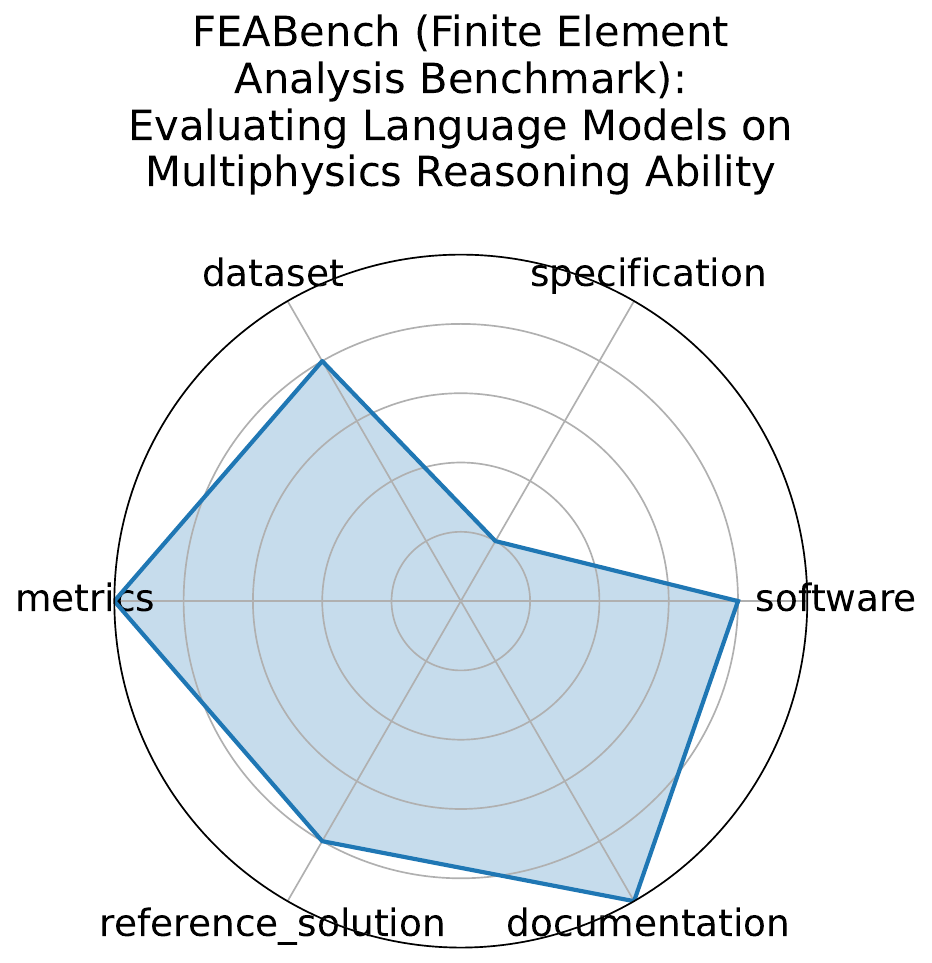} & FEABench (Finite Element Analysis Benchmark): Evaluating Language Models on Multiphysics Reasoning Ability & Mathematics & FEniCS, deal.II & Solve time, Error norm & \cite{mudur2025feabenchevaluatinglanguagemodels} \\ \hline
\includegraphics[width=0.05\textwidth]{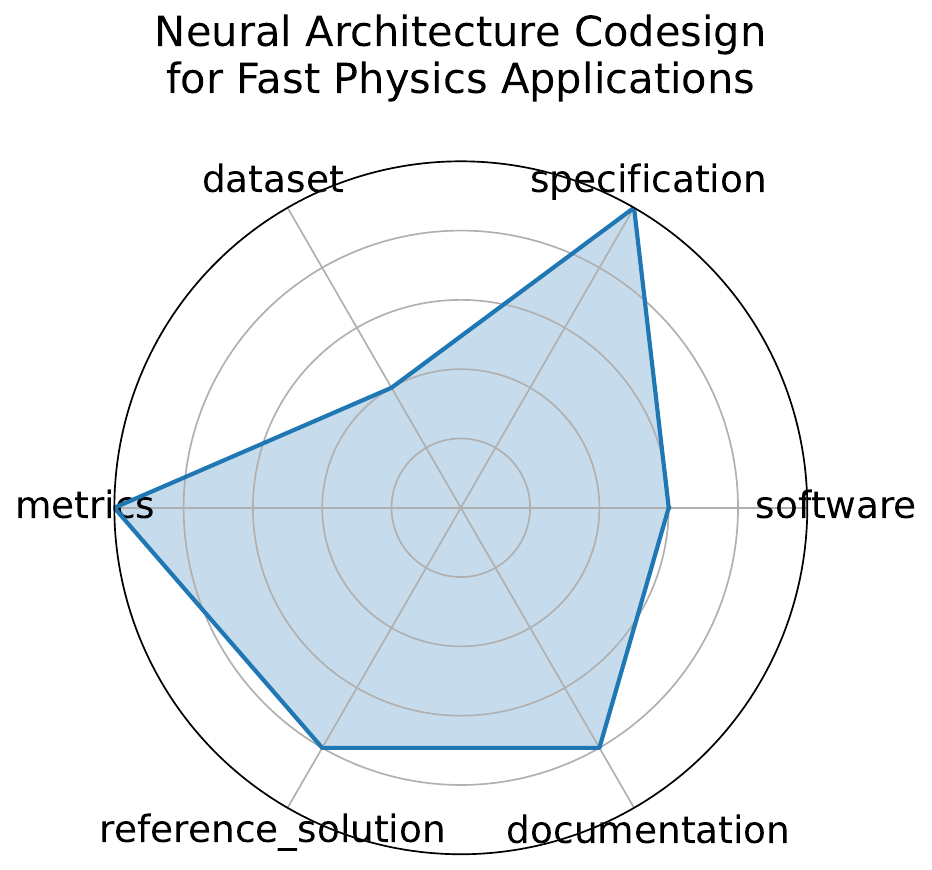} & Neural Architecture Codesign for Fast Physics Applications & High Energy Physics & NAC-based BraggNN, NAC-optimized Deep Sets (jet) & Accuracy, Latency, Resource utilization & \cite{weitz2025neuralarchitecturecodesignfast} \\ \hline
\includegraphics[width=0.05\textwidth]{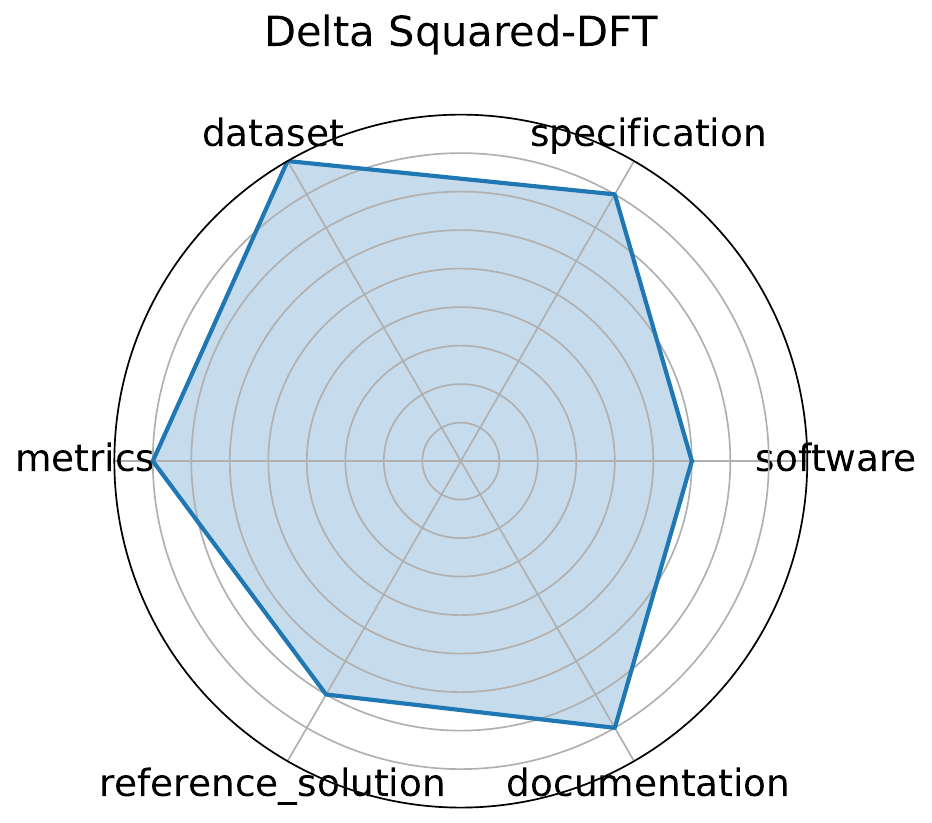} & Delta Squared-DFT & Chemistry, Materials Science & Delta Squared-ML correction networks, Kernel ridge regression & Mean Absolute Error (eV), Energy ranking accuracy & \cite{khrabrov2024nabla2dftuniversalquantumchemistry} \\ \hline
\includegraphics[width=0.05\textwidth]{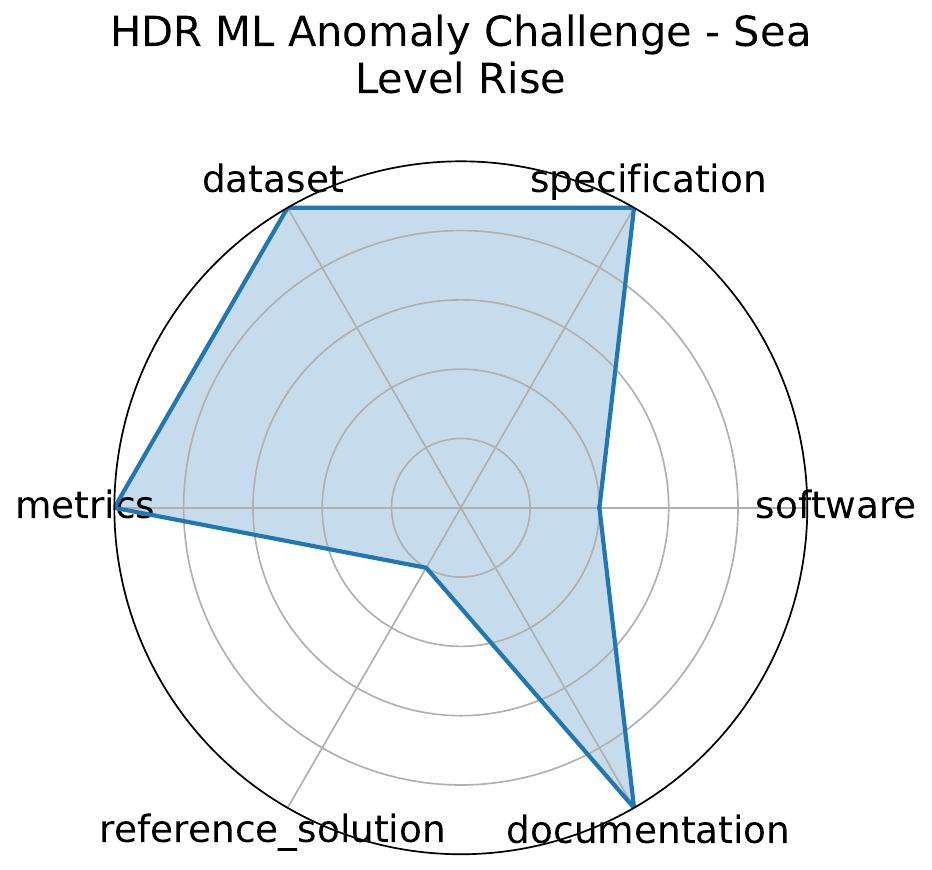} & HDR ML Anomaly Challenge - Sea Level Rise & Climate \& Earth Science & CNNs, RNNs, Transformers & ROC-AUC, Precision/Recall & \cite{campolongo2025buildingmachinelearningchallenges} \\ \hline
\includegraphics[width=0.05\textwidth]{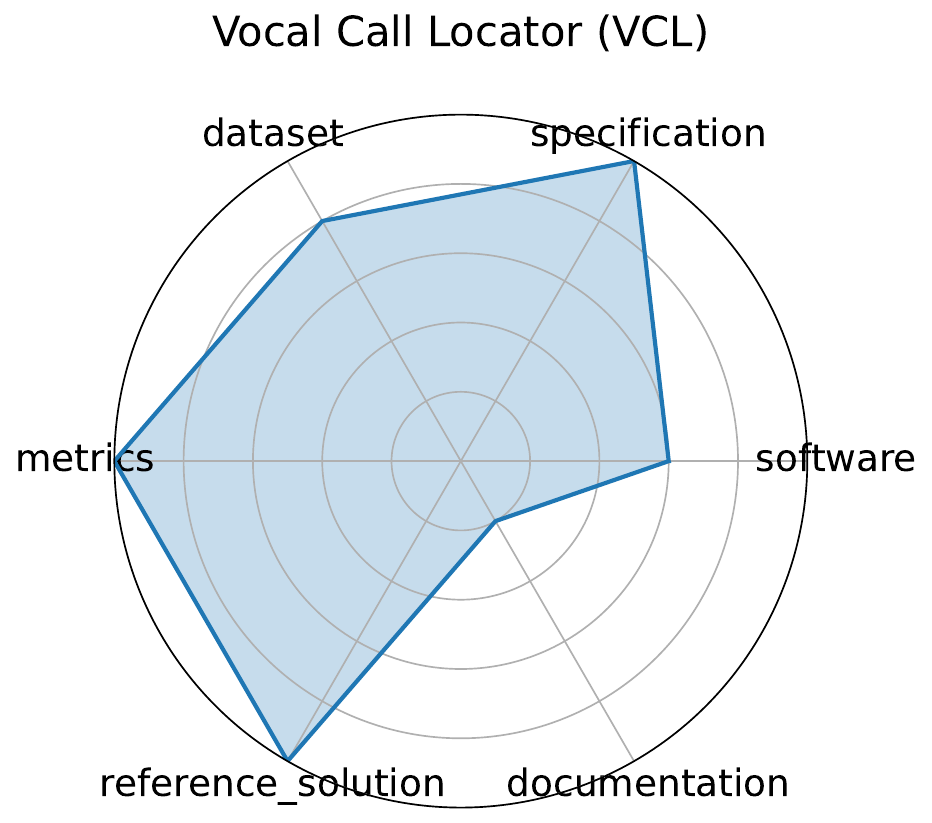} & Vocal Call Locator (VCL) & Biology \& Medicine & CNN-based SSL models & Localization error (cm), Recall/Precision & \cite{neurips2024_c00d37d6} \\ \hline
\includegraphics[width=0.05\textwidth]{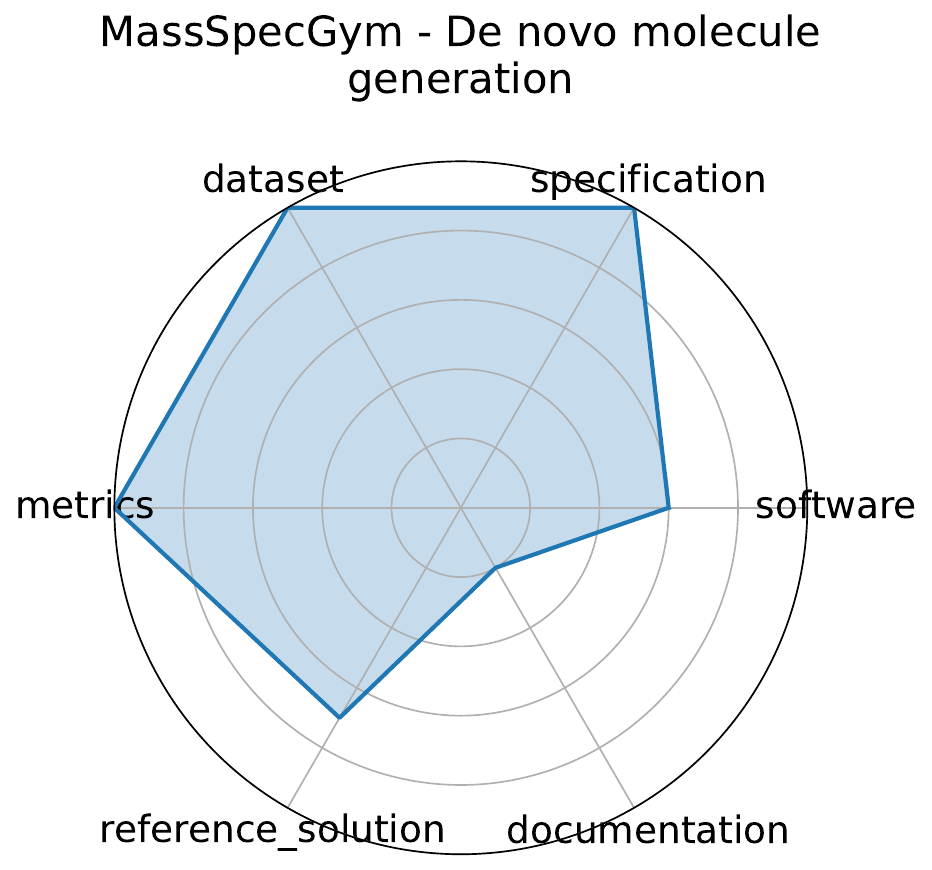} & MassSpecGym - De novo molecule generation & Chemistry & Graph-based generative models, Retrieval baselines & Structure accuracy, Retrieval precision, Simulation MSE & \cite{neurips2024_c6c31413} \\ \hline
\includegraphics[width=0.05\textwidth]{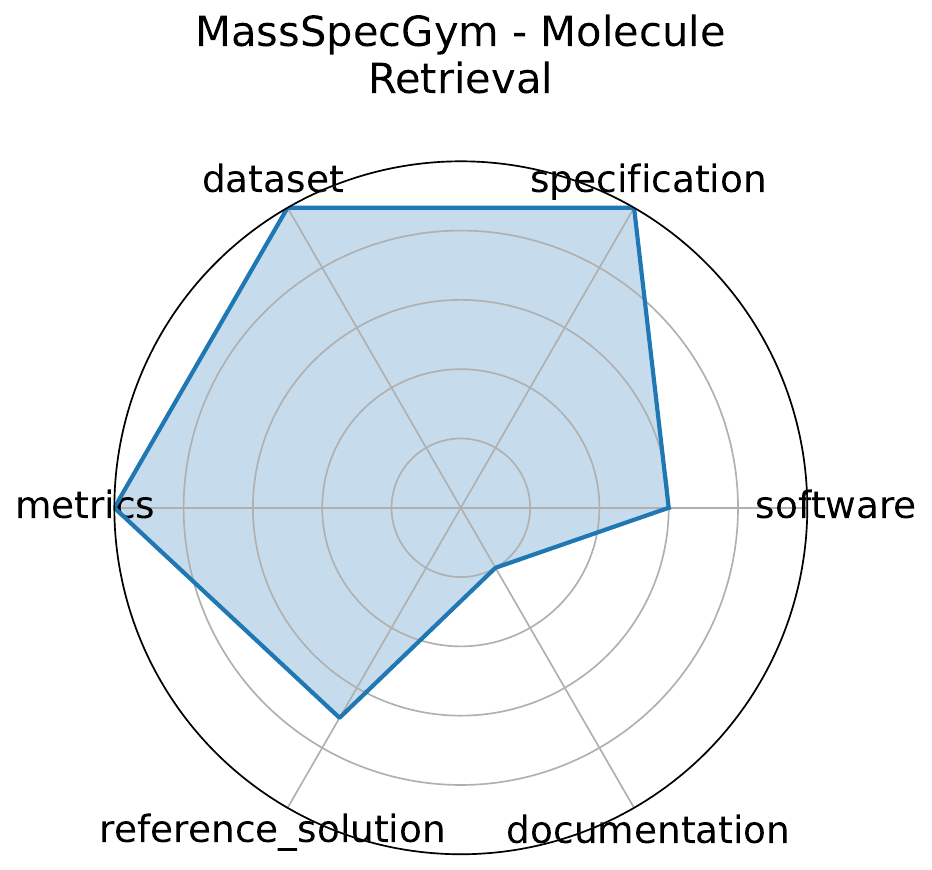} & MassSpecGym - Molecule Retrieval & Chemistry & Graph-based generative models, Retrieval baselines & Structure accuracy, Retrieval precision, Simulation MSE & \cite{neurips2024_c6c31413} \\ \hline
\includegraphics[width=0.05\textwidth]{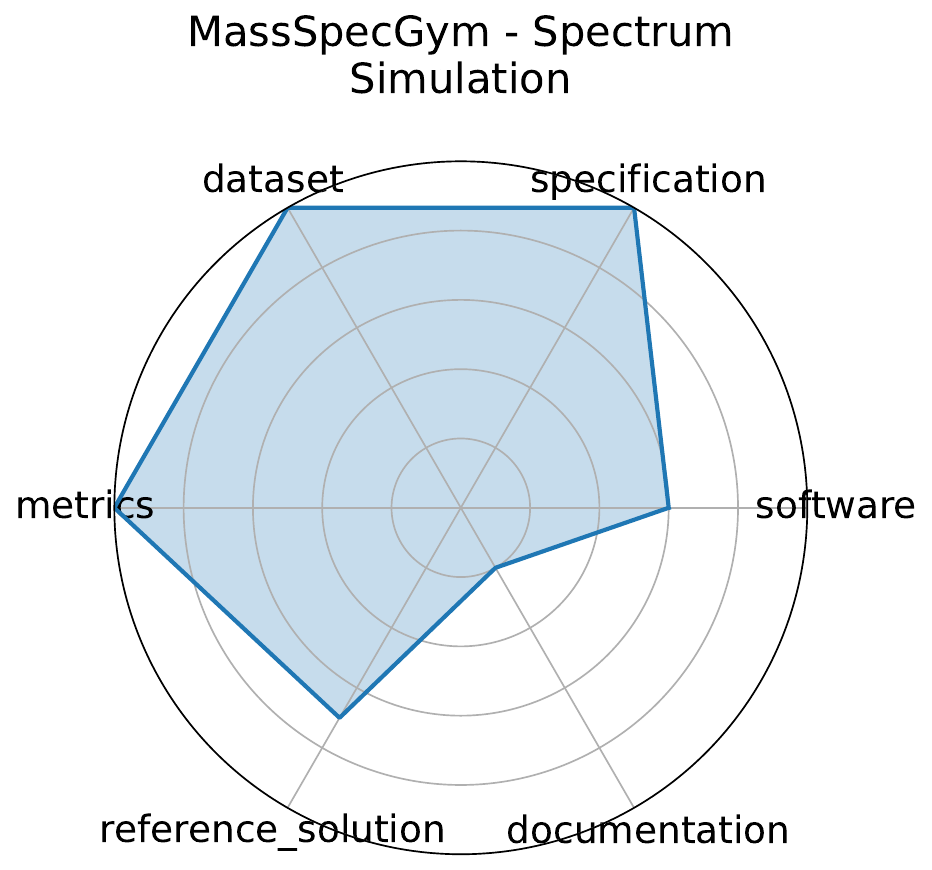} & MassSpecGym - Spectrum Simulation & Chemistry & Graph-based generative models, Retrieval baselines & Structure accuracy, Retrieval precision, Simulation MSE & \cite{neurips2024_c6c31413} \\ \hline
\includegraphics[width=0.05\textwidth]{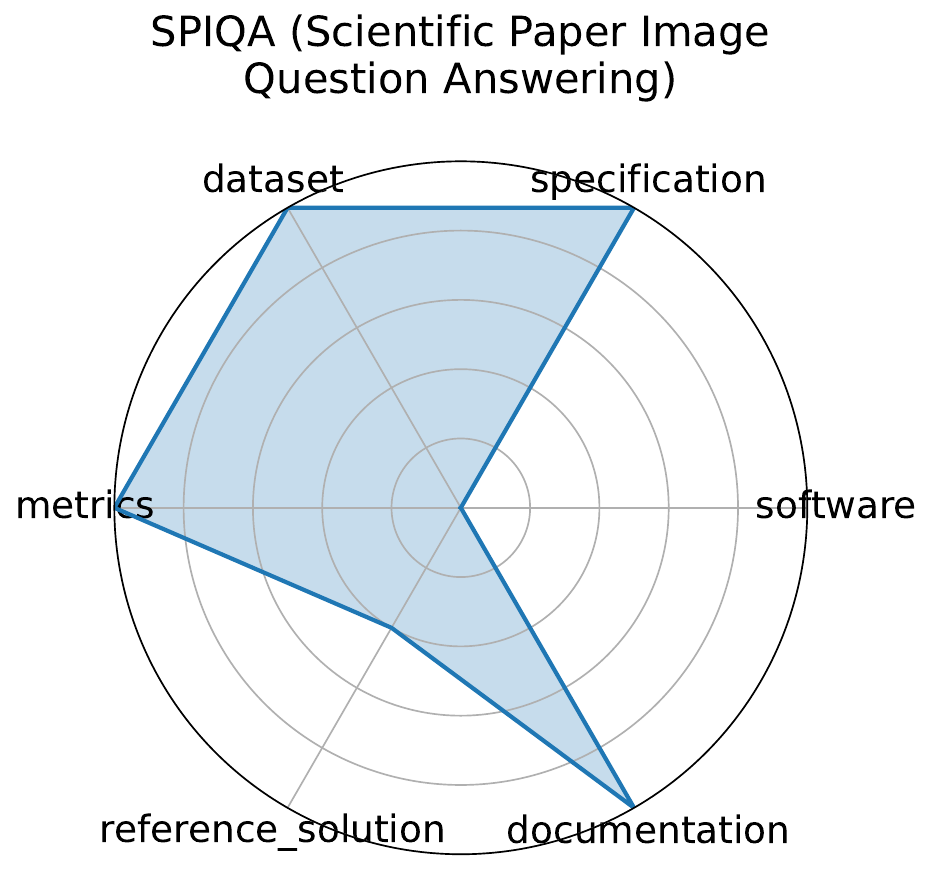} & SPIQA (Scientific Paper Image Question Answering) & Computational Science \& AI & Chain-of-Thought models, Multimodal QA systems & Accuracy, F1 score & \cite{zhong2024spiqa} \\ \hline
\includegraphics[width=0.05\textwidth]{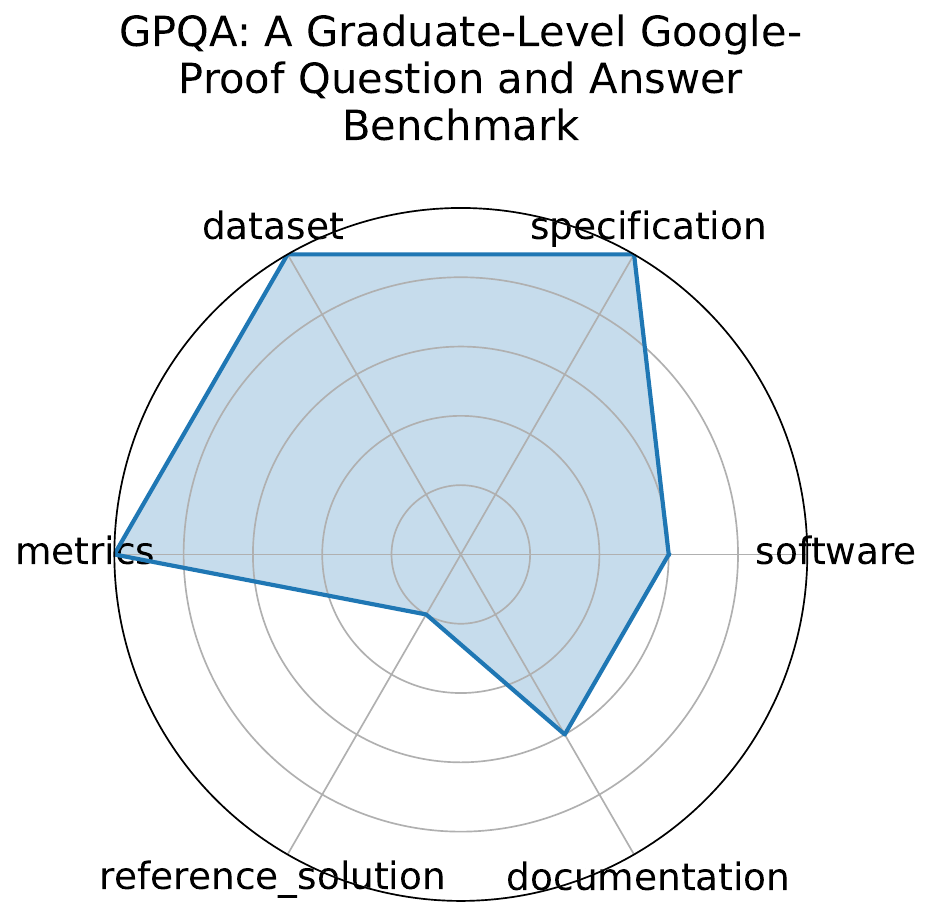} & GPQA: A Graduate-Level Google-Proof Question and Answer Benchmark & Biology \& Medicine, High Energy Physics, Chemistry & GPT-4 baseline & Accuracy & \cite{rein2023gpqagraduatelevelgoogleproofqa2} \\ \hline
\includegraphics[width=0.05\textwidth]{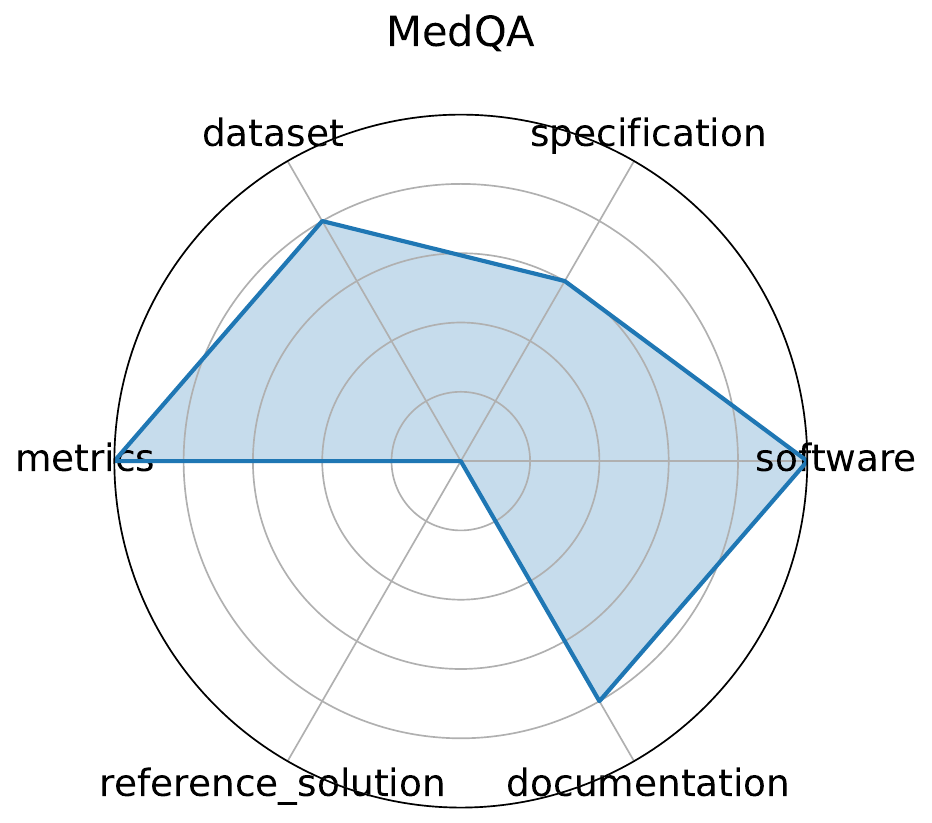} & MedQA & Biology \& Medicine & Neural reader, Retrieval-based QA systems & Accuracy & \cite{jin2020diseasedoespatienthave} \\ \hline
\includegraphics[width=0.05\textwidth]{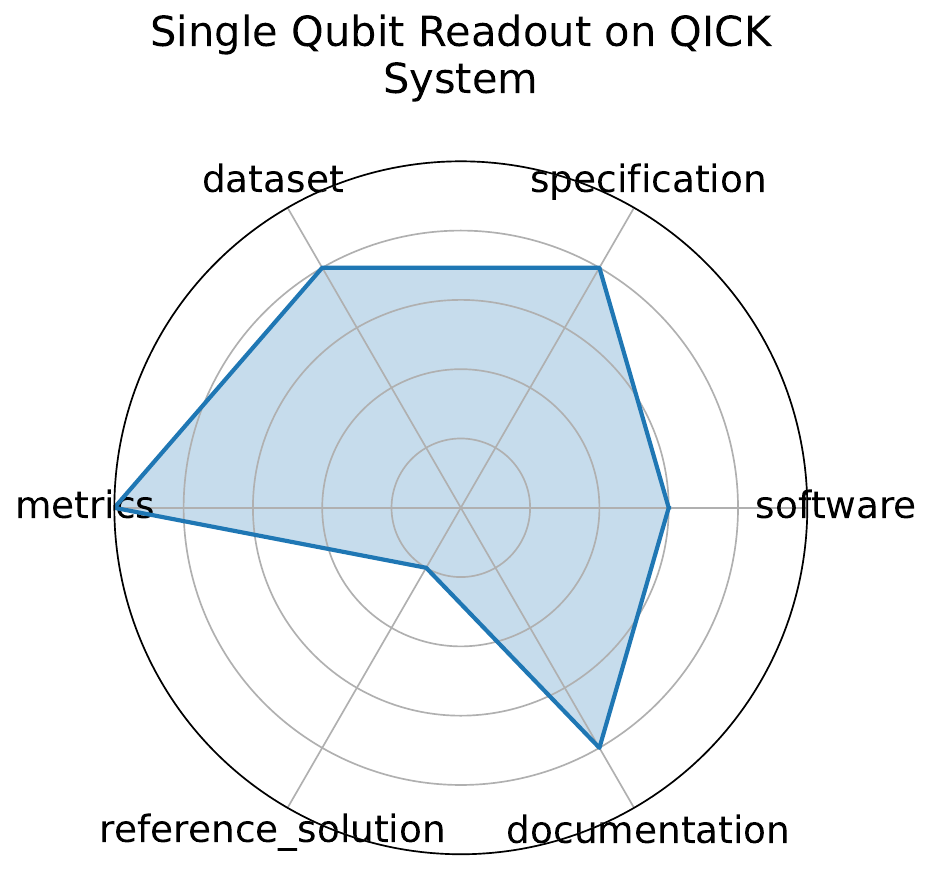} & Single Qubit Readout on QICK System & Computational Science \& AI & hls4ml quantized NN & Accuracy, Latency & \cite{diguglielmo2025endtoendworkflowmachinelearningbased} \\ \hline
\includegraphics[width=0.05\textwidth]{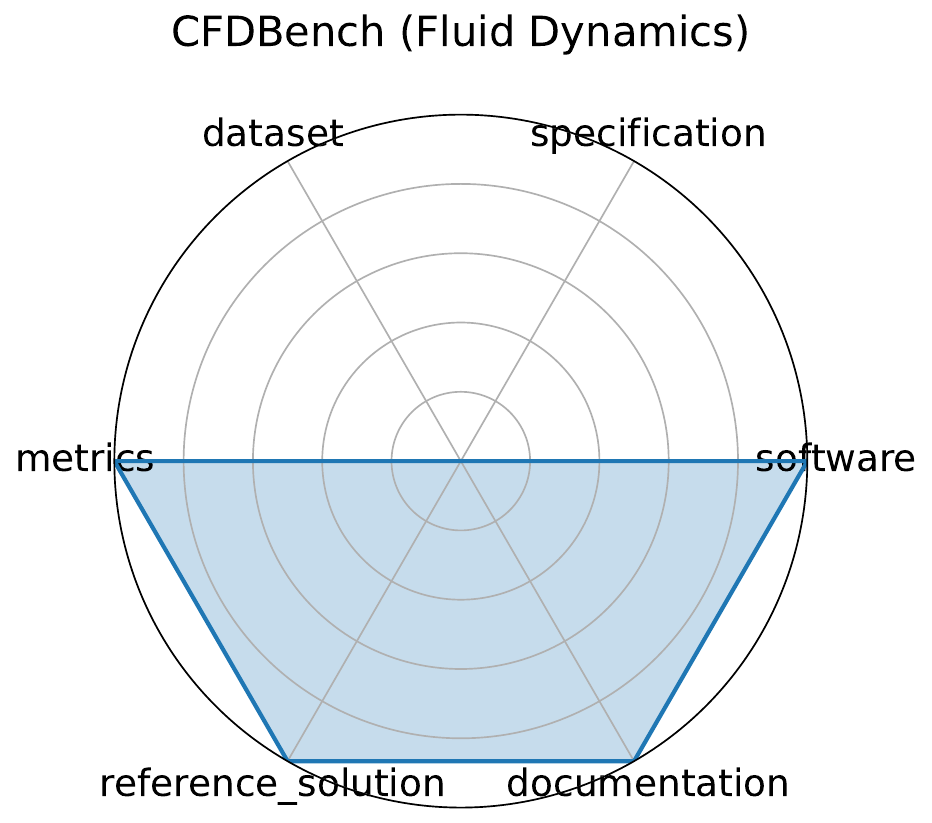} & CFDBench (Fluid Dynamics) & Mathematics & FNO, DeepONet, U-Net & L2 error, MAE & \cite{luo2024cfdbenchlargescalebenchmarkmachine} \\ \hline
\includegraphics[width=0.05\textwidth]{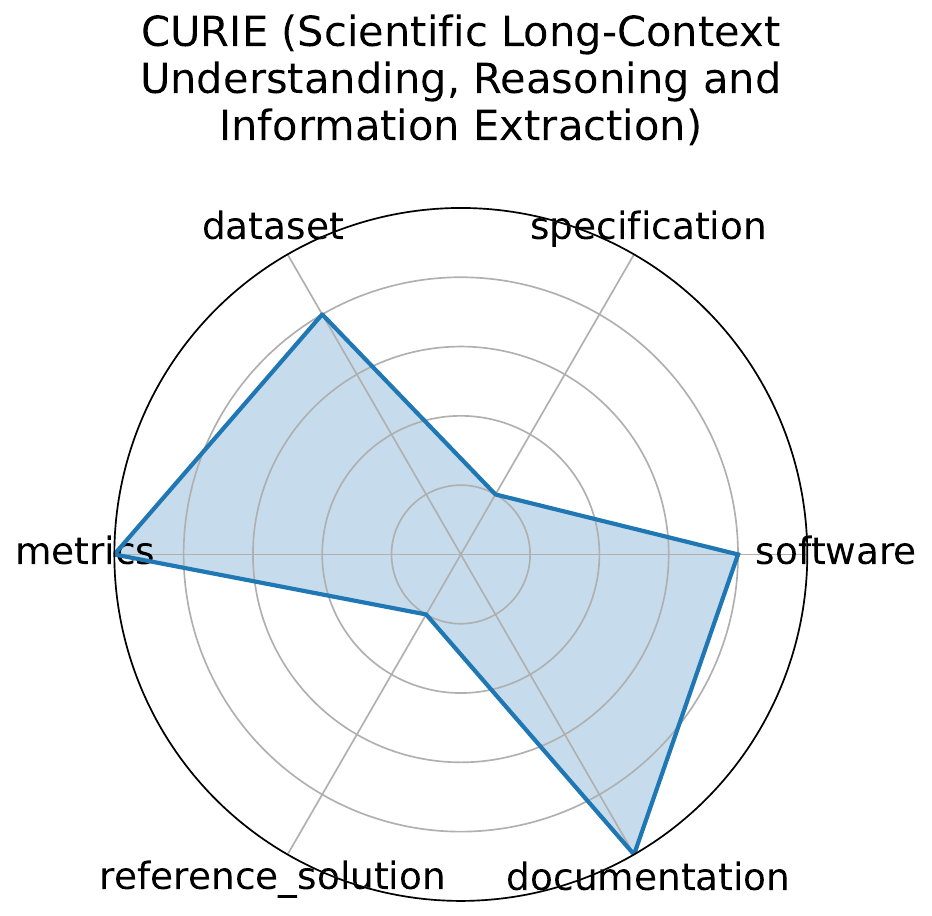} & CURIE (Scientific Long-Context Understanding, Reasoning and Information Extraction) & Materials Science, High Energy Physics, Biology \& Medicine, Chemistry, Climate \& Earth Science & unkown & Accuracy & \cite{cui2025curieevaluatingllmsmultitask} \\ \hline
\includegraphics[width=0.05\textwidth]{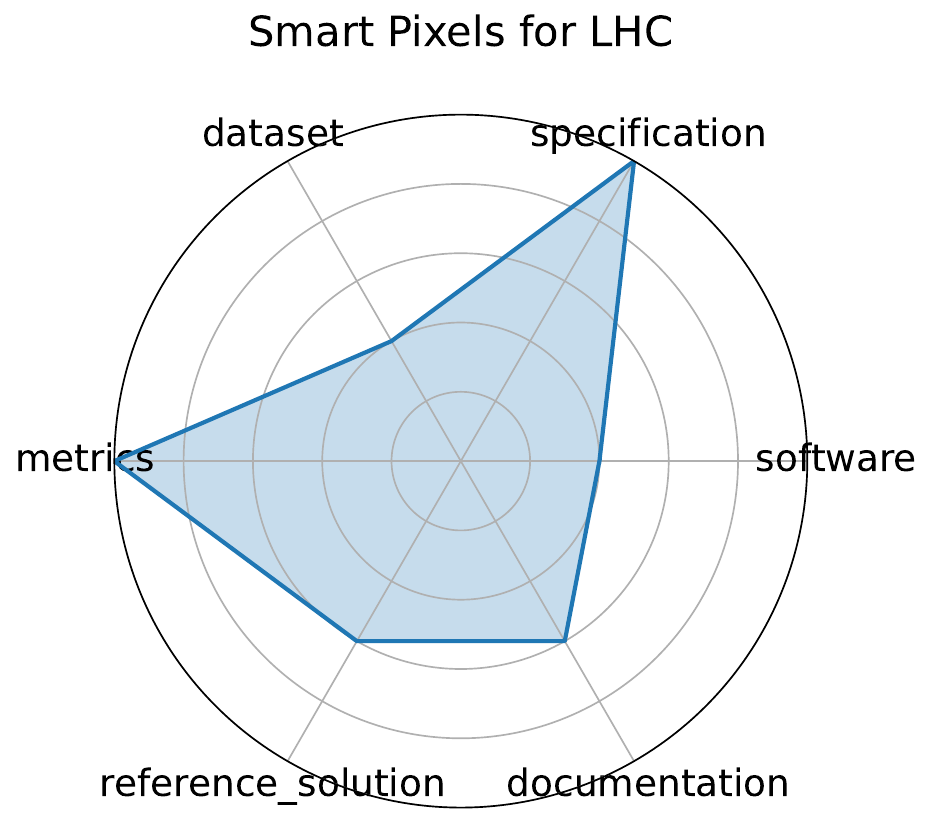} & Smart Pixels for LHC & High Energy Physics & 2-layer pixel NN & Data rejection rate, Power per pixel & \cite{parpillon2024smartpixelsinpixelai} \\ \hline
\includegraphics[width=0.05\textwidth]{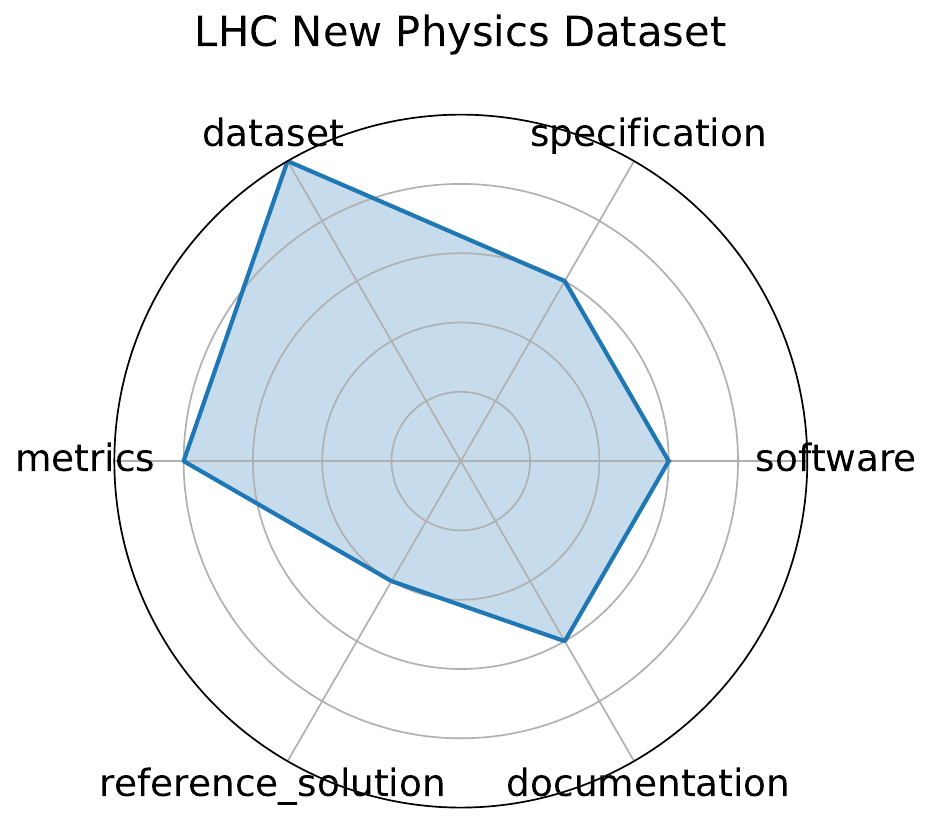} & LHC New Physics Dataset & High Energy Physics & Autoencoder, Variational autoencoder, Isolation forest & ROC-AUC, Detection efficiency & \cite{https://doi.org/10.5281/zenodo.5046389} \\ \hline
\includegraphics[width=0.05\textwidth]{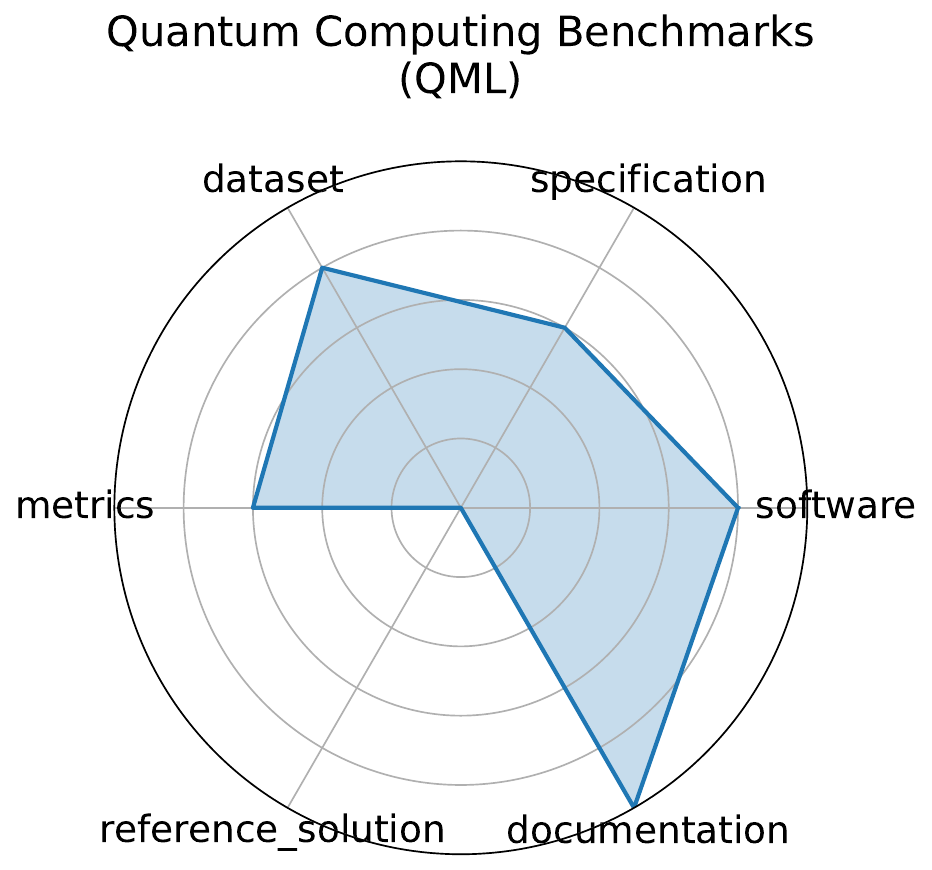} & Quantum Computing Benchmarks (QML) & Computational Science \& AI & IBM Q, IonQ, AQT@LBNL & Fidelity, Success probability & \cite{bowles2024betterclassicalsubtleart} \\ \hline
\includegraphics[width=0.05\textwidth]{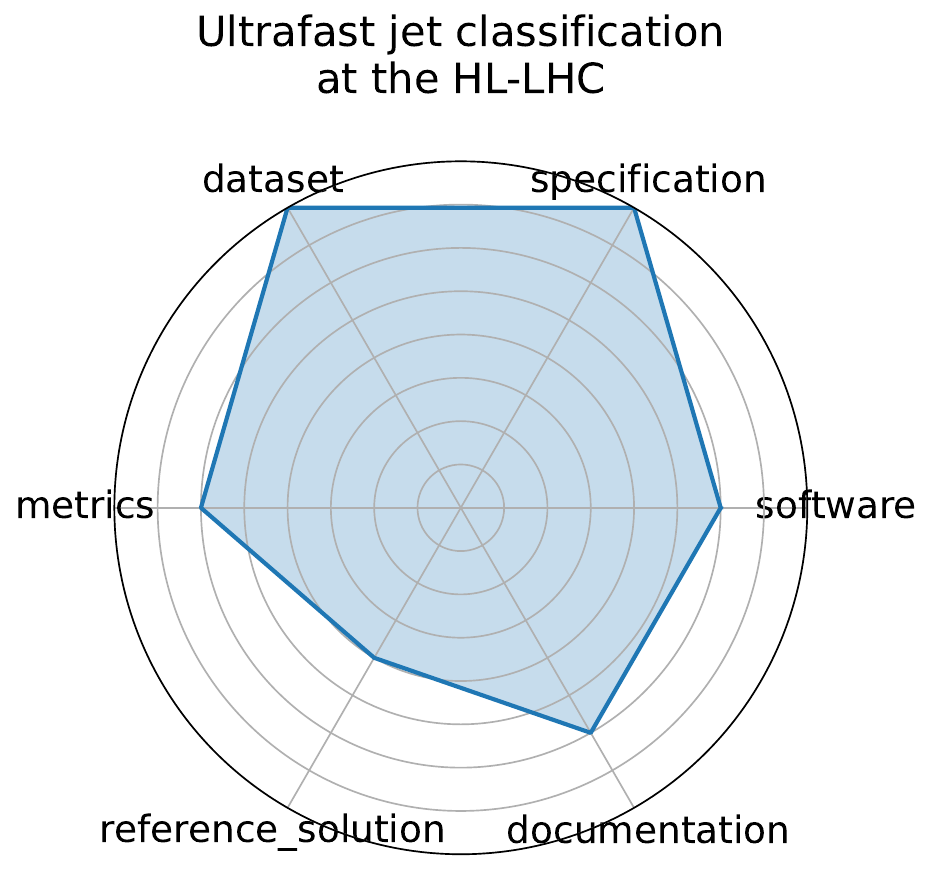} & Ultrafast jet classification at the HL-LHC & High Energy Physics & MLP, Deep Sets, Interaction Network & Accuracy, Latency, Resource utilization & \cite{odagiu2024ultrafastjetclassificationfpgas} \\ \hline
\includegraphics[width=0.05\textwidth]{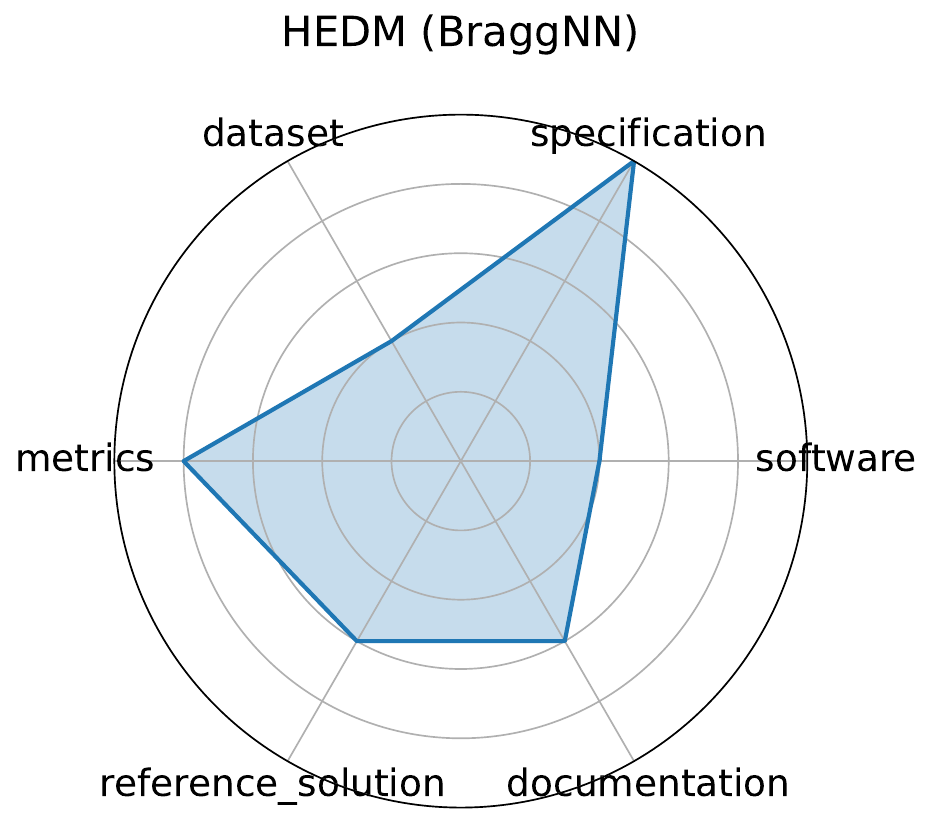} & HEDM (BraggNN) & Materials Science & BraggNN & Localization accuracy, Inference time & \cite{liu2021braggnnfastxraybragg} \\ \hline
\includegraphics[width=0.05\textwidth]{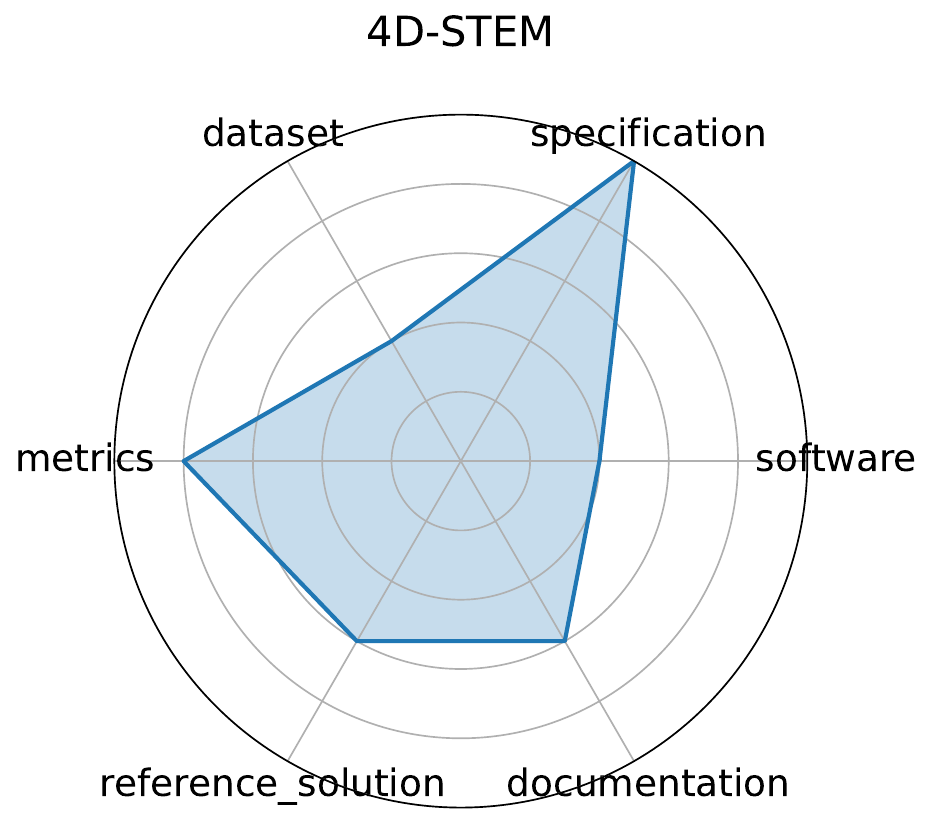} & 4D-STEM & Materials Science & CNN models (prototype) & Classification accuracy, Throughput & \cite{qin2023extremely} \\ \hline
\includegraphics[width=0.05\textwidth]{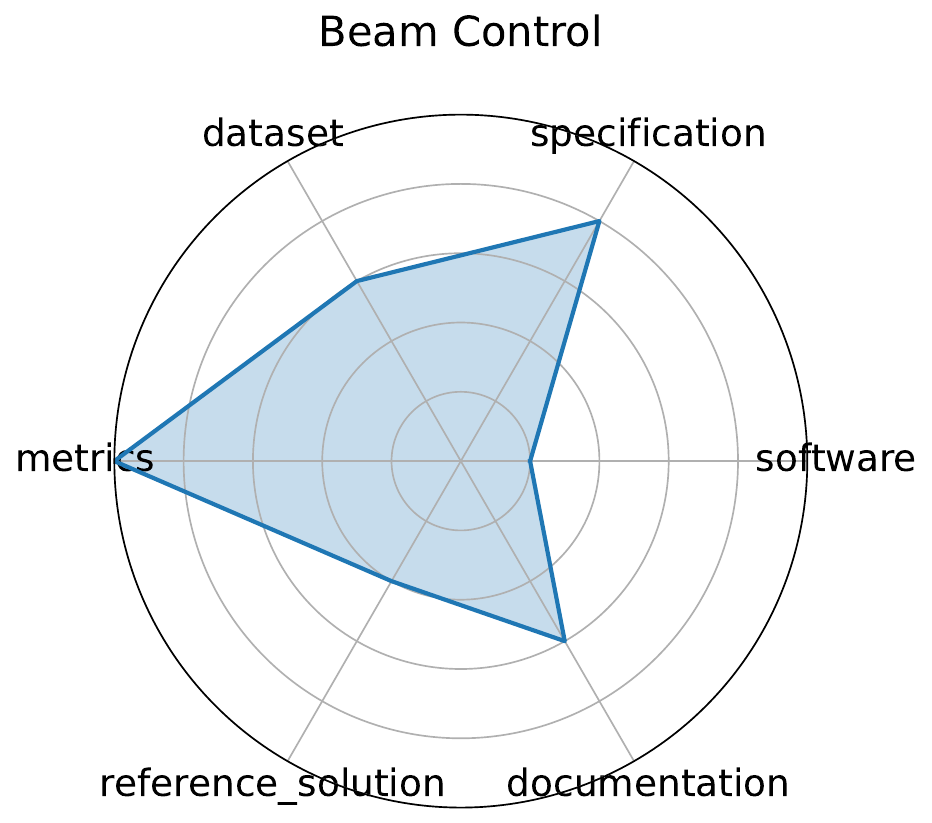} & Beam Control & High Energy Physics & DDPG, PPO (planned) & Stability, Control loss & \cite{duarte2022fastml,kafkes2021boostrdatasetacceleratorcontrol} \\ \hline
\includegraphics[width=0.05\textwidth]{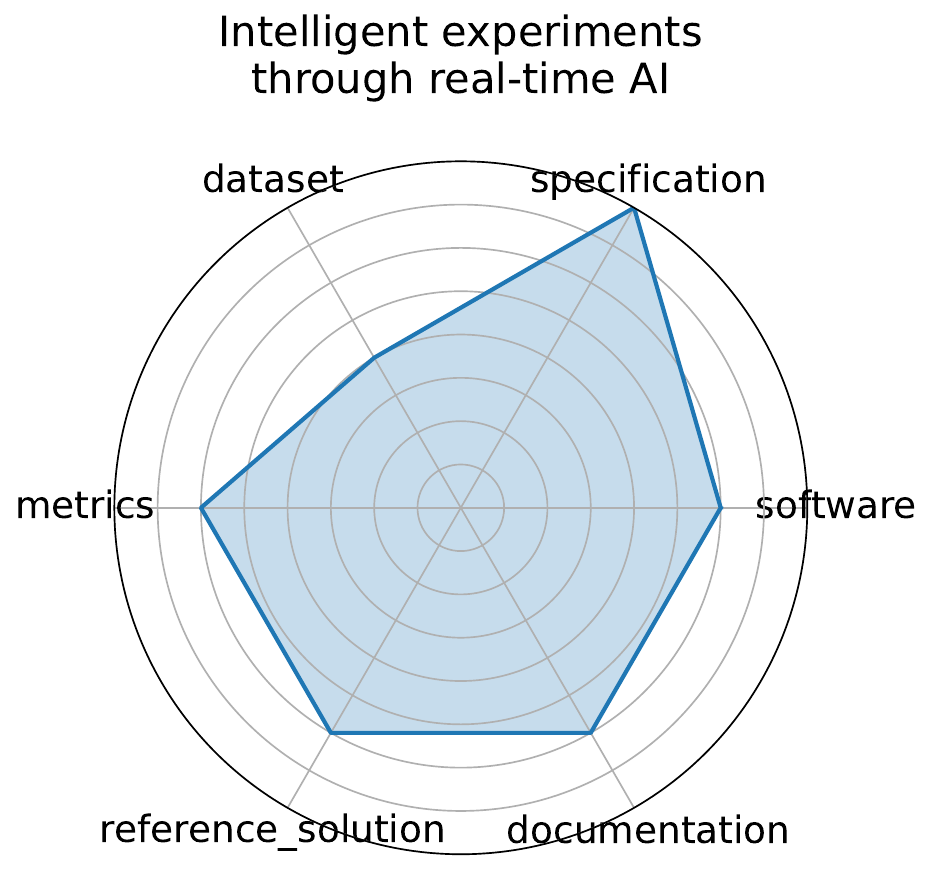} & Intelligent experiments through real-time AI & High Energy Physics & Bipartite Graph Network with Set Transformers (BGN-ST), GarNet (edge-classifier) & Accuracy (charm and beauty detection), Latency (micros), Resource utilization (LUT/FF/BRAM/DSP) & \cite{kvapil2025intelligentexperimentsrealtimeai} \\ \hline
\includegraphics[width=0.05\textwidth]{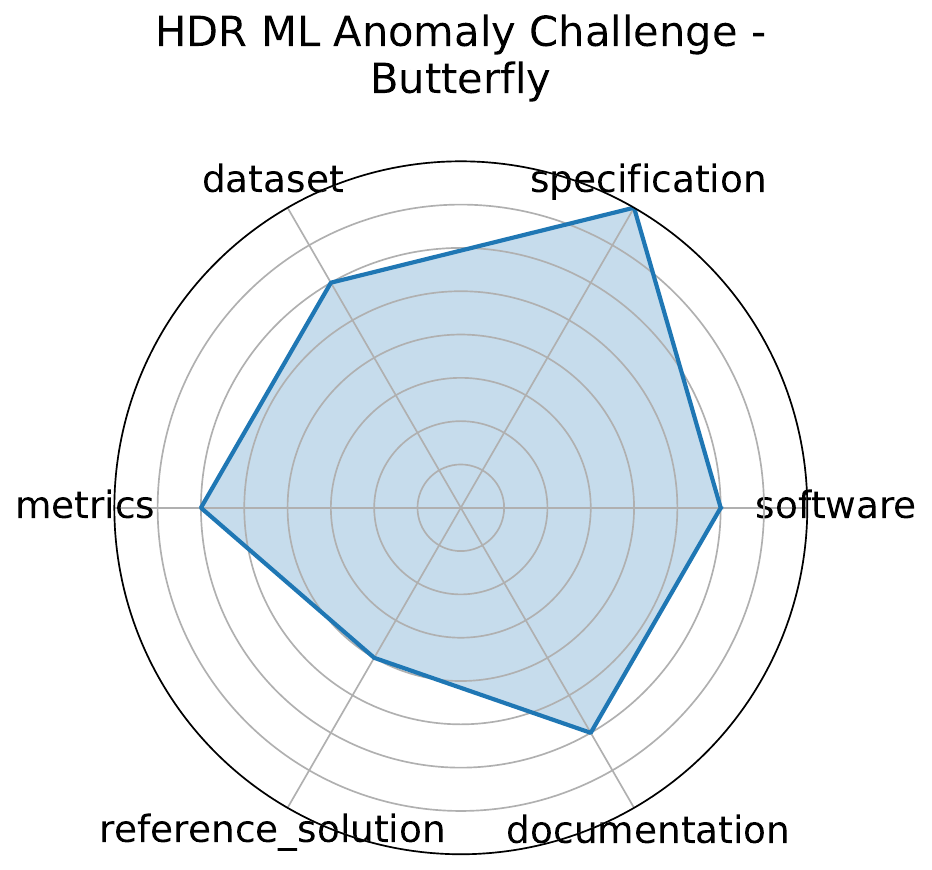} & HDR ML Anomaly Challenge - Butterfly & Biology \& Medicine & CNN-based detectors & Classification accuracy, F1 score & \cite{campolongo2025buildingmachinelearningchallenges} \\ \hline
\includegraphics[width=0.05\textwidth]{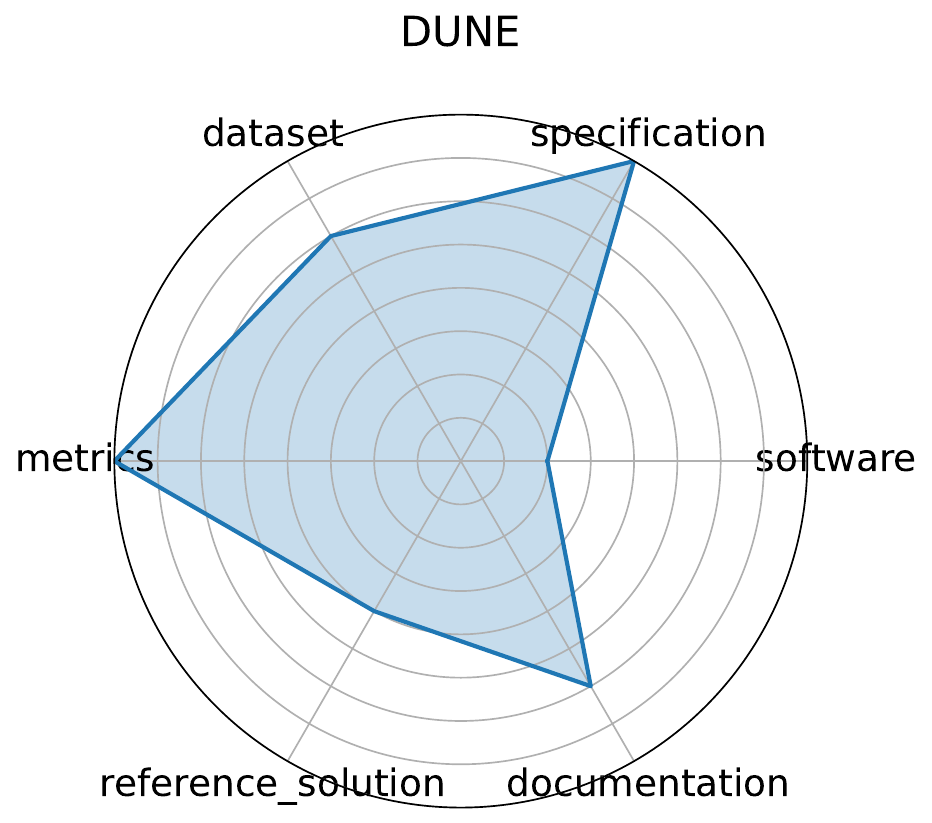} & DUNE & High Energy Physics & CNN, LSTM (planned) & Detection efficiency, Latency & \cite{abud2021deep} \\ \hline
\includegraphics[width=0.05\textwidth]{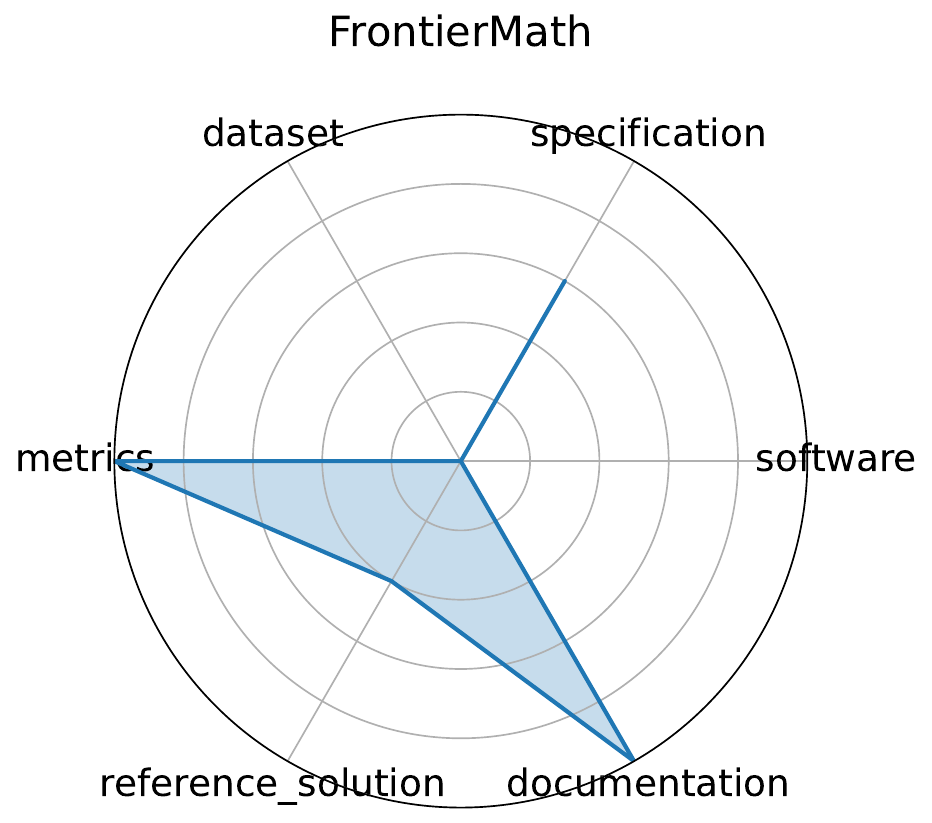} & FrontierMath & Mathematics & unknown & Accuracy & \cite{glazer2024frontiermathbenchmarkevaluatingadvanced} \\ \hline
\includegraphics[width=0.05\textwidth]{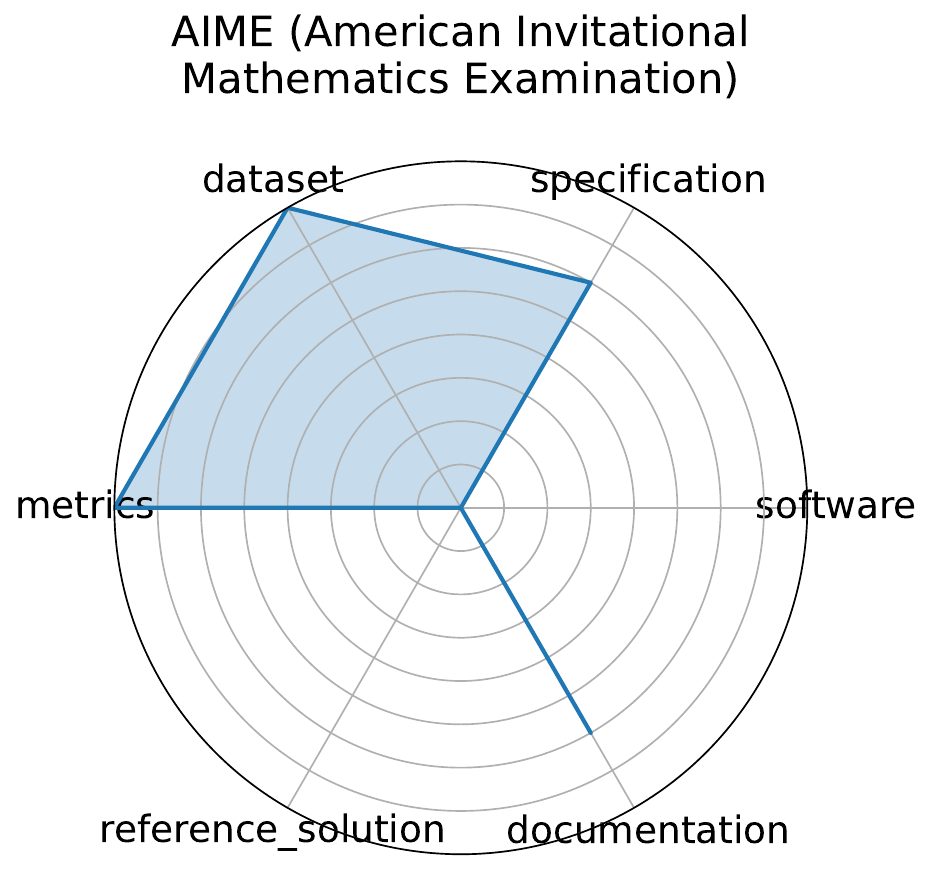} & AIME (American Invitational Mathematics Examination) & Mathematics & unknown & Accuracy & \cite{www-aime} \\ \hline
\includegraphics[width=0.05\textwidth]{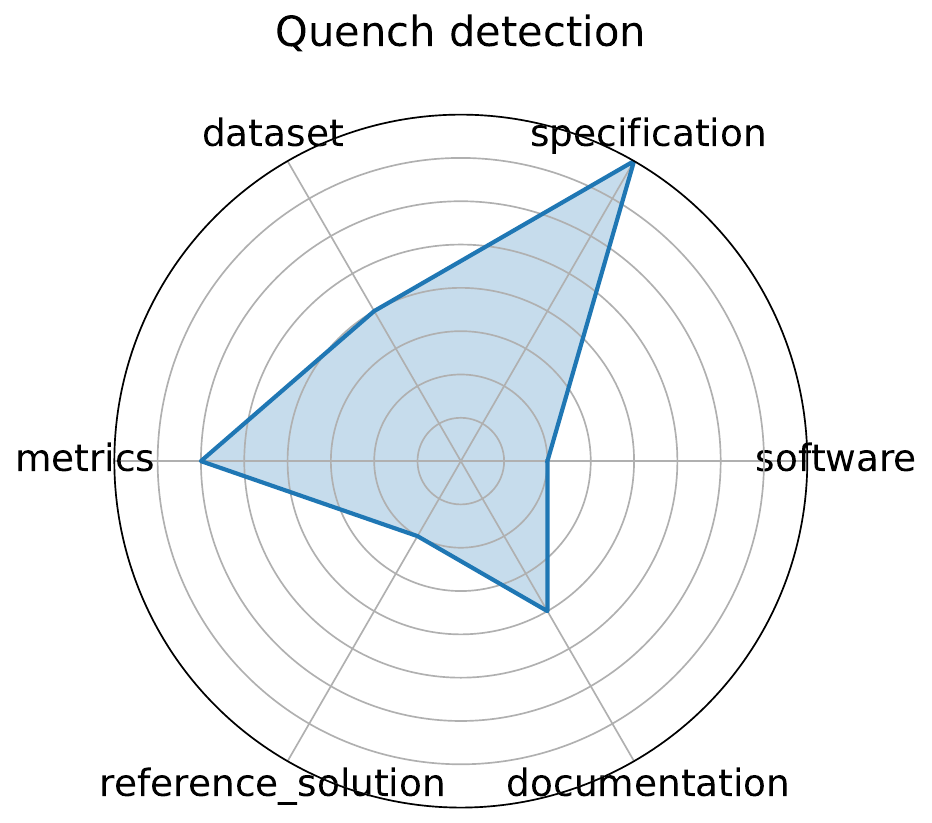} & Quench detection & High Energy Physics & Autoencoder, RL agents (in development) & ROC-AUC, Detection latency & \cite{quench2024} \\ \hline
\includegraphics[width=0.05\textwidth]{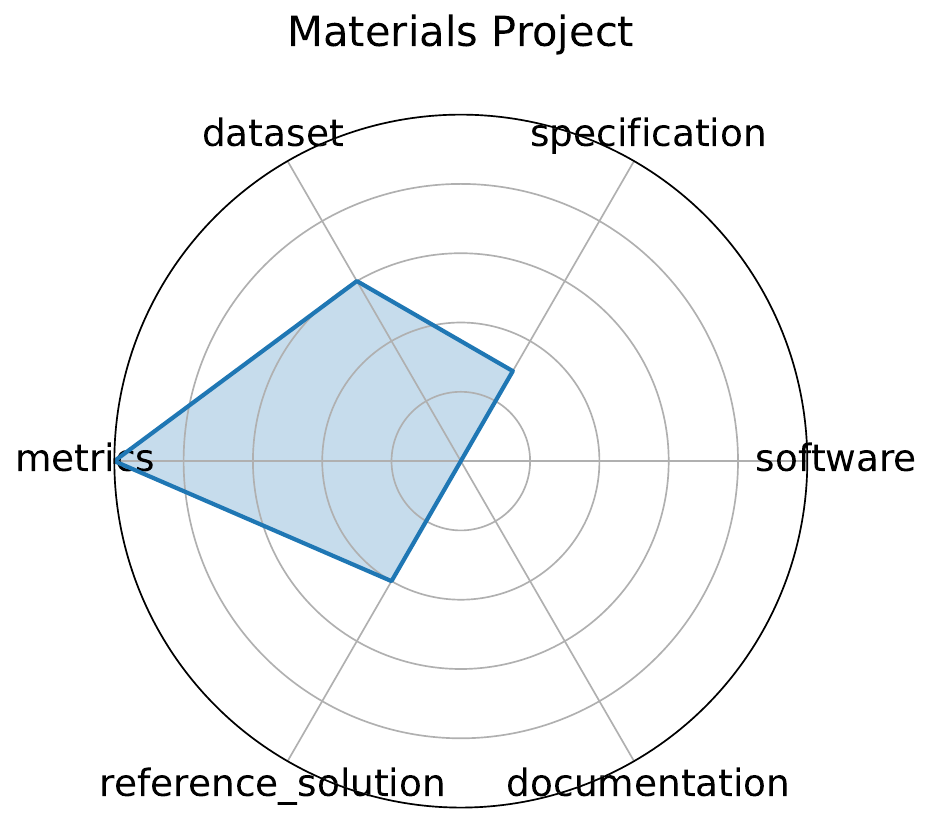} & Materials Project & Materials Science & Automatminer, Crystal Graph Neural Networks & MAE, R{\textasciicircum}2 & \cite{jain2013materials} \\ \hline
\includegraphics[width=0.05\textwidth]{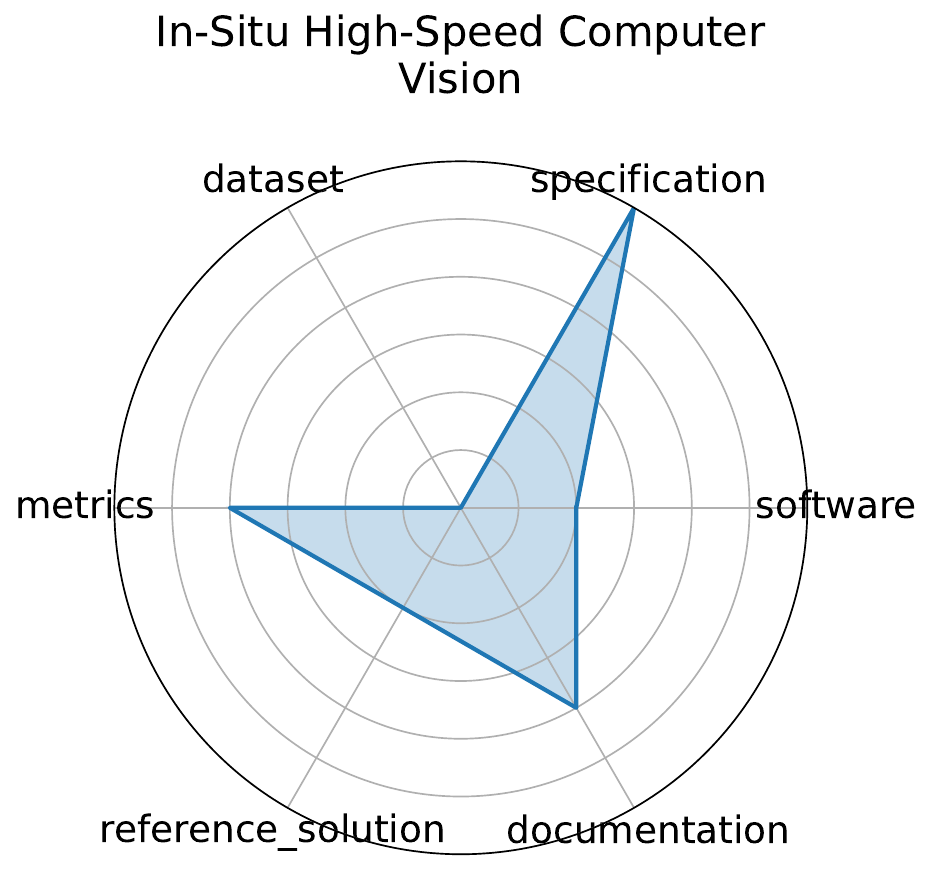} & In-Situ High-Speed Computer Vision & High Energy Physics & CNN & Accuracy, FPS & \cite{wei2024lowlatencyopticalbasedmode} \\ \hline
\end{longtable}
}

\end{landscape}
\twocolumn

%% file: 0100-profile.tex
\subsubsection{Profiling and Performance Analysis}
\label{sec:prof}

Profiling is the process of measuring a program's performance in
association with the locations in the source code in order to reveal
where resources (e.g., time and memory) are spent during execution.
Profiling is important in AI benchmarking for the following reasons:

\begin{itemize}
    \item Profiling helps explain why a particular method or
    implementation variant is faster than another.
    \item Profiling helps support fair and reproducible benchmarking.
    \item Profiling can distinguish between the essential computations
    and extraneous overheads.
    \item In a heterogeneous system, profiling can identify which
    components (e.g., CPU or GPU's CUDA cores vs. tensor cores) are
    being used by different parts of the application.
    \item Profiling can identify which specific library kernels are being used by different parts of the application.
\end{itemize}

Table~\ref{tab:merged_profiling_tools} provides a list of profiling tools that are useful for analysis of deep learning applications.

\begin{table*}[htbp]
\centering
\caption{Summary of Example Profiling Tools Useful for Deep Learning and AI Workloads}
\label{tab:merged_profiling_tools}
\renewcommand{\arraystretch}{1.2}
\begin{tabular}{|p{0.2\textwidth}|p{0.15\textwidth}|p{0.15\textwidth}|p{0.4\textwidth}|}
\hline
\rowcolor{blue!20} \textbf{Tool / Category} & \textbf{Vendor / Maintainer} & \textbf{Level / Primary Use Case} & \textbf{Key Features and Capabilities} \\
\hline
\hline
\rowcolor{gray!30} \multicolumn{4}{|l|}{\textbf{Framework Profilers}} \\ \hline
PyTorch Profiler~\cite{TensorFlow_System} & Meta & Framework-level & Records CPU/GPU activities, memory usage, and operator timings; integrates with TensorBoard and Perfetto; useful for training optimization and layer timing. \\ \hline
TensorBoard / TensorFlow Profiler~\cite{TensorFlow_System} & Google & Framework-level & Visualizes input pipelines, GPU kernels, and op-level timings; includes memory and device utilization tracing; supports bottleneck analysis. \\ \hline
torch.utils.bottleneck~\cite{PyTorch_System} & Meta & Framework-level & Combines autograd and Python profilers for quick bottleneck diagnostics. \\ \hline
JAX Profiler~\cite{JAX_Profiler} & Google & Framework-level & Works with TensorBoard to trace XLA compilation, HLO graphs, and TPU/GPU runtime performance. \\ \hline
NVIDIA DLProf~\cite{NVIDIA_DLProf} & NVIDIA & Framework-level (GPU-focused) & High-level view of deep learning layers and operations; integrates with TensorBoard DLProf plugin. \\ \hline
\rowcolor{gray!30} \multicolumn{4}{|l|}{\textbf{Hardware / System Profilers}} \\ \hline
Nsight Systems~\cite{NVIDIA_NsightSys_Doc} & NVIDIA & System-level & Timeline visualization of CPU–GPU interactions, kernel launch overheads, multi-process analysis, and NCCL tracing. \\ \hline
Nsight Compute~\cite{NVIDIA_NsightComp_Doc} & NVIDIA & Kernel-level & Detailed GPU kernel performance metrics: memory throughput, Tensor Core utilization, occupancy, and roofline analysis. \\ \hline
nvprof (deprecated)~\cite{NVIDIA:CUDA:ProfilerGuide} & NVIDIA & GPU-level & Legacy command-line CUDA profiler, replaced by Nsight tools. \\ \hline
VTune Profiler~\cite{Intel_VTune_Doc} & Intel & CPU/System-level & Hotspot analysis, vectorization, threading efficiency, and CPU performance bottlenecks. \\ \hline
omnitrace / rocprof / rocm-smi~\cite{AMD_ROCM_Doc} & AMD & GPU-level & Profiling and monitoring for AMD GPUs: kernel execution metrics, power, and temperature. \\ \hline
HPCToolkit~\cite{HPCToolkit_Paper} & Rice University & System-level (CPU+GPU) & Hierarchical performance profiling, time attribution to calling context, supports CUDA and HIP. \\ \hline
TAU~\cite{TAU_Paper} & University of Oregon & System-level (CPU+GPU+MPI) & Multi-level performance analysis, MPI integration, supports heterogeneous systems. \\ \hline
Perfetto~\cite{Perfetto_Google} & Google (Open Source) & System-level & High-resolution trace visualization, interoperable with PyTorch/TensorFlow profiler exports. \\ \hline
PAPI~\cite{PAPI_Paper} & University of Tennessee & Hardware counter interface & Provides access to CPU/GPU performance counters for integration with other profiling tools or custom instrumentation. \\ \hline
\rowcolor{gray!30} \multicolumn{4}{|l|}{\textbf{Compiler / Graph Profilers}} \\ \hline
XLA Profiler~\cite{XLA_Paper} & Google & Compiler-level (XLA) & Profiles XLA-compiled operations and execution times; supports JAX/TF and TPU/GPU workloads. \\ \hline
TorchDynamo / TorchInductor Debug Tools~\cite{TorchDynamo_TorchInductor} & Meta & Compiler-level (PyTorch 2.x) & Analyzes graph fusion, compiler optimizations, and operator performance of compiled PyTorch models. \\ \hline
Triton Profiler~\cite{Triton_Paper} & OpenAI & Kernel-level (Custom Kernels) & Reports kernel execution time, register usage, and occupancy for custom Triton GPU kernels. \\ \hline
\rowcolor{gray!30} \multicolumn{4}{|l|}{\textbf{Communication / Distributed Profilers}} \\ \hline
NCCL Profiler~\cite{NVIDIA_NCCL_Doc} & NVIDIA & Communication-level & Profiles NCCL collective communication operations (e.g., all-reduce, broadcast); timeline visualization of multi-GPU communication. \\ \hline
AWS SageMaker Debugger / Azure Profiler~\cite{AWS_SageMaker_Debugger,Azure_Profiler} & AWS / Microsoft & Cloud-level & Distributed GPU/CPU monitoring, training metric collection, and profiling at cloud scale. \\ \hline
Weights \& Biases, Comet, MLflow~\cite{Weights_Biases,Comet_ML,MLflow} & Multiple Vendors & Experiment / Cloud-level & Logs performance traces, GPU utilization, integrates with PyTorch and TensorFlow profilers for real-time monitoring. \\ \hline
\rowcolor{gray!30} \multicolumn{4}{|l|}{\textbf{System \& Memory Profilers}} \\ \hline
Torch / TensorFlow Memory Tools~\cite{Torch_Tensorflow_Memory} & Meta / Google & Framework-level (Memory) & Reports GPU memory allocation, fragmentation, and utilization trends for debugging memory bottlenecks. \\ \hline
Python Profilers (cProfile, py-spy)~\cite{Python_Profilers} & Python Community & CPU-level & Measures Python-level overhead and I/O performance; used for diagnosing data preprocessing bottlenecks. \\ \hline
\end{tabular}
\end{table*}

It is important to note that the tooling and services exist for
supporting different levels of infrastructures. This includes examples
for framework-level, system-level (including CPU and GPU), kernel-level,
compiler-level, communication-level, and cloud-level.

Furthermore, we aim here to provide comprehensive coverage of the AI
profiling stack, which affords users the insights into cross-vendor and
cross-platform capabilities and offerings, and also provide key analysis
of features of the said tools and services.

We believe it is essential to increase awareness and use of profiling
tools through AI benchmarking efforts, enabling a better understanding
of bottlenecks in AI applications. Additionally, we need to educate the
community about policy limitations that may implicitly restrict specific
profiling tools. As discussed previously, one such policy restriction is
that not all profiling information is available for energy benchmarks.
Such restrictions may also be in place for additional hardware profiling
measures.

Lastly, we need to educate the community about the {\em performance
impact} of profiling costs to avoid over-profiling. Therefore, it makes
sense that AI benchmarks should be able to choose the level of profiling
selectively. This information is vital to support the FAIR principles
and ensure that benchmarks are comparable.

%% file: 0060-review-gpu-benchmarks.tex
\subsection{GPU Benchmarking and its Variability}
\label{sec:gpu}



Modern scientific applications frequently require peta- or exascale levels of compute to model topics with high fidelity. To meet these demands in reasonable timeframes, scientists and researchers typically run these workloads on massively parallel systems such as GPUs. For example, workloads such as graph analytics~\cite{CheBeckmann2013,WangPan2017-gunrock}, scientific computing~\cite{coral2,olcf6-bmks, kim2018qmcpack, WuTaylor2019-candle}, ML~\cite{BanburyReddi2021-tinyMLPerf,BaruahShivdikar2021-gnnMark,DongKaeli2017-dnnmark,Narang2017-deepBench,MattsonCheng2019-mlperfTrain,MattsonReddi2020-mlPerf,Reddi2020mlperf-Infer,ReddiCheng2021-mlPerfVision} heavily utilize GPUs.
Increasingly, ML is also impacting scientific applications~\cite{fan2021predicting,jumper2021highly,kates2019predicting,ThiyagalingamShankar2022-mlSci,ThiyagalingamVonLaszewski2022-aiForSciMLCommons} by replacing or supplementing traditional computing methods in application domains like molecular dynamics (e.g., DeePMD~\cite{WangZhang2018-deepmd,ZengZhang2023-deepmd2}), protein folding (e.g., OpenFold2~\cite{openfold2}), and scientific AI models (e.g., AuroraGPT~\cite{Stevens2023-auroraGPT}).
However, given the scale of data these workloads operate on and the large size of the workloads themselves, they typically must partition their work across many GPUs.

Given their widespread use and trend towards many GPU applications, it is desirable from a benchmark carpentry perspective to make GPU experiments repeatable and consistent.
For traditional HPC systems composed of multiple CPUs, prior work showed that this was difficult to achieve: application performance varied by up to 20\%, even for CPUs with the same architecture and vendor SKU (Stock-Keeping Unit)~\cite{AcunLanger2016-power,chasapis2016runtime, ChasapisMoreto2019-powerEfficJobSched, InadomiPatki2015-scVar, PatelWagenhauser2020-hpcPowerConsump, SkinnerKramer2005-perfVarCauses}. This variation occurs due to the manufacturing process and the chip's power constraints~\cite{ChasapisMoreto2019-powerEfficJobSched,Scogland2015-pwrPerspectives}.
Such dynamic behavior makes it challenging for 
repeatable experiments, and can lead to resource underutilization.
Unfortunately, similar issues also arise in modern systems composed of many GPUs. Recent work has demonstrated that GPU-rich systems suffer from significant performance variability~\cite{DeBardeleben-LBNL-EuroPar13, DeSensiDeMatteis2022-cloudPerfVar, Fraternali-EEHPCVar-2018, Scogland2015-pwrPerspectives, SencanKulkarni2025-gpuHPCUtil, sinha2022notall, TopcuKarabacak2025-gpuPowerPerfVar, YouXuan2024-gvarp, ZhongSultanov2025-uncorePowerWaste}.

For example, Sinha, et al. examined variability across five modern GPU-rich clusters with a variety of sizes, cooling approaches, and GPU vendors~\cite{sinha2022notall}.
They found that applications exhibited performance variability of 8\% on average (max 22\%) with outliers up to 1.5$\times$ slower than the median GPU.
Moreover, these results were consistent over time (i.e., not transient) and were unaffected by GPU vendors or cooling type.
Interestingly, this performance variability was also application-specific: the more compute-intensive the application was, the more performance variability the application observed due to effects of the GPU's power management algorithm (e.g., Dynamic Voltage \& Frequency Scaling---DVFS).
Furthermore, performance variability is getting worse as transistors continue scaling~\cite{DRAMthermalissues}.

Although the impact of performance variability is significant for single-GPU workloads, it is even larger for multi-GPU workloads.
Currently, GPU-rich systems focus on scheduling work to minimize the number of nodes an application requests, without considering variability.
In the five clusters from this prior work, users asking for 4 GPUs for a given application would get a slower GPU allocated to them between 22\% (Sandia's Vortex cluster) and 50\% (TACC's Longhorn cluster~\cite{stanzione2020frontera, tacc}) of the time.
Thus, users are likely to get a slow GPU frequently, especially since modern scientific workloads often request 64 or more GPUs for a given experiment.
This can lead to significant resource under-utilization for multi-GPU jobs since all of them must wait for the slowest one to complete due to the bulk synchronous programming (BSP) model used in many data-parallel workloads~\cite{paszke2017-pytorch}.
Accordingly, it is imperative for users to be aware of the impact of performance variability on their experiments, and for benchmark carpentry to propose solutions to minimize its effects.

Although GPU-rich systems are likely to suffer from performance variability for the foreseeable future, there are several steps various stakeholders, such as users, maintainers, and system designers, can take to reduce the impact on obtaining statistically significant results in existing systems.
First, cluster operators can perform periodic performance-variability benchmarking to identify underperforming GPUs and perform targeted maintenance on them. Likewise, users can perform similar benchmarking to identify GPUs that behave similarly, and then use blacklisting or other scheduling approaches to attempt to schedule work on GPUs with similar performance variability profiles. However, doing so can be time- and labor-intensive for clusters with thousands or more GPUs (though it is a one-time cost, since a GPU's performance variability is consistent over time).
Thus, a more scalable, dynamic approach is to redesign job-scheduling policies for GPU clusters to account for performance variability when making scheduling decisions. Recent work has shown that embracing performance variability can transparently and significantly improve job completion time, makespan, and GPU utilization~\cite{JainTran2024-pal}.
Finally, since performance variability is application-specific, we recommend that new, unprofiled applications either rely on other applications with similar profiles as proxies ~\cite{Guerreiro-appClasses} or be profiled during their first execution on a new cluster to determine their sensitivity to performance variability.

In terms of democratizing the availability of multi-GPU systems there are several barriers to overcome, these are the cost, access, skills and complexity The cost barrier means that the large-scale systems are affordable only to national labs and major corporations. Consequently, the access is usually restricted to the staff of these organizations. Using multi‑GPU systems effectively requires specialized knowledge. Users must be trained in containerization technologies, distributed libraries, and orchestration tools that allow applications to scale across many GPUs. There is also a barrier on the conceptual level. The performance of a multi‑GPU system is the result of interactions between hardware, interconnects, and software stacks. At present, we lack high‑level performance prediction model that can reliably describe how applications behave when running on GPUs. This makes it difficult to plan experiments, determine the required resources and generalize findings. 


%% file: 0070-energy-benchmarks.tex
\subsection{Energy Benchmarking}
\label{sec:energy}

Energy consumption is a critical component of  ML benchmarking. Training and inference with modern AI systems can require enormous computational resources.

\begin{figure}[tb]
    \centering
    \includegraphics[width=1.0\linewidth]{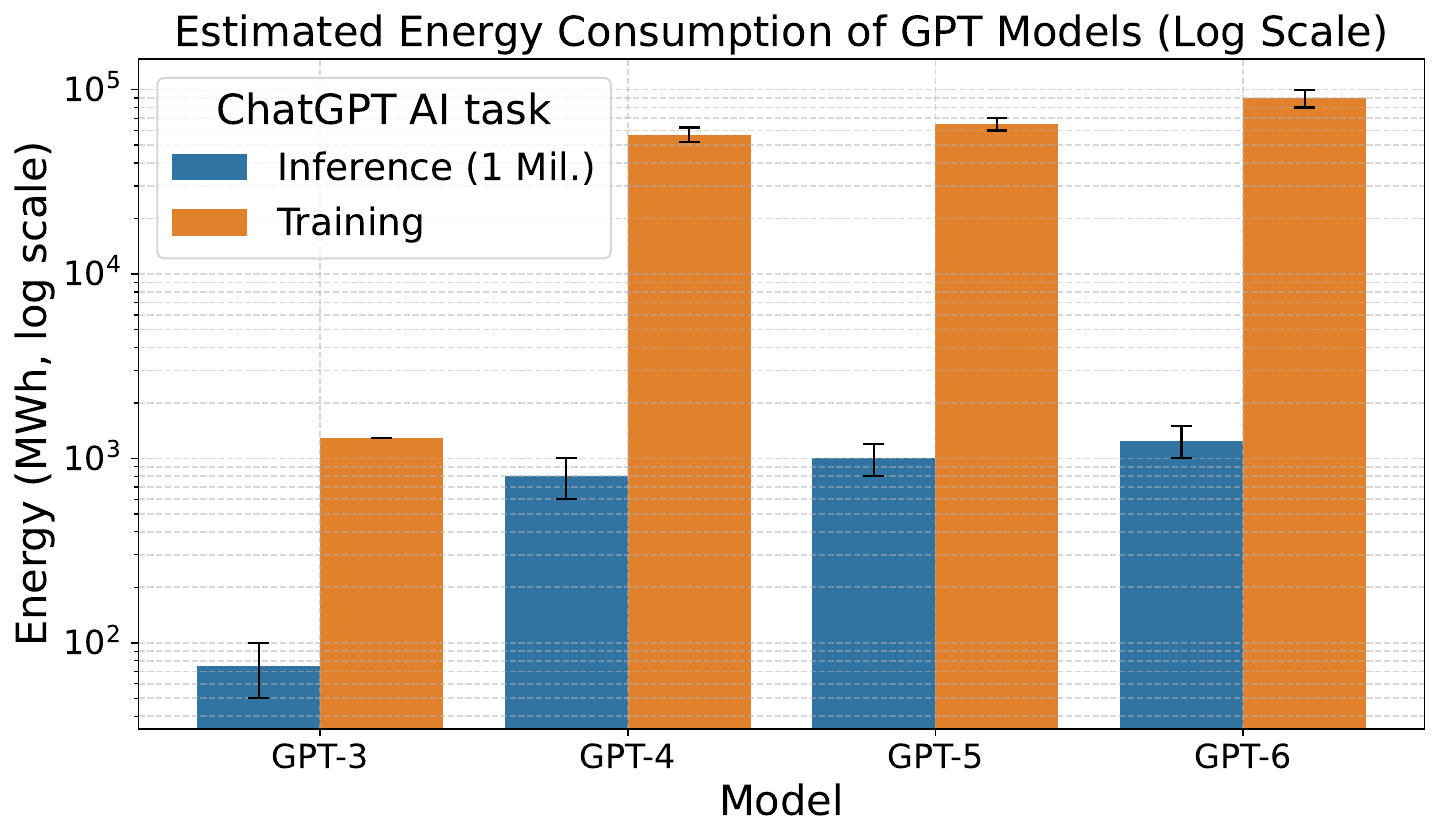}
    \caption{Energy Consumption for ChatGPT Training and Inferencing 1 Million Queries \TODO{(Data for GPT-5 and 6 are estimates).} }
    \label{fig:energy-consumption}
\end{figure}

To illustrate the issue, we have provided in Table \ref{tab:energy-chatgpt} and Figure \ref{fig:energy-consumption} the energy required to train various ChatGPT models (some of which are estimated as no public data has been released \cite{sciencefeedback2024energy,kaplan2020scaling}, such as GPT-5 and GPT-6). The training of a single large-scale language model (GPT-3) consumes approximately 1,287 MWh placing it in the same range as the annual energy usage of about 130 U.S. households, according to U.S. Energy Information Administration (EIA) statistics on average residential electricity consumption \cite{EIA_Electricity_Price_2025,eia2024residential,WECEnergy,patterson2021carbon,jegham2025hungryaibenchmarkingenergy,baeldung2023energy}. 

For the U.S. Department of Energy (DOE) leadership-class machines, such as those hosted at Oak Ridge National Laboratory (see Table \ref{tab:ornl-energy}), we find documented and significant progress toward exascale, but at the cost of increased energy consumption that more than doubled during the last generational upgrade. However, the Peak Performance per energy unit has increased significantly, and compared to Jaguar's initial values, Frontier has improved by a factor of 209, thus becoming relatively more efficient despite overall energy consumption needs.

\begin{table}[tb]
\centering
\caption{Estimated Energy Consumption of GPT Models for Training and Inference 
(Based on \cite{brown2020language, patterson2021carbon, medium2023gpt4carbon, 
extremenetworks2023energy, epochai2024compute, hackernoon2024dirtysecret}).}
\label{tab:energy-chatgpt}
\resizebox{\columnwidth}{!}{
\begin{tabular}{|l|c|c|}
\hline
\textbf{Model} & \textbf{\makecell{Training Energy\\(MWh)}} & \textbf{\makecell{Inference Energy\\(per 1M queries, MWh)}} \\
\hline
\hline
GPT-3 & $\sim$1,287 \cite{baeldung2023energy,patterson2021carbon} & $\sim$50--100 \\
\hline
GPT-4 & 51,773--62,319 \cite{medium2023gpt4carbon,extremenetworks2023energy} & $\sim$600--1,000 \\ \hline 
GPT-5 & $>$60,000 (estimated) \cite{epochai2024compute,hackernoon2024dirtysecret} & $\sim$800--1,200 \\ \hline
GPT-6 & 80,000--100,000 (projected) \cite{epochai2024compute} & $\sim$1,000--1,500 \\ \hline
\end{tabular}
}
\end{table}

\begin{table}[tb]

    \caption{Evolution of the Leadership Class Supercomputer at Oak Ridge National Laboratory}
    \label{tab:ornl-energy}
    
    \resizebox{\columnwidth}{!}{
        \begin{tabular}{|l|c|l|r|r|r|r|r|}
            \hline
            \textbf{Machine} & \textbf{Year} & \textbf{Architecture} & $R_{max}$ Scaling & \textbf{\makecell{$R_{max}$ \\ PFlops/s}} & \textbf{\makecell{$R_{peak}$ \\ PFlops/s}} & \textbf{\makecell{Power\\ (MW)}} & \textbf{\makecell{$R_{max}/$Power\\(PF/MW)}} \\
            \hline
            \hline
            Jaguar\cite{BlandRogers2009-jaguar}       & 2009 & Multi-core CPU  & 1 &    1941 &    2628 & 7  & 277.29 \\
            \hline
            Titan\cite{Bland2012-titan}               & 2012 & Hybrid CPU/GPU  & 9.06 &   17590 &   27113 & 9  & 1954.44  \\
            \hline
            Summit\cite{WombleShankar2019-summit}     & 2017 & Hybrid CPU/GPU  & 76.6 &  148600 &  200795 & 13  & 11430.77  \\
            \hline
            Frontier\cite{AtchleyZimmer2023-frontier} & 2022 & Hybrid CPU/GPU  & 697.1 & 1353000 & 2055717 & 29  & 46655.17  \\
            \hline
        \end{tabular}
    }

    \smallskip
    {\tiny~~ $^*$PF=\emph{Theoretical peta–floating-point operations per second}; $1\;\text{PF}=10^{15}\text{FLOPS}.$
    
    $R_{max}$ = maximal LINPACK performance achieved.
    $R_{peak}$ = theoretical peak performance.
    }
\end{table}

Carbon-emission measurements also help provide a more detailed understanding of associated energy impacts.

If we only focus on traditional benchmarks using metrics such as FLOPS or latency, we provide performance insights but overlook {\em energy-to-solution}, which measures the total energy required to complete a task. Without perspective, researchers and practitioners focus on optimizing for speed at the expense of sustainability and cost efficiency. 

Thus, we believe it is important to make energy benchmarks an important aspect of AI benchmarks. Energy benchmarking ought to address the following:

\begin{itemize}
     \item Quantify the environmental footprint of AI workloads (carbon emissions, renewable vs. non-renewable energy use).
     \item Highlight economic tradeoffs in large-scale computing (cloud costs, datacenter efficiency). 
     \item Guide hardware and algorithmic choices towards a more effective architecture.
     \item Support policy and funding decisions by providing transparent data on sustainability. 
 \end{itemize}

Energy-aware benchmarks help ensure that AI development aligns with broader goals of responsible computing, making results reproducible, performant, and economically and environmentally sustainable. 

Thus, we see several opportunities. First, we need to make energy benchmarks more prominent and provide materials and tutorials as part of AI benchmark carpentry to educate the community. Second, we must ensure that not only the most expensive hardware, such as leadership-class and hyper-scale data centers, is used, but also medium- and even small-scale hardware, so that democratizing energy benchmarks within the community is easy to implement. This way, measurements of even smaller AI-based scientific applications can integrate energy consumption into their benchmarks, and meaningful comparisons with traditional algorithms that do not use AI can be drawn. Third, we must ensure that energy metrics and logs can be accessed and uniformly integrated into the AI benchmarks.

\subsubsection{AI Energy Benchmark Carpentry}

To support AI energy benchmark carpentry efforts, we need to address the following issues:

\begin{itemize}
\item Conduct a relevant survey of existing efforts
\item Identify metrics useful for AI benchmarks 
\item Identify how to leverage existing and create new leaderboards focusing on energy metrics
\item Identify simple-to-use blueprints as part of the carpentry efforts that can not only be replicated and reused, but also serve as a basis for newly developed benchmarks.
\item Conduct community outreach to offer carpentry tutorials that focus on AI benchmarks instead of just AI software and services.
\item Identify how to obtain and integrate meaningful and practical metrics (e.g., data centers may not provide uniform access to energy data) so that energy data collection and access become part of carpentry efforts.
\end{itemize}

Strategies to integrate energy into AI benchmarks for carpentry efforts include improving access to metrics, including the creation of logs during runtime that:

\begin{itemize}
    \item Log ambient temperature and humidity.
    \item Log sample power at regular intervals or averages over the run.
    \item Store the logging data in an easy-to-parse format (CSV, JSON, YAML) 
    \item Upload results as artifacts in support of the FAIR principle and make available for comparison.
\end{itemize}

Next, we discuss some of the aspects that need to be addressed in more detail.

\subsubsection{Energy Metrics}

There are various energy metrics to consider, including metrics that may not historically received attention. It is also important to identify metrics for leaderboards, but they must be obtained in a way that allows fair, informed comparisons. Hence, it is important to document how the experiment should be conducted rather than just referring to the metric. In principle, blueprints should be used and adapted to make comparisons across hardware and software easier. Energy metrics are used across different layers of the AI benchmark infrastructure, which is similar to classical HPC infrastructure. We provide an example of using different metrics on the various layers in Figure \ref{fig:energy-metric-layer}. Such diagrams should be integrated into the blueprints provided to users to simplify understanding the benchmarks' energy scope.

\input{graph-energy-layered.tex}

As part of the energy augmentation, a clear purpose for the benchmark metric should be stated. 
Such examples should be collected as part of the experiment's metadata so they can be leveraged and serve as a motivator for other benchmarks. In our example from Figure \ref {fig:energy-metric-layer}, the purpose for each metric is as follows:

\begin{enumerate}

\item {\bf Device/Micro-architectural Layer (\mmD})
\begin{itemize}
        \item {\em Energy per flop} or {\em Energy per inference}: Measures the energy consumed to perform a single computational operation (a floating-point operation or an inference).
        \item {\em Temperature sensors}: {\em Related Logging (Non-KPI):} {\em Inlet and Outlet Temperature Sensors}: Logged because {\em thermal headroom} directly bounds the safe {\em Dynamic Voltage and Frequency Scaling (DVFS)} ranges.
\end{itemize}
\item {\bf Job/System Layer (\mmJ)}
\begin{itemize}

        \item {\em Kilowatt-hour (kWh)}: The total energy consumed by a specific job or set of jobs over its duration.
        \item {\em Energy--Delay Product (EDP)}: A combined metric of energy and time (energy $\times$ delay) used to assess the overall efficiency of a computation. Lower EDP generally indicates better performance and efficiency.
\end{itemize}

\item {\bf Facilities/Data Center Layer (\mmF)}
\begin{itemize}

        \item {\em Power Usage Effectiveness (PUE)}: A ratio that measures how efficiently a data center uses energy. An ideal PUE is 1.0 (meaning the IT equipment uses all energy).
        {\em Data Center Infrastructure Efficiency (DCiE)}: The reciprocal of PUE, expressed as a percentage. It shows the percentage of total data center energy used by IT equipment.
\end{itemize}

\end{enumerate}

This tiered structure, along with a detailed purpose statement for each metric, allows for meaningful comparisons and decision-making at every level of the computing infrastructure.

To identify commonly used metrics, we conducted an initial survey of tools and benchmarks related to energy, which we present in 
Table \ref{tab:hpc_energy_catalog}, while listing their typical benchmark use.

Common requirements for such metrics include obtaining measurements at low cost, sharing results with metadata augmentations, and integrating them into potential leaderboards. 
We believe we have to go beyond established leaderboards such as {\em Green500} and the {\em MLPerf Power}, which already influence processor road-maps and procurement calls~\cite{Scogland11Green500,Tschand24MLPerfPower}, to raise awareness of the energy impact on real-world scientific applications.

\renewcommand{\arraystretch}{1.1}
\begin{table*}[hptb]
  \centering
  \caption{Energy- or Carbon-Efficiency (B)enchmarks and (T)ools used in Scientific-HPC research.}
  \label{tab:hpc_energy_catalog}

  \resizebox{\textwidth}{!}{%
  \begin{tabular}{|lllll|}
  \hline
\rowcolor{blue!20} 
    & \headerfont\textbf{(B)enchmark or (T)ool}
    & 
    & \headerfont\textbf{Core metric(s) } 
    & \headerfont\textbf{Typical Benchmarking Use}\\ 
\hline
\hline
\rowcolor{gray!20} \multicolumn{5}{|l|}{Benchmark}\\ \hline
\mmB & SPECpower\_ssj2008           & \cite{specpower}            & W/transaction; ops/W                & Enterprise-server rankings; ENERGY STAR compliance \\ \hline
\mmB & SPEC\,SERT$^{2}$             & \cite{sert2}                & Server-Efficiency-Rating = kWh + perf    & EU Lot 9 certification; vendor datasheets \\ \hline
\mmB & TPC-Energy                   & \cite{tpcenergy}            & Wh/DB phase                            & OLTP/warehouse energy cost studies \\ \hline
\mmB & JouleSort                    & \cite{joulesort}            & records/J                              & Storage-I/O contests; I/O-stack tuning \\ \hline
\mmB & Green500                     & \cite{green500}             & GFLOPS/W (HPL or HPL-AI)               & Global supercomputer energy ranking \\ \hline
\mmB & HPCG-Power                   & \cite{hpcgpower}            & GFLOPS/W (HPCG)                        & Memory-bound tuning; procurement add-on to TOP500 \\ \hline
\mmB & HPL-MxP (HPL-AI)             & \cite{hplmxphplai}          & mixed-precision GFLOPS/W               & GPU/TPU evaluation for AI-optimised LINPACK \\ \hline
\mmB & MLPerf Power                 & \cite{mlperfpower}          & J; avg W; J/sample; J/epoch       & Official energy track for MLPerf submissions \\ \hline
\mmB & MLPerf Tiny                  & \cite{mlperftiny}           & $\mu$J/inference (MCU)             & Edge-AI board comparison; ultra-low-power design \\ \hline
\mmB & CoreMark-PRO Power           & \cite{coremarkpro}          & iterations/s/W (SoC)                 & Pre-silicon DVFS sweeps; embedded RFPs \\ \hline 
\mmB & UL Procyon AI Power          & \cite{procyon}              & images/W; fps/W                     & Smartphone \& laptop AI-inference benchmarks \\ \hline
\mmB & CANDLE Power Study           & \cite{candlepowerstud}      & J/epoch; GFLOPS/W                   & DOE accelerator procurement guidance \\ \hline
\mmB & LULESH/miniFE Energy       & \cite{luleshminifeene}      & J/iteration                            & DVFS + autotuning baselines \\ \hline
\mmB & ExaSMR Power Benchmark       & \cite{exasmrpowerbenc}      & J/neutron; energy-vs-accuracy curve   & Energy budget strategy in nuclear simulations \\ \hline
\mmB & EE-HPC-WG Energy Benchmark   & \cite{eehpcwgenergybe}      & draft node/job spec; JSON trace       & Toward common HPC energy standard \\ \hline
\mmB & HPC-AI500 Energy Track       & \cite{hpcai500energyt}      & planned: GFLOPS/W; tokens/J         & Mixed AI/HPC cluster evaluations \\ \hline
\mmB & PARSEC-3.1 Energy Extension  & \cite{parsec31energye}      & W; J via PAPI-RAPL; J/op; EDP       & Pre-silicon DVFS research \\ \hline
\mmB & CosmoFlow-Power              & \cite{cosmoflow2019}        & J/epoch; GFLOPS/W                   & CNN scaling on 15 k+ GPUs \\ \hline
\mmB & HACC Energy Add-on           & \cite{hacc2020power}        & J/particle update                      & N-body cosmology power studies \\ \hline
\mmB & DeepCAM-Energy               & \cite{deepcam2020power}     & J/epoch (UNet)                         & Climate-analytics accelerator studies \\ \hline
\mmB & OpenIFS-Energy               & \cite{openifsenergy2023}    & kWh/model-day; W timeline             & Weather-model node comparison \\ \hline
\mmB & GROMACS-EE                   & \cite{gromacsee2024}        & J/ns; W/GPU                         & MD clock-vs-accuracy trade-offs \\ \hline
\mmB & NAMD-Power                   & \cite{namdpower2019}        & Energy-Delay-Product (ApoA1)             & Summit node DVFS optimisation \\ \hline
\mmB & QE Energy Suite              & \cite{qeenergy2022}         & J/SCF step; GFLOPS/W                & DFT GPU-offload studies \\ \hline
\mmB & VASP-Power Harness           & \cite{vasppower2023}        & W; kWh/MD step                        & Materials-science accelerator compare \\ \hline
\mmB & OpenFOAM-Energy              & \cite{openfoamenergy2021}   & J/1k iterations                       & CFD partitioning \& mesh tuning \\ \hline
\mmB & InSAR-AI Power Kit           & \cite{insarpower2024}       & J/satellite scene                      & Edge-to-cloud EO inference cost \\ \hline
\mmB & H3D-Energy                   & \cite{h3denergy2023}        & J/hydrology timestep                & Hydrology model DVFS exploration \\ \hline \hline\hline  
\rowcolor{gray!20} \multicolumn{5}{|l|}{Tool}\\ \hline
\mmT & PTDaemon/SERT Energy       & \cite{specptdaemonser}      & calibrated W; kWh (node)                & Lab reproducibility; Lot 9 labels \\ \hline
\mmT & Scaphandre                   & \cite{scaphandre}           & W; kWh (process/node, Prometheus)     & Slurm dashboards; power-cap feedback \\ \hline
\mmT & Kepler                       & \cite{kepler}               & W/pod; J/pod (eBPF)                 & Energy observability in K8s clusters \\ \hline
\mmT & CodeCarbon                   & \cite{codecarbon}           & kWh; kg CO\(_2\)e (process)             & Rapid CO\(_2\) estimation in pipelines \\ \hline
\mmT & CarbonTracker                & \cite{carbontracker}        & measured + predicted kWh; CO\(_2\)e     & Scheduling DL jobs in low-carbon hours \\ \hline
\mmT & PowerPACK/Mont-Blanc       & \cite{powerpackmontbl}      & W; J for MPI/OpenMP mini-apps         & Network-topology \& DVFS studies \\ \hline
\mmT & Cray PAT Energy Counters     & \cite{craypatenergyco}      & J/function; avg W                     & Kernel hotspot hunting on Shasta \\ \hline
\mmT & IBM PowerAPI (pmlib)         & \cite{ibmpowerapipmli}      & kWh (job/process)                      & Energy-aware scheduling on Summit \\ \hline
\mmT & NVIDIA DCGM Energy           & \cite{nvidiadcgmenerg}      & W; J (GPU) \@ 1Hz; telemetry           & GPU power-cap discovery; Green500 \\ \hline
\mmT & Intel VTune Power            & \cite{intelvtunepower}      & package W; J/function                 & Roofline-vs-energy tuning on Xeon \\ \hline 
\mmT & Cloudmesh GPU & \cite{cloudmesh-gpu} & Power Draw; Temperature & Temperature and energy frequency traces \\ \hline
  \end{tabular}
  }
\end{table*}

\subsubsection{Leveraging Previous Work}

As we can see from the table, a large number of tools and benchmarks exist, and we can leverage them to work towards a FAIR-based approach on energy benchmarks. This is all the more important when developing concise carpentry and democratization efforts. The distinction in the layered architecture for energy benchmarks also helps, as it is often not possible or desirable to address all layers at once. It is evident that energy benchmarking, in itself, is a complex research topic, and that carpentry efforts must be established to bring this knowledge forward and enhance AI benchmarks into AI energy benchmarks.

%% file: graph-energy-layered.tex
\newcommand{\mmB}{B}
\newcommand{\mmT}{T}
\newcommand{\mmJ}{$J_L$}
\newcommand{\mmD}{$D_L$}
\newcommand{\mmF}{$F_L$}

\begin{figure}[!ht]
\begin{center}
\resizebox{1.0\columnwidth}{!}{
\begin{tikzpicture}[
    box/.style={rectangle, rounded corners, draw=black, very thick, text width=6cm, align=center, minimum height=1.5cm, fill=white!90!gray},
    layer/.style={rectangle, rounded corners, draw=blue!50!black, very thick, text width=8cm, align=center, minimum height=1cm, fill=blue!10},
    metric/.style={rectangle, draw=red!50!black, thick, text width=4cm, align=center, minimum height=0.8cm, fill=red!10},
    labelbox/.style={rectangle, draw=green!50!black, thick, text width=3cm, align=center, minimum height=0.6cm, fill=green!10},
    arrow/.style={-Latex, thick}
]

\node (facility_layer) at (0, 0) [layer] {\textbf{3. Facility/Data Center Layer (\mmF)}};
\node at (0, -0.7) {\footnotesize \emph{Focus: CSR Reporting, Infrastructure Management}};

\node (pue) at (-4, -2) [metric] {\textbf{PUE} \\ (Power Usage Effectiveness)};
\node (dcie) at (4, -2) [metric] {\textbf{DCiE} \\ (Data Center Infrastructure Efficiency)};

\node (pue_eq) at (-4, -3.5) [box, text width=5cm, minimum height=1.2cm] {$$ \text{PUE} = \frac{\text{Total Facility Energy}}{\text{IT Equipment Energy}} $$};
\node (dcie_eq) at (4, -3.5) [box, text width=3.3cm, minimum height=1.2cm] {$$ \text{DCiE} = \frac{1}{\text{PUE}} \times 100\% $$};

\draw [arrow] (facility_layer.south) -- ++(0, -0.5) coordinate (split1);
\draw [arrow] (split1) -- (pue);
\draw [arrow] (split1) -- (dcie);
\draw [arrow] (pue) -- (pue_eq);
\draw [arrow] (dcie) -- (dcie_eq);

\node (job_layer) at (0, -6) [layer] {\textbf{2. Job/System Layer (\mmJ)}};
\node at (0, -6.7) {\footnotesize \emph{Focus: Power-Cap Schedulers, System Optimization}};

\node (kwh) at (-4, -8) [metric] {\textbf{kWh} \\ (Kilowatt-hour)};
\node (edp) at (4, -8) [metric] {\textbf{Energy–Delay Product} (EDP)};

\node (edp_def) at (4, -9.5) [box, text width=3.5cm, minimum height=1.2cm] {$$ \text{EDP} \propto \text{Energy} \times \text{Delay} $$};

\draw [arrow] (job_layer.south) -- ++(0, -0.5) coordinate (split2);
\draw [arrow] (split2) -- (kwh);
\draw [arrow] (split2) -- (edp);
\draw [arrow] (edp) -- (edp_def);

\node (device_layer) at (0, -12) [layer] {\textbf{1. Device/Micro-architectural Layer (\mmD)}};
\node at (0, -12.7) {\footnotesize \emph{Focus: Micro-architectural Hot-spots, Hardware Limits}};

\node (eperop) at (-4, -14) [metric] {\textbf{Energy per flop} or \\ \textbf{Energy per inference}};
\node (temp) at (4, -13.7) [labelbox] {\textbf{Temperature~~Sensors~} (\emph{Non-KPI})~};

\node (eperop_def) at (-4, -15.5) [box, text width=3.5cm, minimum height=1.2cm] {Energy per single \\ computational unit};

\node (temp_reason) at (4, -15.5) [box, text width=3.5cm, minimum height=1.5cm] {Logged because \textbf{thermal headroom} bounds safe \textbf{DVFS} ranges.};

\draw [arrow] (device_layer.south) -- ++(0, -0.5) coordinate (split3);
\draw [arrow] (split3) -- (eperop);
\draw [arrow] (split3) -- (temp);
\draw [arrow] (eperop) -- (eperop_def);
\draw [arrow] (temp) -- (temp_reason);

\draw [arrow, line width=2pt, dashed, gray] (job_layer.north) -- (facility_layer.south);
\draw [arrow, line width=2pt, dashed, gray] (device_layer.north) -- (job_layer.south);

\node at (0, -5) [align=center] {\textbf{Drives Schedulers} \\ $\uparrow$};
\node at (0, -11) [align=center] {\textbf{Informs System Config} \\ $\uparrow$};

\end{tikzpicture}
}
\end{center}
\caption{Illustration of an Example for Metrics as Used in the Layered System Architecture for Large-Scale AI Benchmarking.}
\label{fig:energy-metric-layer}
\end{figure}
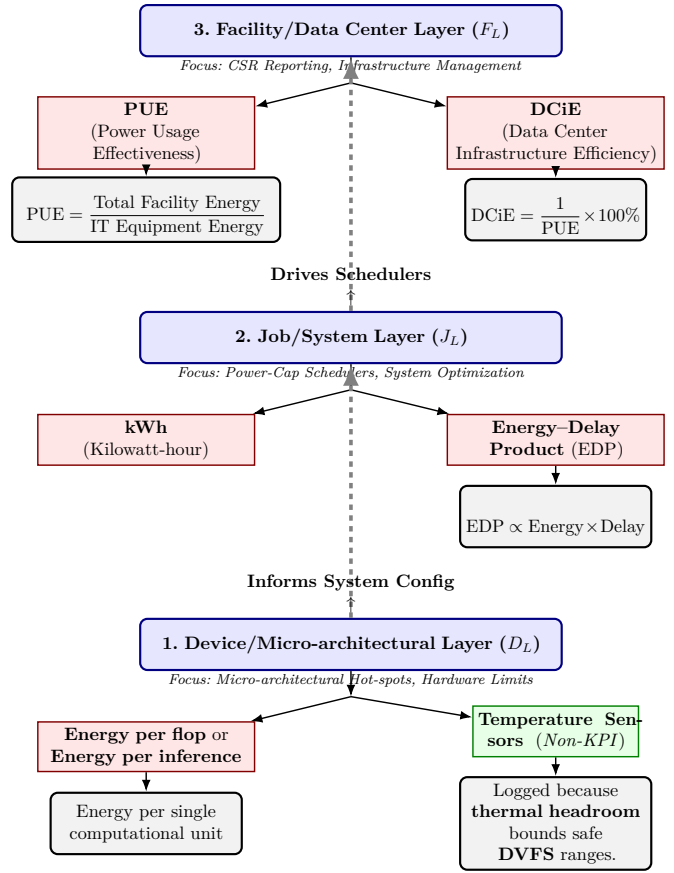

%% file: 0110-sim.tex
%
%

\subsection{Simulation as a Tool to Benefit AI Benchmark Carpentry and Democratization}
\label{sec:sim}


Simulating AI hardware and software infrastructures offers an opportunity to democratize AI benchmarking and impact AI development. 
This is especially useful for those (a) without direct access to the hardware on which the AI benchmarks run, and therefore can use simulations to estimate its behavior; and (b) planning large-scale experiments, who can use simulations to assess the impact on real hardware and infrastructure.

As part of this, recent work in the modeling and simulation community has significantly expanded users' options for studying how their ML workload optimizations affect them.
Although there is a wide array of tools that can be used, we focus on four of the most popular, widely used tools: Accel-Sim~\cite{KhairyShen2021-accelSim}, gem5~\cite{binkert2011gem5, LowePowerAhmad2020-gem520}, SST~\cite{RodriguesHemmert2011-sst, SST}, and Digital Twins~\cite{brewer2024digital}.
These tools are often used in academia, industry, and national labs because they enable high-fidelity, early-stage design exploration.
Moreover, they enable users who do not have access to real hardware or are prototyping optimizations for hardware that does not yet exist to simulate the behavior of popular ML workloads while balancing performance and power trade-offs.

\noindent
\textbf{Accel-Sim}~\cite{KhairyShen2021-accelSim}: For users interested in simulating ML workloads on modern NVIDIA (Volta through Blackwell) GPUs, Accel-Sim offers a great combination of high fidelity and usability.
Accel-Sim builds upon the popular GPGPU-Sim~\cite{BakhodaYuan2009-gpgpuSim}, and has an integrated power model~\cite{KandiahPeverelle2021-accelWattch}.
This allows users to examine power and performance tradeoffs for ML workloads.

Currently, Accel-Sim supports running ML workloads in three formats: (1) direct CUDA source code, (2) CUDA programs with library calls where the library includes the PTX for the library calls (only for CUDA 8.1 and earlier~\cite{LewShah2018-gpgpusimML}), (3) and direct SASS (NVIDIA's machine assembly language) execution.
As NVIDIA's libraries (e.g., cuDNN, cuBLAS) grow increasingly complex, and software like PyTorch add additional complexity on top of these libraries, the third option is the most popular as it can trace through multiple layers of software (e.g., PyTorch, cuBLAS).
Moreover, to make the simulator's runtime more tractable, recent work has demonstrated how to identify and simulate a representative subset of a given workload without significantly compromising accuracy~\cite{AvalosKhairy2021-pka}.
Thus, Accel-Sim is widely used by users who want to improve the efficiency of a given GPU.
However, since Accel-Sim focuses on the GPU, it may not be best for users who want to study interactions with other system components (e.g., the CPU or other accelerators).
Accel-Sim also does not heavily focus on the GPU cache coherence or memory consistency.

\noindent
\textbf{gem5}~\cite{binkert2011gem5, LowePowerAhmad2020-gem520}: 
The gem5 simulator is another popular tool used in computer system research to evaluate novel hardware designs. 
It provides a robust API for researchers to modify and extend current models and to create new models in the gem5 infrastructure.
The gem5 simulator implements many models for system component including CPUs (out-of-order designs, in-order designs, and others), AMD and ARM GPUs~\cite{GutierrezBeckmann2018-gem5GPU}, accelerators~\cite{RogersSlycord2020-gem5Salam, SpencerRogers2024-gem5Salam2, ChaudhariSinclair2025-gem5Accel}, various memories, on-chip interconnects, coherent caches, I/O devices, and many others.
These gem5 models have enough fidelity to boot Linux, run unmodified workloads, and investigate cross-layer designs.

Thus, gem5 enables rapid prototyping of hardware-software co-designs across the computing stack.
For example, users can prototype optimizations to the compiler, OS, or runtime in tandem with architectural changes and study the implications of their design choices.
Like Accel-Sim, gem5 has an integrated power model~\cite{SmithBruce2024-gem5Power} and also supports running popular ML workloads both natively and through frameworks like PyTorch -- including adding support for advanced techniques to tradeoff simulation time for reduced fidelity in less important application regions~\cite{RamadasPoremba2023-gem5GPUFS, RamadasPoremba2024-gem5MLSim, RamadasSinclair2024-gem5MLSim}.
However, gem5's support for ML workloads differs in three key ways from Accel-Sim's.
First, unlike Accel-Sim, gem5's support for ML workloads spans across different types of compute devices, including CPUs and accelerators.
Second, gem5 currently focuses its support on AMD GPUs.
Since AMD's GPU runtime and drivers are open-source, this enables gem5 to model co-design between additional layers of the computing stack because it simulates all of those layers (unlike Accel-Sim).
Third and finally, gem5 also has highly accurate models for cache coherence, memory consistency, and interfaces between components in the system like the GPU's Command Processor.
Thus, gem5 may be a good choice for users wanting to study how ML workloads behave across system components or who want to prototype optimizations across layers of the computing stack.
However, since many users focus on NVIDIA GPUs and gem5 currently does not support them, users deeply tied to NVIDIA's ecosystem will not find it useful.

\noindent
\textbf{SST}~\cite{RodriguesHemmert2011-sst,SST}: Accel-Sim and gem5 focus on modeling a single GPU (Accel-Sim, gem5) or a single system-on-a-chip (gem5).
However, modern, large-scale computing systems frequently have hundreds or thousands of processors (e.g., GPUs) integrated together.
Thus, the Structural Simulation Toolkit (SST) is a good option for users who want to study ML workloads in rack-scale systems.
Instead of using high fidelity, but often slow models for components like processors (like Accel-Sim and gem5 do), SST utilizes analytical models for these components and focuses on modeling the network across many components, making it faster and scalable.
However, for users who want to focus on both smaller- and larger-scale systems, both Accel-Sim~\cite{sandia_2, sandia_3} and gem5~\cite{hsieh2012gem5sst, nguyen2022gem5sst} have integrated their models with SST -- potentially providing the best of both worlds.

\textbf{ExaDigiT}~\cite{brewer2024digital}: 
To study AI workloads at datacenter or supercomputer scale, ExaDigiT provides a holistic digital twin framework that models the coupled behavior of workloads, compute, power, and cooling subsystems.
Unlike simulators such as Accel-Sim, gem5, or SST, which operate at device- or node-level timescales, ExaDigiT enables large-scale modeling of system dynamics over operational timescales—capturing interactions that are difficult to observe or measure directly in production environments.
This framework further provides a means to evaluate operational strategies, perform “what-if” analyses, and uncover complex, cross-disciplinary transient behaviors that emerge from the tight coupling of workloads, compute, power, and cooling.

\begin{table*}[tb!]
  \caption{Example Simulation Tools that Benefit AI Benchmark Simulations.}
  \begin{center}
  \begin{tabular}
  {|p{0.15\textwidth}|p{0.25\textwidth}|p{0.25\textwidth}|p{0.25\textwidth}|}
  \hline
  \rowcolor{blue!30!white}
  Tool/Software & Scale & Benefits & Application\\
  \hline
  \hline
  Accel-Sim~\cite{KhairyShen2021-accelSim} & Single- and multi-\textbf{GPU} & High fidelity, usability, integrated power model, supports NVIDIA GPUs. & Examining power/performance tradeoffs; improving GPU efficiency.\\
  \hline
  gem5~\cite{binkert2011gem5, LowePowerAhmad2020-gem520} & Single- and multi-\textbf{CPU, GPU, and system-on-a-chip} & High fidelity, hardware-software co-design, models cache coherence, interconnects, and memory consistency, supports accelerators and AMD/ARM GPUs. & Studying ML workload behavior across components; prototyping optimizations across layers.\\
  \hline
  SST~\cite{RodriguesHemmert2011-sst, SST} & \textbf{Rack-scale} systems & Faster, scalable, models networking, utilizes analytical models. & Studying ML workload behavior in large-scale systems.\\
  \hline
  ExaDigiT~\cite{brewer2024digital} & \textbf{Datacenter- or supercomputer-level} & Models interactions among workloads, scheduling, power, networking, and cooling, including physical footprint. & Examining ML workload behavior at the largest scales.\\
  \bottomrule
  \end{tabular}
  \end{center}
\end{table*}

ExaDigiT consists of three coupled modules: (1) a \emph{resource allocator and power simulator (RAPS)} for replaying telemetry, simulating the scheduling of real or synthetic workloads, and dynamically estimating energy consumption; (2) a \emph{thermo-fluid cooling module} for predicting pressures, temperatures, flow rates, system-level control responses, and overall power-usage effectiveness (PUE); and (3) a \emph{visual analytics module} that integrates both a web-based dashboard and extended reality (XR) interfaces for immersive exploration of system behavior in augmented, virtual, or mixed reality. 

Operating at coarser timescales than cycle-accurate simulators, ExaDigiT enables comprehensive studies of power, cooling, and scheduling interactions across the full supercomputer. 
It has been applied to analyze how scheduling policies influence power and cooling dynamics~\cite{maiterth2025hpc}, used as a reinforcement learning environment for training optimal scheduling agents~\cite{brewer2025trace}, and to perform ``virtual'' benchmarking of large-scale LLM training workloads~\cite{kalepu2025virtual}.

\noindent
\textbf{Summary}: 
Collectively, these simulation frameworks span a continuum of modeling fidelity and scale—from device-level, cycle-accurate simulators such as Accel-Sim and gem5, to system- and datacenter-level models such as SST and ExaDigiT. 
By enabling controlled, repeatable, and cost-effective experimentation, they serve to democratize AI benchmarking in the design-space exploration of emerging architectures. 
As AI workloads continue to push the limits of power, cooling, and scheduling efficiency, such simulation-based tools will become indispensable for evaluating new ideas before committing to physical deployment.

%% file: 0090-share.tex
\section{Sharing Benchmarks}
\label{sec:share}

Beyond the creation of new benchmarks, \textit{sharing} benchmakrs is an essential aspect of benchmark carpentry. To this end, integrating the  FAIR principles is of paramount importance.

Benchmark sharing is best supported through hosting the code in a public repository that provides well-documented, executable workflows, thereby enabling others to reproduce the benchmark and compare results. Standard development practices, such as using Python Notebooks or scripts in other programming languages, as well as standard libraries, are recommended. More complex benchmarks may benefit from formal build processes (e.g., using makefiles) and dependency management through package managers. Containerization offers additional advantages, simplifying configuration and improving portability across environments.

To further support FAIRness, benchmark results should include standardized metadata, facilitating consistent comparison and analysis across studies.

While existing platforms such as Hugging Face and Kaggle provide mechanisms for sharing benchmarks, results, and leaderboards, fostering community capacity to host them independently remains valuable. Initiatives such as MLCommons illustrate how communities can maintain open, transparent benchmarking ecosystems. Educational efforts could be developed to train researchers and practitioners in these practices.

Finally, with the growing prominence of agentic AI, it is worth exploring its potential for automating the benchmarking lifecycle—including benchmark execution, result generation, and report synthesis. For example, the MLCommons Science Working Group is investigating how agentic AI can be applied to scientific benchmarks, particularly those involving time series analysis.

%% file: 0400-carpentry.tex
%
%
%
%

\section{Towards an AI Benchmark Carpentry Curriculum}
\label{sec:edu}

Based on the lessons learned and our observations from domain experts, we have devised the following exemplary curriculum addressing AI benchmark carpentry.

\begin{itemize}
    \item \textbf{Software Carpentry Foundational Tools and Practices:}

    Before addressing benchmark carpentry, we recommend that participants will review and learn about basic fundamental tools and practices. As they already exist as part of Software Carpentry, they can be reused. However, it may be of advantage to adapt certain aspects to explicitly utilize examples that focus on AI benchmarks and not just any arbitrary software carpentry project.
    
    \begin{itemize}
        \item Programming Skills: Proficiency in Python, Jupyter Notebooks, focusing on reproducible coding practices, including documentation, and reproducibility.
        \item Version Control: Git for tracking changes and collaboration.
        \item Command-Line Proficiency: Unix shell for efficient data manipulation.
        \item Data Management: Techniques for data cleaning, transformation, and visualization.
        \item Learning from Online AI/LLM Resources: Leveraging large language models and online tutorials for benchmarking insights and guidance.
    \end{itemize}
    
    \item \textbf{AI Benchmarking Fundamentals:}

    Having a basic understanding of AI Benchmarking is important for designing, evaluating, and improving AI systems. Benchmarks provide a standardized way to measure performance, compare models, and identify areas for optimization. By introducing benchmarking methodologies, examples, and metrics, participants gain the tools to critically assess AI models. Effective simple visualization practices help communicating results in a transparent, reproducible fashion related to real-world examples.

    \begin{itemize}
        \item Benchmarking Methodologies: Introduction to frameworks such as MLPerf and AIBench.
        \item Scenario-Based Benchmarks: Creating benchmarks that simulate real-world AI applications.
        \item Performance Metrics: Throughput, latency, accuracy, and resource utilization.
        \item Displaying Information with Graphs: Visualizing benchmark results for better analysis and interpretation.
    \end{itemize}

\item \textbf{Reproducibility and Experiment Management:}

Especially for benchmarks, it is not only important to document the code, but to document the results so we enable reproducibility. This includes documenting workflows and data provenance in case prior work and data are integrated. Thus, applying the FAIR principles—making data and experiments Findable, Accessible, Interoperable, and Reusable—enhances transparency and promotes collaboration across teams and institutions. 

\begin{itemize}
    \item Experiment Documentation: Importance of detailed documentation for reproducibility and adherence to FAIR principles.
    \item Automated Workflows: Using Docker and CI/CD pipelines to automate benchmarking processes.
    \item Data Provenance: Tracking data sources and transformations for transparency, traceability, and reuse.
    \item FAIR: Apply the FAIR principle to AI benchmarks.
\end{itemize}

    \item \textbf{Ethical Considerations and Bias Mitigation:}

    It is important to address the ethical implications of conducting Benchmarks. Here, we not just focus on societal impacts, but also on  the reporting of bias, fairness conducted potentially through hardware, software, and even vendor impacts.

    \begin{itemize}
        \item Bias Detection: Methods to identify and mitigate biases in AI models and datasets.
        \item Fairness Metrics: Metrics to assess and ensure fairness in AI systems.
        \item Ethical Implications: Discussion on societal impacts and ethical decision-making.
    \end{itemize}

    \item \textbf{Carpentry Principles in Practice:}

    A practical experience will be introduced to showcase the principles of AI benchmarking techniques. For this, a small, manageable datasets, and AI algorithm are used. The project may be conducted individually or in groups, while a walkthrough will also be available. An expansion to this AI-based benchmark will be the hosting and deployment of a leaderboard. Contributors can post their results in this shared leaderboard for the compute systems they have access to.
    
    \begin{itemize}
        \item Hands-On Workshops: Practical sessions applying benchmarking techniques to real datasets.
        \item Collaborative Projects: Group projects to foster teamwork and problem-solving skills.
        \item Open-Source Contributions: Participation in community AI benchmarking initiatives.
    \end{itemize}

    \item \textbf{Special Topics:}

    As we have seen from the previous section, several aspects have a great impact on AI benchmarking, which is so far not covered by other carpentry efforts. This includes energy benchmarking, simulation of hardware to estimate performance, and performance tuning with a focus on AI. Instead of just setting up a leaderboard through, for example, a Docker container, selected parties may have an interest in finding out more about setting up such leaderboards and hosting them.
    
    \begin{itemize}
        \item Energy Efficiency: Measuring power consumption and optimizing AI workloads for lower energy usage.
        \item Simulation: Using synthetic data and simulated environments for benchmarking when real data is limited.
        \item Performance Tuning: Techniques for optimizing model execution, hardware utilization, and system throughput.
        \item Leaderboard Management: Designing, maintaining, and validating AI benchmark leaderboards for reproducibility and fairness.
        \item To provide users a starting point, presenting the community with a collection of benchmarks can be useful and has been spearheaded at \cite{www-las-mlcommons-benchmark-coolection}.
    \end{itemize}
\end{itemize}

From the extensive surveys and numerous examples it is important that to start one ought to begin with the most elementary efforts and grow them continuously. As such, we recommend adding specific lessons when we discover they need to be added by the community. Also, we must involve the community itself and allow for contributions of tutorials from a wide variety of groups.

%% file: 0410-democratizing.tex
%
%

\section{Towards AI Benchmark Democratizing}
\label{sec:dem}

Our goal is to make AI benchmarking transparent, reproducible, and community-driven. Democratization empowers a broader range of participants to contribute to and learn from AI performance evaluation.

Introducing democratization tools, datasets, and evaluation frameworks that are openly accessible and easy to use can allow anyone—from students to independent researchers—to measure, compare, and improve AI models. 

One of the biggest hurdles we find is that some benchmarks, probably rightfully so, target hyperscale or leadership-class machines. However, in order to increase the community and raise awareness, smaller scale benchmarks need to be available.

As such, the following aspects can improve democratization:

\begin{itemize}
    \item \textbf{Accessibility:}
    \begin{itemize}
        \item Benchmarks, datasets, and tooling ought to be open-source or freely available.
        \item Users may not need to rely on expensive hardware or proprietary software to participate.
        \item Examples can be leveraged to develop new benchmarks. One can start with examples provided by MLCommons open datasets, pre-built benchmarking pipelines, and Jupyter notebooks with ready-to-run benchmarks.
    \end{itemize}

    \item \textbf{Usability:}
    \begin{itemize}
        \item Interfaces, documentation, and examples in existing efforts can serve as starting point to developing user-friendly, allowing non-experts to run benchmarks.
        \item Providing automated scripts and tutorials reduces the barrier to entry.
    \end{itemize}

    \item \textbf{Transparency:}
    \begin{itemize}
        \item Specifying clear definitions of metrics, scoring methods, and evaluation procedures ensures everyone understands the results.
        \item Improved transparency addresses the hide everything in a “black box” approach,  where only insiders can interpret outcomes.
    \end{itemize}

    \item \textbf{Community Participation:}
    \begin{itemize}
        \item Anyone with minimal but sufficient knowledge should be able to contribute to benchmarks, improve tools, or submit models.
        \item Democratization also means encouraging collaboration and reproducibility across institutions and geographies (e.g., engaging the broader community). 
    \end{itemize}

    \item \textbf{Impact:}
    \begin{itemize}
        \item Through democratization, smaller teams or educational institutions can contribute and benefit from learning, competing, and comparing AI benchmarks.
        \item Through democratization, fairness and innovation is fostered because knowledge and evaluation methods are disseminated.
    \end{itemize}
\end{itemize}

%% file: 0530-conclusion.tex
\section{Conclusion}
\label{sec:conclusion}

Overall, this comprehensive paper has explored the motivations and pathways for creating a holistic benchmark  carpentry effort, paying specific attention to aspects that can democratize AI benchmarks.
This was achieved by (a) providing standardized and formal definitions of benchmarks, and (b) identifying a representative set of benchmarks related to AI activities.
Finally, we propose an AI Benchmark Carpentry curriculum that integrates the various topics discussed into a structured learning activities 
to empower  
practitioners with reproducible coding practices, experiment‑management 
skills, and an ethical lens on benchmarking. By embedding FAIR principles, 
bias‑mitigation techniques, and performance‑tuning modules, the curriculum 
offers a scalable pathway for communities—from academic labs to industry 
R\&D—to build, share, and improve benchmarks in a collaborative, 
transparent manner.

Together, these activities foster democratization of AI benchmarks and can be utilized to grow the community and the understanding on how benchmarks may effect an individual activity or even community. While deploying such activities, we hope to grow community awareness and overcome the lack of well defined activities to educate the community in this regard. While fostering these activities we also address the need for more easily develop dynamic and adaptable benchmarks.

%% file: 0910-acknowledgments.tex
\section*{Acknowledgment}

We have used at one point ``ChatGPT'' to improve upon the grammar of selected sections with the question: ``Improve the grammar of ...''. However, we stopped that practice early on due to wrong corrections, and have used Grammarly throughout the paper.

The work was in part sponsored by NSF Grant \#2346173 and \# 2303700, POSE: Phase II: MLCommons Research for Science: Enabling Open-Source Ecosystems for Scientific Foundation Models by Community Standards and Benchmarks

The portion of this work done at UW-Madison is supported in part by NSF grant CNS-2312688 and by the U.S. Department of Energy, Office of Science, Office of Advanced Scientific Computing Research, under Award Number DE-SC-0026036.

This manuscript has been in part authored by FermiForward Discovery Group, LLC under Contract No. 89243024CSC000002 with the U.S. Department of Energy, Office of Science, Office of High Energy Physics.
Fermilab Report Number FERMILAB-PUB-25-0835-CSAID. 

This work was supported by DOE ASCR Microelectronics Science Research Center Projects, BIA.  This material is based upon work supported by the U.S. Department of Energy, Office of Science, under contract number DE-AC02-06CH11357.

This research was in part sponsored in part by and used resources of the Oak Ridge Leadership Computing Facility (OLCF), which is a DOE Office of Science User Facility at the Oak Ridge National Laboratory (ORNL) supported by the U.S. Department of Energy under Contract No. DE-AC05-00OR22725.

Shirley Moore's work on this paper was supported by the Department of Energy Office of Science under award \#DE-SC0024352.

Kirkpatrick's work was made possible through the National Science Foundation award \#2226453.

This research was funded in part by and used resources at the Argonne Leadership Computing Facility, which is a DOE Office of Science User Facility supported under Contract DE-AC02-06CH11357.

Research was sponsored by the Department of the Air Force Artificial
Intelligence Accelerator and was accomplished under Cooperative
Agreement Number FA8750-19-2-1000. The views and conclusions contained
in this document are those of the authors and should not be interpreted
as representing the official policies, either expressed or implied, of
the Department of the Air Force or the U.S.\ Government.
The U.S.\ Government is authorized to reproduce and distribute reprints
for Government purposes notwithstanding any copyright notation herein.

Gary Mazzaferro has participated in discussions as part of the working group meetings surrounding benchmark definitions and applicability.

%% file: 0930-glossary.tex
\section*{Appendix A. Abbreviations}
\label{sec:abbrev}

\begin{description}

\item[$\bf A, P$] Application and Parameters – Formal components defining a scientific task $T = (A, P)$.
\item[AI] Artificial Intelligence – Computational systems performing tasks that typically require human intelligence, such as learning, reasoning, or perception.
\item[API] Application Programming Interface – A set of functions and protocols for building and integrating software applications.
\item[ARC] AI2 Reasoning Challenge – Benchmark for commonsense and science question answering at the K–12 level.
\item[AristoBench] A benchmark based on U.S. science exam questions.
\item[$\bf B$] Benchmark – A formal specification of an evaluation setup $B = (I, D, T, M, C_B, R)$.
\item[BioASQ] A benchmark for biomedical question answering and document retrieval.
\item[BSP] Bulk synchronous programming
\item[CAPEX/OPEX] Capital Expenditure / Operating Expenditure – Financial measures of investment and operational costs.
\item[CarbonTracker] Tool for tracking and reporting ML training energy use and CO$_2$ emissions.
\item[$\bf C_B$, $C_c$] Benchmark and Component Constraints – Rules applied to benchmarks or components for fairness, reproducibility, and comparability.
\item[CodeCarbon] Python package estimating CO$_2$ emissions based on compute energy and grid intensity.
\item[CO$_2$e] Carbon Dioxide Equivalent – Standardized unit comparing greenhouse gas emissions by global warming potential.
\item[cPUE] Compute-Only Power Usage Effectiveness – Refinement of PUE focusing on compute power versus total facility power.
\item[CORE-Bench] A benchmark for computational reproducibility.
\item[CPU] Central Processing Unit – The main processor of a computer, optimized for sequential execution.
\item[CSR] Corporate Social Responsibility – A framework for a company’s environmental and social accountability.
\item[Data Carpentry] A Carpentries initiative focused on teaching data management and analysis skills.
\item[DCGM] Data-Centre GPU Manager – NVIDIA’s toolkit for GPU monitoring, management, and power measurement.
\item[DCiE] Data-Centre-infrastructure Efficiency – Metric defined as $P_{\mathrm{IT}} / P_{\mathrm{fac}}$, the reciprocal of PUE.
\item[DeepONet] Deep Operator Network – A neural architecture for learning operators mapping between function spaces.
\item[DOE] U.S. Department of Energy – Federal agency supporting large-scale AI and HPC research.
\item[DOI] Digital Object Identifier – Persistent identifier for published datasets, software, and benchmarks.
\item[DVFS] Dynamic Voltage and Frequency Scaling – Power management technique adjusting CPU/GPU frequency to balance performance and energy.
\item[EDP] Energy–Delay Product – A metric combining performance (time-to-solution) and energy efficiency.
\item[EIA] Energy Information Administration – U.S. agency providing official energy statistics and analysis.
\item[EE-HPC WG] Energy Efficient HPC Working Group – Community standardizing energy measurement and reporting in HPC.
\item[ElectricityMap] API providing real-time carbon intensity of electricity grids worldwide.
\item[Energy-to-Solution] Total energy consumed to complete a computational task.
\item[EU] European Union – Political and economic union of 27 member states.
\item[FAIR] Findable, Accessible, Interoperable, Re-usable – Principles for data and software management.
\item[FNO] Fourier Neural Operator – Deep learning architecture for solving partial differential equations efficiently.
\item[Galactica Eval] A benchmark measuring scientific writing and knowledge recall capabilities.
\item[GFLOPS] Giga Floating-Point Operations per Second – Measure of computational performance.
\item[GFLOPS/W] GFLOPS per Watt – Energy efficiency metric used in Green500.
\item[GPU] Graphics Processing Unit – Specialized processor for graphics and parallel computation, widely used in ML and HPC.
\item[Green500] Ranking of the most energy-efficient supercomputers in the world.
\item[HPC] High-Performance Computing – The use of supercomputers and parallel processing to solve large, complex problems.
\item[HPC AI500] Benchmark suite evaluating deep learning performance at HPC scale.
\item[HPC Carpentry] Carpentry modules teaching HPC skills and workflows.
\item[HPCG] High Performance Conjugate Gradients --- a benchmark that
represents real-world scientific workloads, such as partial
differential equations' solvers, better than the LINPACK benchmark.
\item[HPCG-Power] Energy-aware version of the HPCG benchmark.
\item[HPL] High-Performance LINPACK Benchmark – Measures floating-point computing power.
\item[HPL-MxP] Mixed-Precision variant of the LINPACK benchmark that
uses multiple floating-point formats for improved FLOP and energy metrics.
\item[IO500] Benchmark suite evaluating HPC storage performance.
\item[IOPS] Input/Output Operations Per Second – Metric measuring storage system performance.
\item[$\bf I$] Infrastructure – Hardware and software environment required to execute a benchmark.
\item[$\bf J/epoch$] Joules per training epoch – ML training energy efficiency metric.
\item[$\bf J/sample$] Joules per sample – Energy consumed per inference or training sample.
\item[$\bf J/step$] Joules per simulation step – Energy metric for iterative simulation workloads.
\item[JouleSort] Benchmark measuring energy efficiency of sorting tasks.
\item[Kaggle] Online platform hosting AI and data science competitions.
\item[Kepler] Kubernetes-based Efficient Power Level Exporter – Collects power metrics in containerized workloads.
\item[KPI] Key Performance Indicator – Quantitative metric for performance or efficiency.
\item[kg CO$_2$e] Kilograms of CO₂ equivalent – Emission unit measuring carbon footprint.
\item[kg CO$_2$e/job] Kilograms CO₂e emitted per completed computational job.
\item[$kWh$] Kilowatt-hour – Standard unit of electrical energy.
\item[LAB-Bench] Benchmark evaluating AI models on laboratory research assistance tasks.
\item[LLM] Large Language Model – AI model trained on massive text corpora, capable of natural language generation.
\item[ML] Machine Learning – A subset of AI focused on algorithms that improve automatically through data.
\item[MLCommons] Open engineering consortium developing ML benchmarks, datasets, and best practices.
\item[MLflow] Open-source platform for managing the ML lifecycle, including experimentation and deployment.
\item[MLPerf] MLCommons benchmark suite for ML training and inference performance.
\item[MLPerf HPC] MLPerf extension for measuring ML workloads in HPC environments.
\item[MLPerf Power] MLPerf extension tracking energy efficiency (e.g., Joules/sample).
\item[MLPerf Tiny Power] MLPerf subset focused on energy use in constrained edge devices.
\item[MMLU] Massive Multitask Language Understanding benchmark – Evaluates cross-domain academic knowledge.
\item[MPI] Message Passing Interface – Communication standard for parallel programming in distributed-memory systems.
\item[M] Metrics – Quantitative measures such as accuracy, throughput, latency, or energy.
\item[NVML] NVIDIA Management Library – API for monitoring and managing NVIDIA GPU performance and power.
\item[OmniStat] AMD’s monitoring and management interface, similar to NVIDIA’s NVML.

\item [$P_{fac}$]  Facility Power

\item [$P_{IT}$] IT Equipment Power

\item[PDEBench] Benchmark for solving partial differential equations using ML and operator learning.
\item[PM\_COUNTER] Node-level power counter (e.g., HPE-Cray) for system-level power tracking.
\item[PowerAPI] Software library and framework for energy measurement and analysis.
\item[PPA] Power-Purchase Agreement – Contract defining terms for renewable energy procurement.
\item[PUE] Power Usage Effectiveness – Data center efficiency metric $P_{\mathrm{fac}} / P_{\mathrm{IT}}$.
\item[PTDaemon] SPEC Power Temperature Daemon – Tool for calibrated power measurement in benchmarks.
\item[PyTorch Profiler] Tool for performance analysis of PyTorch models.
\item[QA] Question Answering – NLP task where systems respond to questions posed in natural language.
\item[R] Programming language for statistical computing and data analysis.
\item[RAPL] Running Average Power Limit – Intel/AMD interface for on-chip power measurement.
\item[RFP] Request for Proposal – Formal request soliciting project bids or benchmark implementations.
\item[$R$] Results – Outputs of a benchmark including accuracy, energy, and performance metrics.
\item[SCIQ] Science Question Answering benchmark focusing on inference and reasoning.
\item[SciEval] Benchmark for multi-domain scientific reasoning.
\item[SciSafeEval] Benchmark measuring safety alignment of AI systems in scientific contexts.
\item[Scaphandre] Open-source energy monitoring agent integrating with Prometheus.
\item[SERT] Server Efficiency Rating Tool – SPEC benchmark for server energy efficiency.
\item[Slurm] Simple Linux Utility for Resource Management – Widely used HPC workload manager.
\item[SM] Streaming Multiprocessor – GPU core building block for parallel computation.
\item[SME] Small and Medium-sized Enterprise – Business classification by size and revenue.
\item[Software Carpentry] Community teaching foundational computing skills for researchers.
\item[SKU] Stock-Keeping Unit
\item[SPEC] Standard Performance Evaluation Corporation – Organization developing performance benchmarks.
\item[SPEC HPC] SPEC benchmark suite for HPC system performance.
\item[SPECpower] SPEC benchmark suite measuring server energy efficiency.
\item[SQL] Structured Query Language – Standard for managing and querying relational databases.
\item[STREAM] Synthetic benchmark measuring sustainable memory bandwidth.
\item[$T$] Scientific Task – Core task evaluated by a benchmark, e.g., classification or prediction.
\item[TDP] Thermal Design Power – Maximum heat generated by a processor under typical workloads.
\item[TensorBoard] Visualization toolkit for TensorFlow training and profiling.
\item[TorchInfo] Python package summarizing PyTorch models (layers, parameters, memory use).
\item[TOP500] Official list of the 500 most powerful supercomputers in the world.
\item[TPC-Energy] Benchmark suite for transaction workloads with energy measurement.
\item[TVA] Tennessee Valley Authority – U.S. regional energy provider supporting grid research.
\item[Vampir] Performance analysis tool for parallel applications with detailed execution tracing.
\item[VTune] Intel VTune Profiler – Performance and energy analysis tool for fine-grained profiling.
\item[WattTime] API providing marginal emissions signals for carbon-aware workload scheduling.
\item[Wh/DB-phase] Watt-hours per database phase – Metric used in TPC-Energy benchmarks.

\end{description}

%% file: 0940-contribution.tex
\section*{Appendix B. Contributions}
\label{sec:contrib}

\definecolor{darkgreen}{rgb}{0.0,0.5,0.0}

\newcommand{\AUTHOR}[2]{%
    \item \textbf{%
        \ifthenelse{\equal{#2}{notified}}%
            {\parindent0pt #1}%
            {\textcolor{red}{x #1}}%
    }%
}

\begin{itemize}[label={}, leftmargin=0pt]
    
\AUTHOR{Gregor von Laszewski}{notified} is the lead author of the paper. He identified first that efforts in benchmark carpentry and democratization are needed. He has lead the organization of this paper in the MLCommons Science Working Group. He also created the initial version of \cite{www-las-mlcommons-benchmark-coolection} which is related and relevant to this effort.

\AUTHOR{Piotr Luszczek}{notified} has contributed to integration of many
decades of experiences from designing, implementing, running, and
collecting results from HPC benchmarks.

\AUTHOR{Wesley Brewer}{notified} has contributed to the simulation section.

\AUTHOR{Jeyan Thiyagalingam}{notified} has worked on the GPU benchmarking section. Reviewed the paper and made corrections.

\AUTHOR{Juri Papay}{notified} has worked on  the GPU benchmarking section. Updated the GPU HW details of MLCommon benchmarks.

\AUTHOR{Geoffrey C. Fox}{notified} is leading the MLCommons Science Working group and has contributed to many of the ideas. The experiences and discussions with Gregor von Laszewski around improvements to the earthquake benchmark have significantly contributed to this effort. The educational effort of using the earthquake benchmark with a number of students motivated this effort.

\AUTHOR{Armstrong Foundjem}{notified} has provided an early version and led the Energy section, and has contributed to the paper writing and overall improvement.

\AUTHOR{Gregg Barrett}{notified} has participated in discussions as part of the working group meetings and contributed to an early version of this paper.

\AUTHOR{Murali Emani}{notified} has participated in discussions as part of the working group meetings and improved the article.

\AUTHOR{Shirley V. Moore}{notified} has written text for the Profiling and Performance Analysis Section. 

\AUTHOR{Vijay Janapa Reddi}{notified} has participated in discussions as part of the working group meetings and improved the article.

\AUTHOR{Matthew D. Sinclair, Shivaram Venkataraman and Rutwik Jain} {notified} participated in discussions as part of the working group meetings and wrote the variability section of the article. Sinclair also wrote the simulation section of the paper and helped improve the paper in other sections.

\AUTHOR{Christine Kirkpatrick}{notified} has worked on conceptualizing the ideas and discussion, and helping with the Carpentries background section.

\AUTHOR{Kartik Mathur}{notified} Has worked on improving an early version of the Energy section.

\AUTHOR{Victor Lu}{notified} Has participated in writing the paper.

\AUTHOR{Tianhao Li}{notified} has participated in discussions as part of the working group meetings and participated in identification of limitations of current benchmarks.

\AUTHOR{Sebastian Lobentanzer}{notified} 
Has participated in the discussions in the working group and has contributed to the abstract, intro, definitions, formalization, and benchmark sections with content and editing.

\AUTHOR{Sujata Goswami}{notified} has worked on the MLCommons benchmark details in Table \ref{tab:benchmarks-mlcommons}.

\AUTHOR{Abdulkareem Alsudais}{notified} has reviewed the motivation to AI Benchmark Carpentry and contributed to the writing of this paper.

\AUTHOR{Kongtao Chen}{notified} has worked on the monitoring sections, related benchmarks, and participated in discussions as part of the working group meetings.

\AUTHOR{Tejinder Singh}{notified} has edited and improved AI hardware benchmarking and infrastructure sections and provided new KPIs for AI hardware benchmarking. 

\AUTHOR{Kirsten Morehouse
knmorehouse@gmail.com}{notified} has participated in discussions as part of the working group meetings. Morehouse also reviewed
the paper and made improvements.

\AUTHOR{Marco Colombo, Benjamin Hawks, and Nhan Tran}{notified} have worked on the benchmark ontology and Table \ref{tab:ontology}.

\AUTHOR{Khojasteh Z. Mirza}{notified} has participated in discussions as part of the working group meetings and worked an a very early version of the energy section.

\AUTHOR{Renato Umeton}{notified}
revised the manuscript for consistency and coherence.

\AUTHOR{Sasidhar Kunapuli and Gavin Farrell
gavinmichael.farrell@phd.unipd.it}{notified} have participated in discussions as part of the working group meetings.

\end{itemize}